\documentclass[fleqn,10pt]{wlscirep}
\usepackage[utf8]{inputenc}
\usepackage[T1]{fontenc}
\usepackage{multirow}
\usepackage{placeins}
\usepackage{hyperref}

\title{Vision-based Deep Learning Analysis of Unordered Biomedical Tabular Datasets via Optimal Spatial Cartography}

\author[1]{Sakib Mostafa}
\author[2]{Tarik Massoud}
\author[1,3,4]{Maximilian Diehn}
\author[1,*]{Lei Xing}
\author[1,*]{Md Tauhidul Islam}
\affil[1]{Department of Radiation Oncology, Stanford University, Stanford, California, USA}
\affil[2]{Department of Radiology, Stanford University, Stanford, California, USA}
\affil[3]{Institute for Stem Cell Biology and Regenerative Medicine, Stanford University, Stanford, California, USA}
\affil[4]{Stanford Cancer Institute, Stanford University, Stanford, California, USA}

\affil[*]{Correspondence: tauhid@stanford.edu and lei@stanford.edu}

\begin{abstract}
Tabular data constitutes a dominant representation in biomedical science, underpinning applications ranging from liquid biopsy and bulk and single-cell transcriptomics to structured electronic health records and engineered phenotypic descriptors. Yet, unlike images, texts, or time series, tabular datasets lack intrinsic spatial organization: features are treated as unordered dimensions, and their interrelationships must be inferred implicitly by learning algorithms. This fundamental structural limitation constrains the ability of convolutional neural networks, vision transformers, and other spatially aware vision architectures to exploit local correlations and higher-order interactions that encode underlying biological and clinical signals. Here we introduce Dynamic Feature Mapping (Dynomap), an end-to-end deep learning framework that learns a task-optimized spatial topology of features directly from data. Dynomap discovers inter-feature dependencies and jointly optimizes their spatial arrangement with a predictive objective through a fully differentiable rendering mechanism, without reliance on heuristics, predefined feature groupings, or external priors. By transforming high-dimensional tabular vectors into learned feature maps, Dynomap enables vision-based deep learning architectures to operate effectively on non-spatial biomedical inputs and achieves substantially better predictive performance compared to existing state-of-the-art (SOTA) methods across several clinical and biological datasets. When applied to liquid biopsy datasets, Dynomap autonomously organizes clinically actionable gene signatures into coherent spatial structures and improves multiclass cancer subtype prediction accuracy by up to 18\% relative to classical and SOTA deep learning tabular analysis methods. In a Parkinson’s disease voice dataset, Dynomap spatially clusters disease-associated acoustic descriptors—including tunable Q-factor wavelet energy and entropy measures linked to vocal instability—and yields accuracy gains of up to 8\% compared to SOTA techniques. Similar performance improvements and biologically meaningful feature organizations were observed when Dynomap was applied to datasets from other biomedical domains. By converting unordered feature spaces into learned spatial representations, Dynomap establishes a general and principled strategy for bridging tabular and vision-based deep learning and enables the discovery of structured, clinically actionable patterns from unordered high-dimensional biomedical data.
\end{abstract}

\begin{document}

\flushbottom
\maketitle

\thispagestyle{empty}

\section{Introduction}
Tabular data \textit{constitute} a central and fundamental representation in biomedical research and other data-intensive scientific domains, encompassing gene expression profiles, liquid biopsy measurements, clinical variables, and engineered phenomic descriptors~\cite{Hasin2017,Heitzer2019,Wan2017}. In these row-column settings, each sample is represented as a high-dimensional feature vector that captures complex biological, technical, and environmental characteristics of the system under study~\cite{Hasin2017,Libbrecht2015}. Although such representations are information-rich, a fundamental limitation is their lack of intrinsic structural organization~\cite{Gorishniy2021,Grinsztajn2022}. Unlike images, texts, signals, or time series data, tabular inputs contain no explicit encoding of relationships among features, treating variables as unordered dimensions in a high-dimensional space~\cite{Gorishniy2021,Grinsztajn2022,Shavitt2018}. In these unordered tabular representations, relationships among biomedical variables arising from correlated sources such as coordinated molecular programs, shared regulatory pathways, and physiologic dependencies are not reflected. As a result, dependencies, interactions, and higher-order biological structure must be inferred entirely by the learning algorithm, without leveraging inherent inductive biases such as locality, compositionality, or spatial coherence in the data~\cite{LeCun2015,Goyal2022}. This limitation substantially reduces model capabilities and leads to suboptimal performance in many practical applications~\cite{Gorishniy2021,Grinsztajn2022}. This issue is particularly consequential in biomedical research, where data are often acquired in the presence of substantial technical noise, pronounced biological heterogeneity, and severe class imbalance~\cite{Heitzer2019,DagogoJack2018,Marusyk2010}. In these settings, the absence of an explicit structural prior makes pattern discovery and interpretation especially challenging, as models must simultaneously denoise measurements, identify coordinated feature interactions, and generalize from limited samples~\cite{Grinsztajn2022,Libbrecht2015}.

There are several categories of existing methods for analyzing tabular biomedical data. First, linear models such as logistic regression combine features using a simple weighted sum. They are suitable when feature relationships in the dataset are mostly linear, but they fail to model complex interactions among features that are common in real biomedical datasets~\cite{Hosmer2013,Hastie2009}. Second, tree-based ensemble methods such as Random Forests, AdaBoost, and XGBoost model nonlinear relationships by repeatedly splitting the data into smaller regions based on feature thresholds. They can model the nonlinear relationships among features better than linear models, but compared to advanced models such as deep learning, they lack hierarchical representation learning and cannot automatically discover complex latent feature structures or transferable embeddings from raw high-dimensional data. As a result, they may struggle to capture subtle nonlinear interactions and cross-modal feature relationships that neural networks can model more flexibly and scalably~\cite{Breiman2001,Freund1997,Chen2016}. Third, distance and kernel-based methods such as k-nearest neighbors (KNN) and support vector machines (SVM) compare samples to one another based on overall similarity. Because they operate in sample space rather than organizing relationships among features, they fail to consider complex feature relationships and offer little insight into how the predictive variables themselves interact~\cite{Cover1967,Cortes1995}. Fourth, neural network models such as MLPs and modern tabular approaches like TabM or ModernNCA learn complex nonlinear interactions by transforming features into high-dimensional embeddings. However, the structure they learn remains implicit within the network and may not be optimal. Without an explicit structural map of feature relationships from these methods, it remains a challenge to explain their decision-making process~\cite{Gorishniy2021,Grinsztajn2022,Kadra2021}. Fifth, foundation-style models such as TabPFN rely on strong meta-learned priors~\cite{Hollmann2022}. However, they still process features as an unordered vector and do not learn an explicit organization that reveals how features interact~\cite{Gorishniy2021,Grinsztajn2022}. Sixth, fixed vector-to-image methods such as DeepInsight~\cite{sharma2019deepinsight}, IGTD~\cite{zhu2021converting}, NCTD~\cite{alenizy2025transforming}, and REFINED~\cite{bazgir2020representation} create a spatial layout before training and then apply convolutional models on top of it. Because this layout is predefined and not learned with the task, it cannot adapt or reorganize to reflect the specific biological interactions relevant to the prediction objective~\cite{Picard2021,Kossen2021}. Together, these limitations highlight a representational gap. Current methods either treat features as unstructured or impose a fixed structure, underscoring the need to learn a task-adaptive feature organization that makes the spatial arrangement of predictive biological signals directly interpretable~\cite{Bengio2013}.

\begin{figure}[pt]
\centering
\includegraphics[width=0.95\linewidth, page = 2]{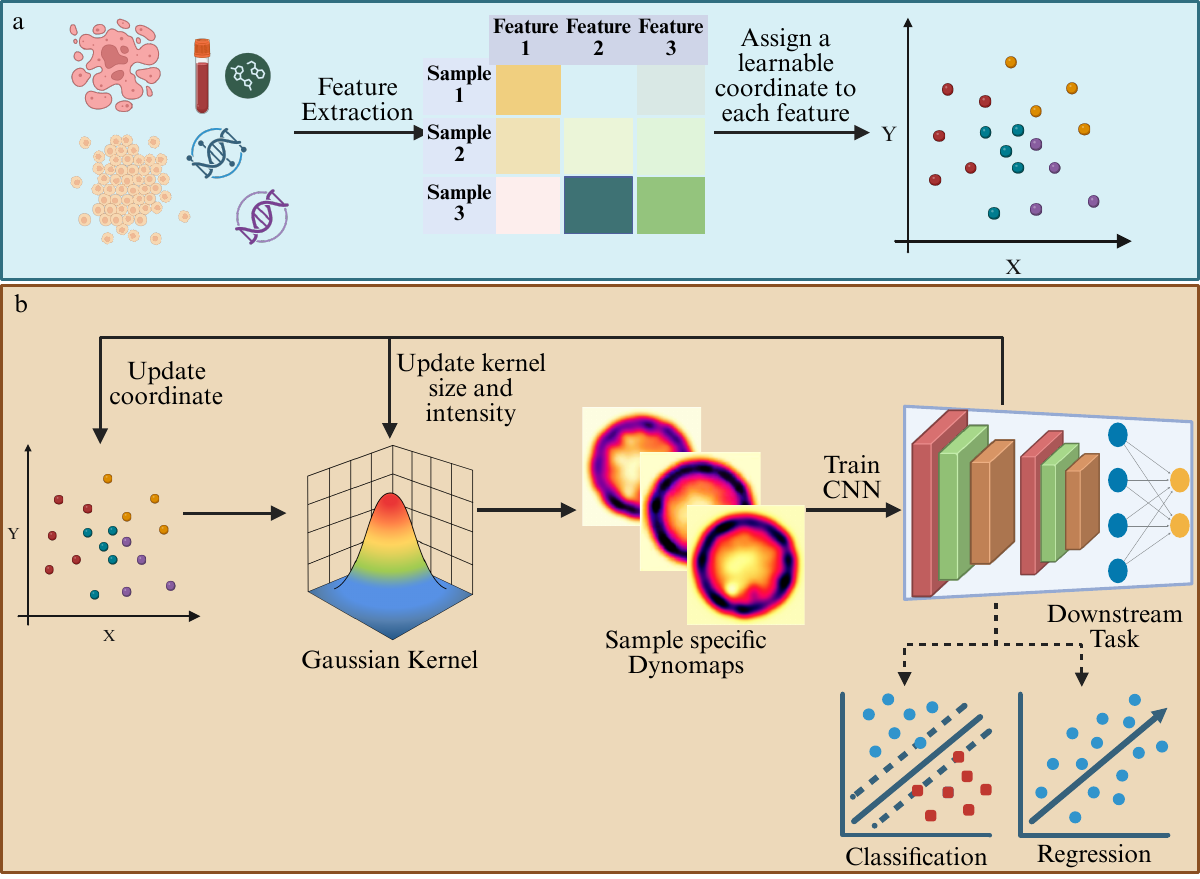}
\caption{\textbf{Overview of the Dynomap framework for spatial representation learning of tabular data.}
\textbf{a}, Tabular biomedical measurements are first represented as a feature matrix in which each column corresponds to a biological variable (e.g., gene expression, molecular measurement, or phenotypic descriptor) and each row corresponds to a sample. Dynomap assigns each feature a learnable coordinate in a continuous two-dimensional space, allowing relationships among variables to be represented as spatial proximity. 
\textbf{b}, During training, feature coordinates and rendering parameters are optimized jointly with the predictive model. Each feature contributes to the image through a Gaussian kernel whose location and intensity depend on the learned coordinates and feature values, producing a continuous spatial map for each sample (Dynomap). These sample-specific spatial representations are processed by a convolutional neural network to perform downstream tasks such as classification or regression. Because the spatial layout is learned directly from the prediction objective, features that contribute to the same biological signal tend to form coherent spatial neighborhoods.}
\label{fig:fig1_pipeline}
\end{figure}

Deep learning architectures that utilize spatial relationships among features, such as convolutional neural networks (CNNs) and vision transformers (ViTs), have demonstrated strong capabilities in domains including natural images and structured signals~\cite{LeCun2015,Esteva2021}. These models leverage locality, hierarchical composition, and parameter sharing to model structured patterns of input data efficiently~\cite{LeCun1998,LeCun2015}. Such architectural properties are particularly effective when relationships among input dimensions are organized in space, allowing computation of local correlations and progressive abstraction of correlation signals through the layers of the network for reliable decision-making~\cite{Goodfellow2016}. In these networks, spatial proximity enables convolutional filters or attention mechanisms to capture complex and higher-order interactions that would be difficult to capture in an unstructured vector representation~\cite{LeCun2015,Bengio2013}. Establishing a data-driven spatial arrangement of features is therefore essential for enabling spatial architectures to capture coordinated biological or phenotypic interactions in tabular data~\cite{sharma2019deepinsight,Picard2021}.

Here, to address the limitations of existing methods and develop a general strategy for transforming unordered biomedical tabular data into a spatially organized format suitable for vision deep learning models, we introduce Dynomap, a framework that learns a spatial organization of tabular features in an end-to-end manner (Fig.~\ref{fig:fig1_pipeline}). Dynomap assigns each feature a learnable coordinate in a two-dimensional continuous space. Individual samples are rendered into images through a fully differentiable process, in which feature values modulate localized kernels placed at their learned positions. The feature layout and the predictive model are optimized jointly through backpropagation, allowing the spatial arrangement of features to evolve in response to the prediction objective. In this formulation, spatial neighborhoods emerge directly from the structure of the predictive signal rather than being imposed in advance. This design enables convolutional or vision transformer models to operate directly on tabular data while preserving feature identity and supporting direct visualization of how predictive biological signals are organized across features. By coupling spatial representation learning with prediction, Dynomap provides an adjustable layout transformation of unordered tabular data. The learned spatial organization is task-adaptive and can reorganize as objectives change, allowing coherent biological or phenotypic modules to emerge from the data. This representation supports interpretation through direct visualization and attribution analysis, enabling examination of how predictive signal is distributed across features and spatial neighborhoods.

In the following sections, we apply Dynomap across diverse biomedical datasets spanning liquid biopsy transcriptomics, bulk RNA sequencing, single-cell expression profiling, and engineered phenomic measurements. Across datasets varying in dimensionality, class imbalance, and biological complexity, Dynomap reveals structured and task-adaptive spatial organization of features, enabling identification of coherent, class-dependent biological modules across molecular and phenomic domains. The learned representations exhibit spatial neighborhoods in which predictive variables cluster according to disease state, cellular lineage, or phenotypic condition. These spatial patterns are supported by attribution analysis and null-controlled spatial statistics, indicating that the learned topography reflects biologically meaningful signal rather than arbitrary arrangement. By jointly learning feature coordinates and prediction in an end-to-end framework, Dynomap provides an adaptive spatial representation of tabular data enabling convolutional models to operate effectively on traditionally non-spatial inputs while preserving interpretability and facilitating discovery of clinically relevant patterns.

\section{Results}
\subsection{Dynomap outperforms classical and modern tabular models in early cancer diagnosis and biomarker discovery on an ultrasensitive cfRNA profiling dataset}
First, we evaluated Dynomap on the RARE-Seq cfRNA liquid biopsy dataset~\cite{Nesselbush2025}. RARE-Seq is an ultrasensitive profiling platform designed to detect low-abundance tumor-derived transcripts in plasma. cfRNA measurements typically exhibit low signal-to-noise ratios, pronounced biological heterogeneity across patients, and substantial class imbalance~\cite{Heitzer2019,Wan2017,Marusyk2010}. To examine the effect of the number of genes on the performance and representations of different approaches, we considered two regimes. For the whole-transcriptome scenario, we used 4,000 highly variable genes (HVGs) from the dataset of over 60,000 genes. In the targeted transcriptome setting, we restricted inputs to the 622-gene signature panel defined in the original study~\cite{Nesselbush2025}. This setup enables direct comparison between a broad transcriptomic regime and a compact, biologically curated feature space.

\begin{figure}[pt]
\centering
\includegraphics[width=0.95\linewidth, page = 3]{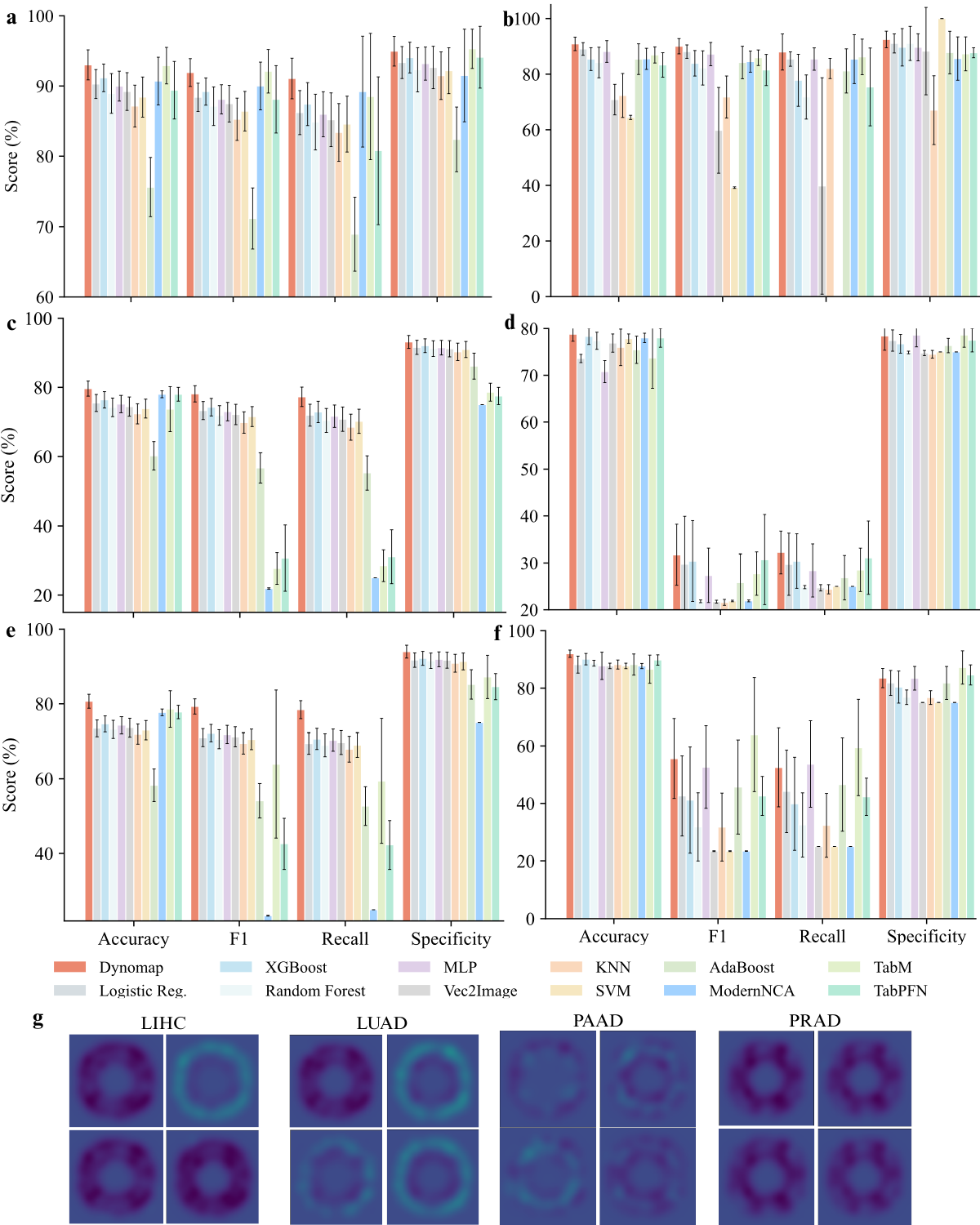}
\caption{\textbf{Dynomap performance and learned representations for liquid biopsy cfRNA classification.} 
\textbf{a–f}, Comparative evaluation of Dynomap and eleven baseline methods across three prediction tasks using circulating cell-free RNA profiles. Models were assessed on binary healthy versus cancer classification (\textbf{a}, \textbf{b}), multiclass cancer subtype prediction (\textbf{c}, \textbf{d}), and cancer stage classification (\textbf{e}, \textbf{f}). For each task, performance is reported under two feature regimes: highly variable genes (left panels) and a curated gene signature panel (right panels). Metrics include accuracy, macro-F1 score, recall, and specificity. Error bars indicate standard deviation across five cross-validation folds. Dynomap achieves the highest performance across most reported metrics in both feature settings.
\textbf{g}, Representative task-optimized image representations generated by Dynomap for subtype classification. Each group of four images corresponds to samples from a distinct cancer class. The images exhibit reproducible spatial patterns within classes, reflecting task-adaptive spatial organization learned directly from tabular gene expression inputs.}
\label{fig:fig2_rareseq_perf}
\end{figure}

Across all three tasks---healthy versus cancer discrimination, cancer subtype classification, and cancer stage classification---Dynomap consistently outperformed both classical machine learning methods and state-of-the-art deep tabular models. In the 4,000 HVG setting, Dynomap achieved 93.0\% accuracy for binary classification, with a macro-F1 of 92.0\% and macro-sensitivity of 91.1\%~(Fig.~\ref{fig:fig2_rareseq_perf}a). This exceeded strong classical baselines including Logistic Regression (90.3\%), XGBoost (91.2\%), and Random Forest (88.9\%), as well as modern neural tabular models such as TabM (92.9\%) and ModernNCA (90.7\%)~\cite{Chen2016,Breiman2001,Hollmann2022,Gorishniy2021}. While TabM achieved comparable performance to Dynomap in overall accuracy, Dynomap achieved much higher macro-sensitivity, indicating more balanced performance across classes. These results demonstrate that Dynomap surpasses both ensemble-based learners, widely regarded as strong baselines for tabular data~\cite{Hastie2009}, and recently proposed deep tabular architectures.

The advantage of Dynomap became more notable in the multiclass subtype task. Using 4,000 HVGs, Dynomap achieved 80.7\% accuracy with a macro-F1 of 79.3\%, substantially outperforming Logistic Regression (73.5\%), XGBoost (74.6\%), and TabM (78.6\%). The margin was even higher when compared to tree-based methods and MLPs. For state classification, Dynomap achieved 79.6\% accuracy and a macro-F1 of 78.1\%, again substantially exceeding classical and modern deep learning-based tabular approaches. Although performance margins were narrower for stage than subtype, Dynomap consistently maintained superior macro-sensitivity, indicating robustness under class imbalance. For the 622-gene signature panel in binary classification, Dynomap achieved 90.9\% accuracy with a macro-F1 of 90.0\%, remaining superior to existing models. Despite nearly a tenfold reduction in dimensionality, binary classification performance declined only modestly relative to the 4,000 HVG regime for Dynomap, indicating that the curated panel preserves the dominant discriminative signal.

For Subtype classification under the signature panel, Dynomap reached 92.0\% accuracy, which represents a significant increase relative to the 80.7\% accuracy obtained using 4,000 HVGs, suggesting that subtype-discriminative signal is concentrated within the curated gene set,. Even under this constrained feature space, Dynomap maintained higher macro-F1 compared with classical and deep learning-based baselines. For, stage prediction : using the 622-gene panel, Dynomap achieved 78.8\% accuracy, similar to the 4,000 HVG regime, but with reduced macro-F1 and macro-sensitivity. This asymmetry indicates that subtype information is concentrated within curated markers, whereas stage-associated signal appears more distributed across the broader transcriptome. Across both feature regimes, Dynomap remained or superior to all baselines, including TabPFN and TabM~\cite{Hollmann2022,Gorishniy2021}, demonstrating that its performance does not rely on brute-force dimensionality but adapts to the concentration of available signal.

Beyond predictive accuracy, Dynomap exhibited systematic changes in spatial organization and attribution patterns between the 4,000 HVG and 622-gene regimes. Visualization of generated images revealed reproducible task-specific spatial structures (Fig.~\ref{fig:fig2_rareseq_perf}g). Spatial autocorrelation analysis using Moran's I and neighborhood purity (kNN consistency) showed that observed spatial organization deviated from null layouts (Figs.~\ref{fig:figS10_rareseq_binary_spatial}, \ref{fig:figS11_rareseq_stage_spatial}), indicating non-random clustering of predictive signal.

\begin{figure}[pt]
\centering
\includegraphics[width=0.9\linewidth, page = 4]{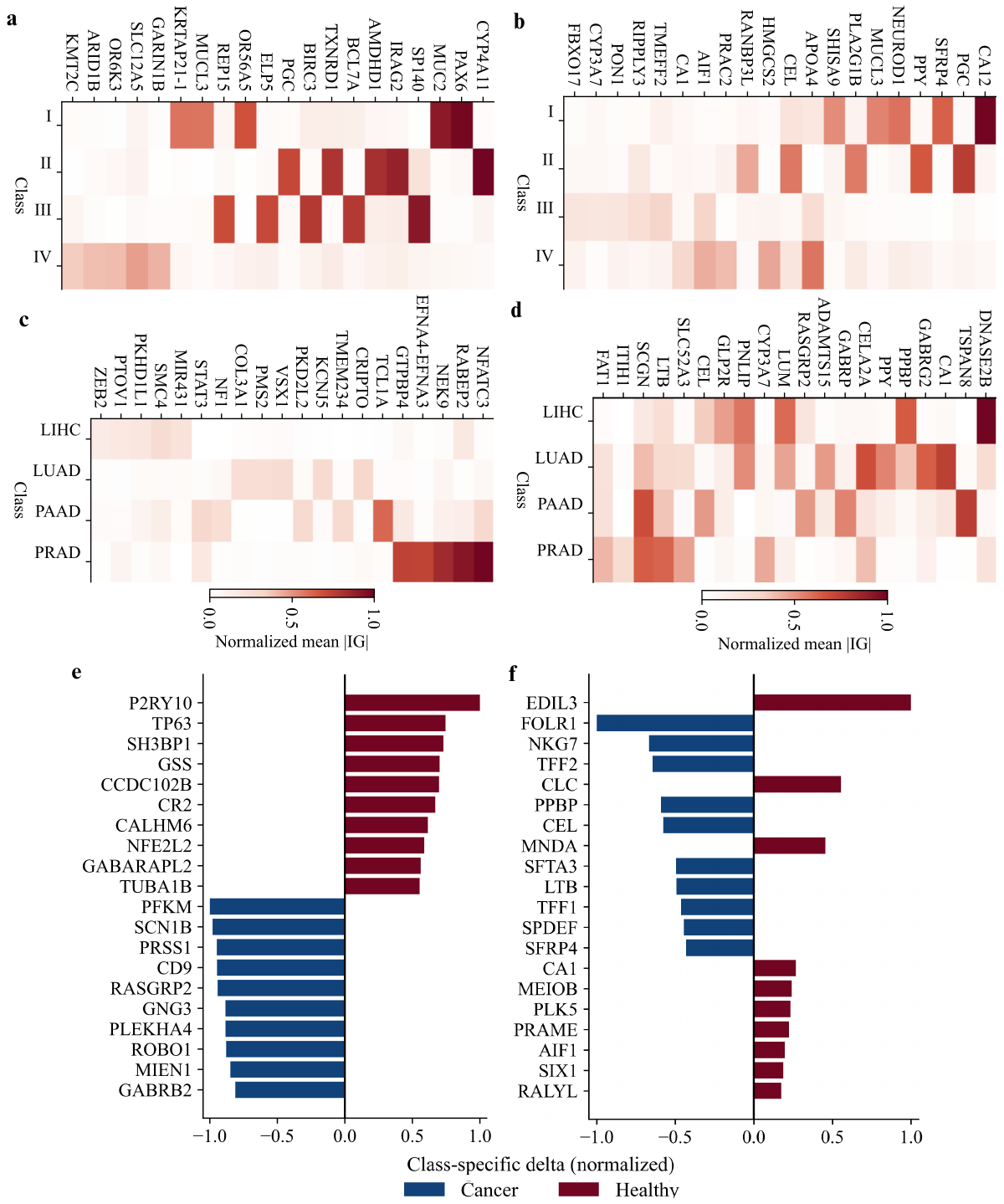}
\caption{\textbf{Task-dependent gene attribution patterns learned by Dynomap in liquid biopsy cfRNA.} 
\textbf{a}, Normalized mean absolute Integrated Gradients (|IG|) for cancer stage classification using highly variable genes. Rows correspond to stage classes (I–IV), and columns denote the highest-attribution genes. Distinct stages are associated with different gene attribution profiles. 
\textbf{b}, Stage classification using the 622-gene signature panel. Attribution patterns remain class-dependent. 
\textbf{c}, Multiclass subtype classification using highly variable genes. Each tumor type (LIHC, LUAD, PAAD, PRAD) is characterized by a distinct subset of high-attribution genes. 
\textbf{d}, Subtype classification under the signature panel setting, showing task-adaptive shifts in emphasized genes relative to the full-gene regime. 
\textbf{e}, Differential attribution for binary healthy versus cancer prediction using highly variable genes. Bars indicate normalized class-specific attribution differences (delta |IG|), with positive values favoring healthy prediction and negative values favoring cancer prediction. 
\textbf{f}, Differential attribution under the signature panel setting. Cancer and healthy-associated genes exhibit minimal overlap and form distinct attribution profiles. Together, these panels demonstrate that Dynomap learns task and class-dependent gene importance patterns that vary across prediction objectives and feature regimes.}
\label{fig:fig3_rareseq_interpret}
\end{figure}

To examine how predictive signals are distributed across genes, we computed Integrated Gradients (IG) and summarized normalized mean absolute attribution values for each task and feature regime (Fig.~\ref{fig:fig3_rareseq_interpret})~\cite{Sundararajan2017}. Attribution patterns were analyzed at the class level to identify genes contributing disproportionately to specific predictions. In subtype classification under the 4,000 HVG setting, distinct tumor classes were associated with separable high-attribution gene sets (Fig.~\ref{fig:figS21_rareseq_stage_ig}). For prostate cancer (PRAD), genes such as \textit{NFATC3}, \textit{RABEP2}, and \textit{BIRC3} exhibited elevated attribution relative to other classes. Lung adenocarcinoma (LUAD) predictions were associated with higher contributions from \textit{CRIPTO} and \textit{CA9}, whereas pancreatic cancer (PAAD) showed stronger attribution for \textit{STAT3} and \textit{NF1}~\cite{TCGA_PRAD_2015,TCGA_LUAD_2014,TCGA_PDAC_2017,Chen2021STAT3,CA9_NSCLC_2020,Ramakrishnan2022NF1}.

When restricting input to the 622-gene signature panel, class-dependent attribution structure remained evident, but the dominant genes shifted, and attribution mass was concentrated within a smaller subset of features. Although subtype separability improved under the signature regime, the specific high-contributing genes differed from those identified in the 4,000 HVG setting. In the binary healthy versus cancer task, butterfly plots revealed clear class-specific attribution asymmetry (Fig.~\ref{fig:figS19_rareseq_binary_ig}). Cancer predictions showed elevated attribution for genes including \textit{PFKM}, \textit{RASGRP2}, and \textit{SMAD4}, whereas healthy predictions were associated with higher attribution for \textit{P2RY10}, \textit{TP63}, and \textit{NFE2L2}~\cite{TCGA_PDAC_2017,Tsanov2025SMAD4,Bilodeau2021TP63,Panda2025NRF2}. Stage prediction exhibited partially overlapping but stage-dependent attribution patterns, with genes such as \textit{PAX6} and \textit{MUC2} more prominent in early-stage samples and \textit{BIRC3} and \textit{CTLA4} contributing more strongly in later stages (Fig.~\ref{fig:figS21_rareseq_stage_ig})~\cite{TCGA_PRAD_2015}. Under the 622-gene regime, stage-associated patterns persisted and used \textit{CA12}, \textit{PPY}, and \textit{TMEFF2}.

Across all tasks, the magnitude and distribution of IG attributions reorganized systematically when transitioning from 4,000 HVGs to the 622-gene signature panel. The full-gene regime distributed predictive contributions across a broader gene set, whereas the curated panel concentrated attribution within fewer features. No single subset of genes dominated across all tasks or regimes. Instead, both the spatial layout and attribution hierarchy adapted to the prediction objective and the available gene space, indicating that Dynomap dynamically reallocates predictive weight according to task and dimensional context.

Taken together, these results show that Dynomap surpasses strong classical learners and modern deep tabular models on ultrasensitive cfRNA data, while exhibiting task-dependent and feature-regime-dependent representational adaptation. Performance gains were accompanied by systematic changes in spatial organization and attribution concentration, indicating that Dynomap learns structured representations that reflect both the predictive objective and the available molecular feature space.

\subsection{Dynomap learns class-structured spatial representations in bulk transcriptomic profiling}
We next analyzed the GSE68086 platelet bulk RNA-seq dataset~\cite{GSE68086,best2015rna}, which included healthy controls and five cancer types profiled under controlled experimental conditions. Compared with cfRNA liquid biopsy, bulk transcriptomes generally provide stronger signal-to-noise ratios and clearer inter-class separation, making them well suited for assessing whether Dynomap’s learned spatial organization reflects stable class-specific transcriptional programs rather than noise-driven structure~\cite{Libbrecht2015,TCGA_PanCancer_2013}.

\begin{figure}[pth]
\centering
\includegraphics[width=0.9\linewidth, page = 6]{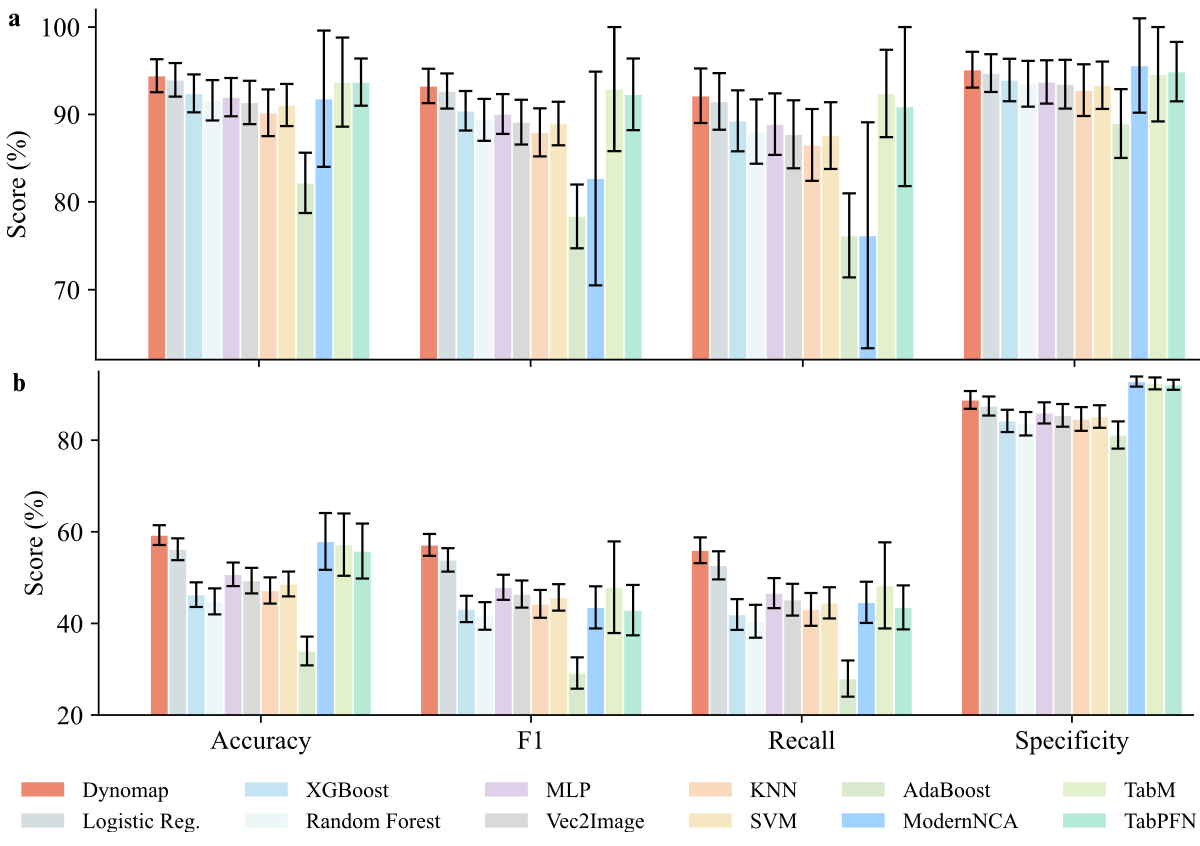}
\caption{\textbf{Benchmarking Dynomap on bulk transcriptomic classification (GSE68086).} 
\textbf{a}, Binary healthy versus cancer classification. Performance is reported across accuracy, macro-F1 score, recall, and specificity for Dynomap and eleven baseline methods. Dynomap achieves the highest performance across all reported metrics, with consistent gains in both sensitivity- and specificity-based measures. 
\textbf{b}, Multiclass tissue-of-origin classification. Despite increased class complexity, Dynomap maintains superior accuracy, macro-F1 score, and recall relative to conventional tabular baselines. Specificity remains high across methods, while Dynomap preserves balanced class discrimination. Error bars represent standard deviation across five cross-validation folds.}
\label{fig:fig5_gse_perf}
\end{figure}

Across both binary and multiclass tasks, Dynomap achieved the strongest performance among all evaluated models~(Fig.~\ref{fig:fig5_gse_perf}). In the binary healthy versus cancer setting, Dynomap reached 94.4\% accuracy with a macro-F1 of 93.3\% and macro-sensitivity of 92.1\%. This performance exceeded classical linear and tree-based baselines, including Logistic Regression (94.0\%) and XGBoost (92.4\%), and surpassed MLPs and Random Forests. Notably, Dynomap also outperformed modern deep tabular architectures, TabM (93.7\%) and TabPFN (93.7\%)~\cite{Gorishniy2021,Hollmann2022}. For binary classification, Dynomap consistently maintained higher macro-F1 and macro-sensitivity. The performance advantage became more prominent in the multiclass setting spanning five cancer types. Dynomap achieved 59.28\% accuracy with a macro-F1 of 57.14\%, exceeding Logistic Regression (56.19\%), TabM (57.2\%), and ModernNCA (57.9\% accuracy but with lower macro-F1), and substantially outperforming tree-based ensembles and MLPs. Importantly, Dynomap exhibited the highest macro-sensitivity (55.97\%) while maintaining strong macro-specificity (88.79\%), indicating that improvements were not confined to dominant classes. In contrast, several deep learning-based baselines displayed elevated specificity but reduced recall across minority classes. This pattern suggests that Dynomap captures structured differences among cancer types rather than exploiting imbalanced separability.

\begin{figure}[pt]
\centering
\includegraphics[width=0.9\linewidth, page = 5]{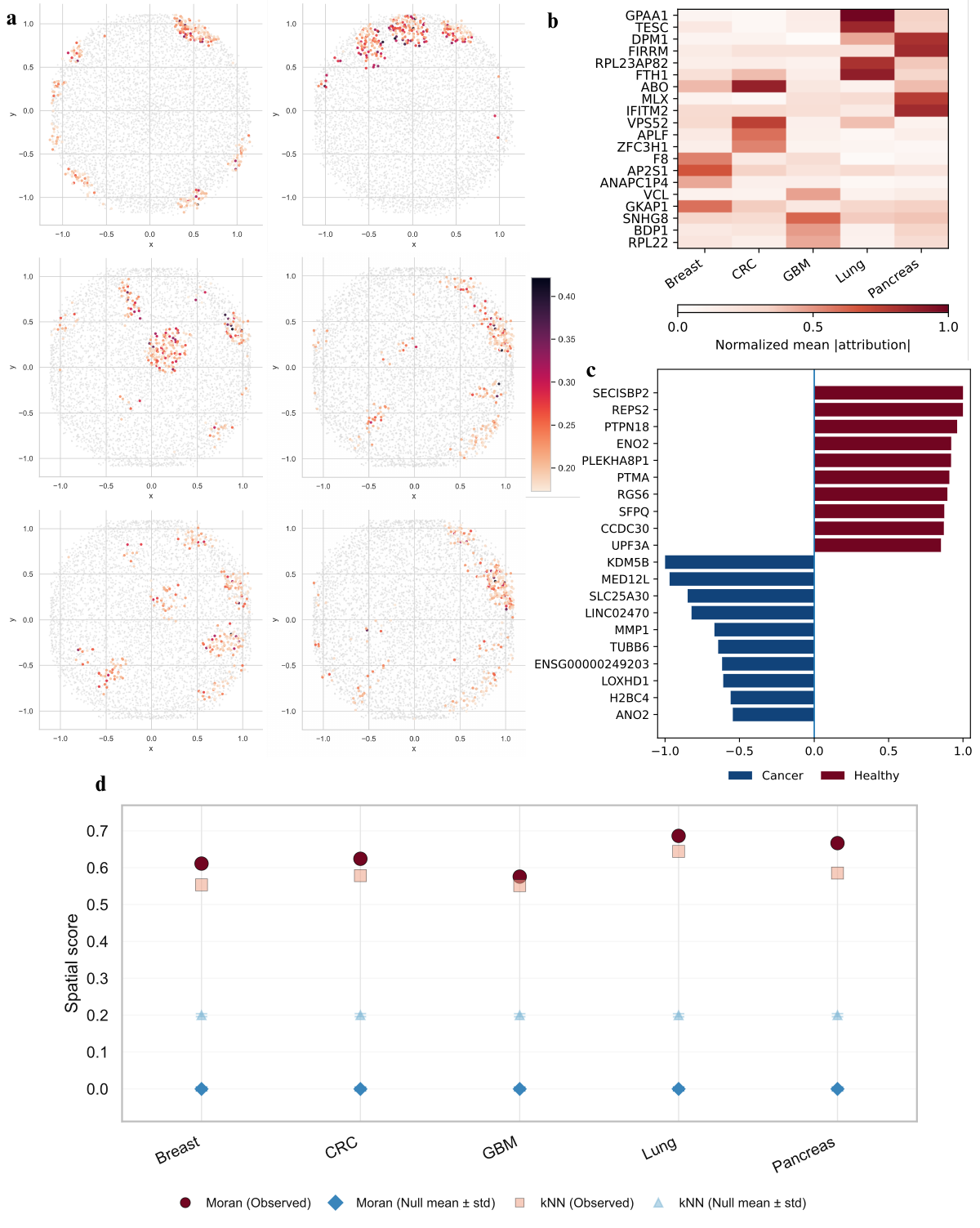}
\caption{\textbf{Class-specific spatial organization and attribution structure in bulk transcriptomic data (GSE68086).} 
\textbf{a}, Learned gene layouts with class-specific attribution overlaid. Gray points represent all genes positioned in the optimized two-dimensional layout. Colored points indicate genes with high Integrated Gradients attribution for the specified class. Top row: binary classification (Cancer, left; Healthy, right). Middle row: multiclass setting (Pancreas, left; Breast, right). Bottom row: multiclass setting (Glioblastoma, left; Colorectal cancer, right). Distinct classes activate spatially localized gene regions rather than overlapping uniformly across the map. 
\textbf{b}, Heatmap of normalized mean absolute attribution across cancer types for the highest-ranked genes. Rows correspond to genes and columns to tumor classes. Distinct attribution profiles are observed for different cancer types, with limited overlap among the most emphasized genes. 
\textbf{c}, Differential attribution for binary classification. Bars represent normalized class-specific attribution differences, with positive values favoring healthy prediction and negative values favoring cancer prediction. 
\textbf{d}, Quantification of spatial structure. Moran’s I (spatial autocorrelation) and kNN purity scores are shown for each cancer type. Observed values exceed null expectations derived from shuffled layouts, indicating that class-associated genes form statistically significant spatial neighborhoods in the learned representation.}
\label{fig:fig4_gse_interpret}
\end{figure}

Beyond predictive performance, Dynomap learned spatial layouts that exhibited clear and reproducible class-dependent organization (Fig.~\ref{fig:fig4_gse_interpret}a). Visualization of the learned embeddings revealed that genes emphasized for each cancer type occupied distinct spatial regions, forming localized clusters rather than diffuse or globally mixed configurations. These clusters were consistent across samples within the same class and differed systematically across classes, indicating that the learned layout preserves class-specific transcriptional structure. We then quantified this organization using global spatial autocorrelation (Moran’s I) and local neighborhood consistency (kNN purity~\cite{Moran1950,Anselin1995}) in Fig.~\ref{fig:fig4_gse_interpret}d. For all cancer classes, observed Moran’s I values substantially exceeded their corresponding null expectations, indicating that high-attribution genes were spatially clustered rather than randomly dispersed. Similarly, kNN purity scores were elevated across breast, colorectal, glioblastoma, lung, and pancreatic cancers, demonstrating that genes with similar attribution profiles formed coherent local neighborhoods in the learned layout. These findings establish that Dynomap’s spatial structure is statistically non-random and consistent across classes.

Integrated Gradients analysis further clarified how predictive signal is organized within this learned spatial manifold~\cite{Sundararajan2017}. In the binary task, cancer-associated predictions were characterized by elevated attribution for genes including \textit{KDM5B}, \textit{MED12L}, \textit{SLC25A30}, and \textit{MMP1}, which are connected to transcriptional regulation and cancer-linked signaling or metabolic programs, as well as extracellular matrix remodeling and invasion~\cite{Jose2020KDM5B,Gonzalez2022MED12,Wohlrab2022SLC25,Xu2022MMP1}, whereas healthy-associated predictions emphasized genes such as \textit{SECISBP2}, \textit{REPS2}, \textit{PTPN18}, and \textit{ENO2}, which have been linked to redox and selenoprotein regulation, receptor trafficking and signaling control, and lineage-associated marker expression in non-malignant contexts~\cite{Taguchi2021SECISBP2,Penninkhof2004REPS2,Wang2014PTPN18,BioSciRep2019NSE} (Fig.~\ref{fig:fig4_gse_interpret}c). Importantly, attribution patterns showed minimal overlap between cancer and healthy classes and were spatially localized to distinct regions of the learned map, indicating that class-discriminative signal is not only separable but also spatially segregated.

In the multiclass setting, each cancer type was associated with a distinct subset of high-attribution genes occupying separate spatial regions. Breast cancer predictions emphasized genes such as \textit{F8}, \textit{AP2S1}, and \textit{DOCK11}, whereas pancreatic cancer predictions were associated with elevated attribution for \textit{MMP1} and related extracellular remodeling genes~\cite{TCGA_BRCA_2012,TCGA_PDAC_2017}. Colorectal and glioblastoma cancers exhibited their own class-specific attribution signatures, including genes such as \textit{RHOC}, \textit{CAMTA1}, and \textit{RPL23A}, which have been reported in tumor-type--specific transcriptional analyses~\cite{TCGA_COAD_2012,TCGA_GBM_2008}. Although certain genes recur across cancer types, their attribution magnitude and spatial localization differed by class, producing non-overlapping attribution hotspots across the map. This pattern indicates class-dependent reweighting of predictive signal.

Notably, the spatial regions highlighted by Integrated Gradients aligned closely with areas of elevated Moran’s I and kNN purity. Genes with high attribution for a given cancer type were concentrated within contiguous neighborhoods, whereas low-attribution genes were more diffusely distributed. This concordance between attribution magnitude and spatial clustering suggests that Dynomap organizes transcriptional signals into coherent spatial modules that correspond to class-specific predictive structures. Moreover, genes emphasized in the multiclass task did not necessarily occupy identical spatial regions or exhibit identical attribution strengths in the binary task, demonstrating task-dependent reorganization of the layout. The learned spatial map is therefore not static, but adapts as class structure changes.

Dynomap achieves strong classification performance on bulk transcriptomic data while learning spatial representations in which class-specific predictive signal is organized into statistically quantifiable clusters. The combination of superior macro-balanced performance, elevated spatial autocorrelation, increased neighborhood consistency, and distinct class-dependent attribution patterns indicates that Dynomap captures structured transcriptional variation aligned with the underlying classification objective rather than relying on a fixed gene ordering or purely global separability.

\subsection{Dynomap discovers disease-specific features from phenomic voice measurements in a Parkinson’s dataset}
To test whether Dynomap performs well in other biomedical datasets beyond molecular profiling, we applied it to a phenomic Parkinson’s disease dataset derived from sustained voice recordings~\cite{Sakar2019}. The dataset contains 754 engineered acoustic features, including time-domain perturbation measures, entropy-based metrics, and tunable Q-factor wavelet transform (TQWT) coefficients. Unlike transcriptomic or cfRNA measurements, these variables are constructed descriptors of vocal dynamics and do not arise from an underlying biological network with a known relational structure. This example, therefore, demonstrates Dynomap’s ability to learn useful spatial organization directly from predictive signals in a high-dimensional engineered feature space.

\begin{figure}[pt]
\centering
\includegraphics[width=0.9\linewidth, page = 7]{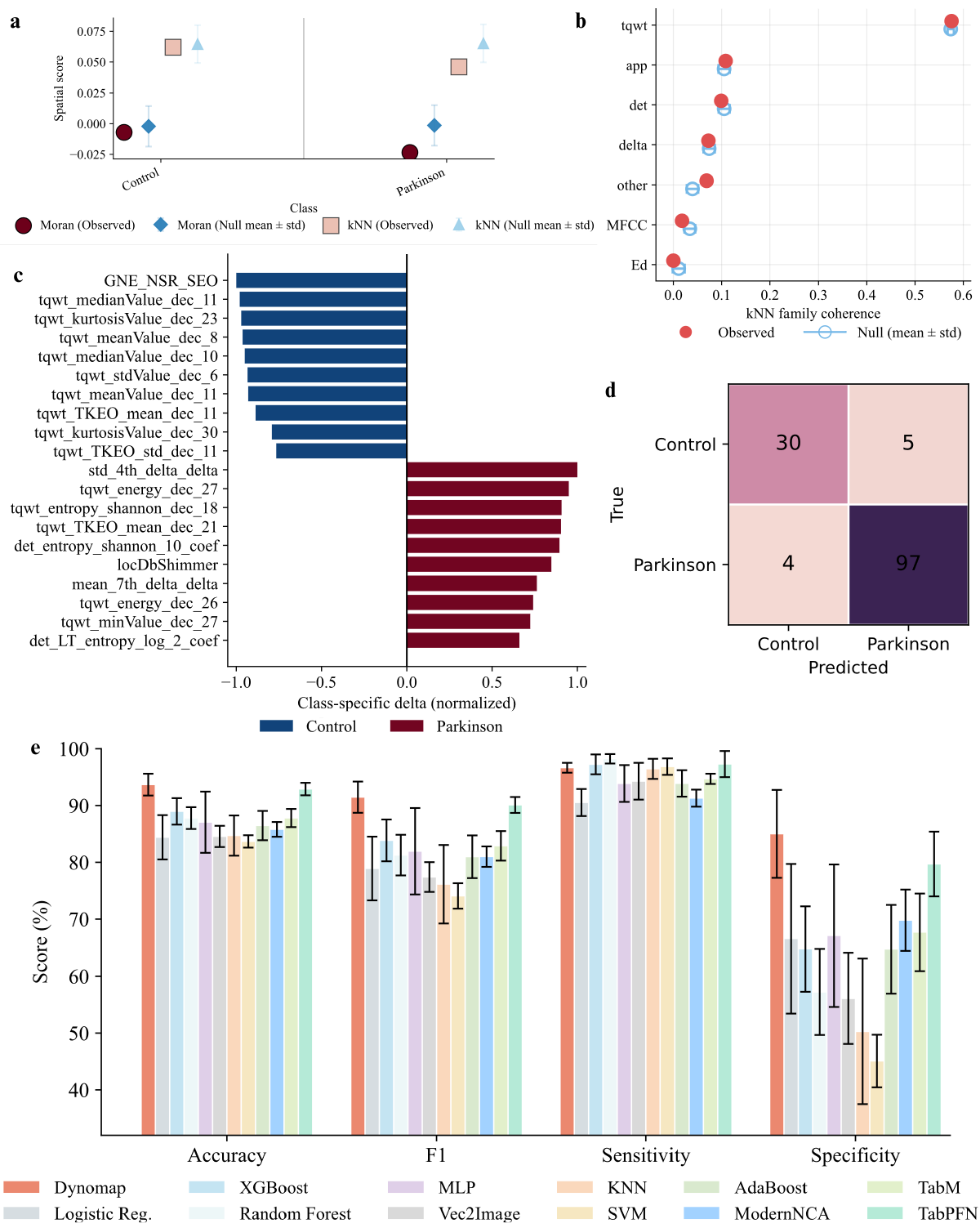}
\caption{\textbf{Spatial organization and phenomic feature attribution in Parkinson’s disease classification.} 
\textbf{a}, Quantification of spatial structure in the learned feature layout. Moran’s I (spatial autocorrelation) and kNN purity are shown for Control and Parkinson classes. Observed values are compared to null expectations derived from shuffled layouts, indicating non-random spatial clustering of class-associated features. 
\textbf{b}, kNN family coherence across engineered feature families (e.g., tqwt, MFCC, delta, and related descriptors). Observed coherence exceeds null distributions, indicating that features derived from similar extraction families occupy nearby spatial regions in the learned layout.
\textbf{c}, Differential feature attribution for binary classification. Bars represent normalized class-specific attribution differences (delta |IG|), with positive values favoring Parkinson prediction and negative values favoring Control prediction. Distinct sets of acoustic features contribute to each class. 
\textbf{d}, Confusion matrix showing counts of true and predicted labels for Control and Parkinson samples. 
\textbf{e}, Comparative performance across models. Accuracy, macro-F1 score, recall, and specificity are reported for Dynomap and eleven baseline methods. Dynomap achieves the strongest overall performance across class-balanced metrics. Error bars represent standard deviation across five cross-validation folds.}
\label{fig:fig6_parkinson}
\end{figure}

Dynomap achieved an accuracy of 93.7\% with a macro-F1 score of 91.5\% for discrimination between Parkinson’s disease and control samples (Fig.~\ref{fig:fig6_parkinson}). Macro-sensitivity reached 96.7\%, while macro-specificity was 85.0\%. Across all reported metrics, Dynomap substantially outperformed classical machine learning baselines, including Logistic Regression, Random Forest, XGBoost, and PCA+SVM, as well as modern neural tabular models such as TabM, ModernNCA, and TabPFN. Notably, several comparison methods achieved moderately high sensitivity but substantially lower specificity, indicating a bias toward predicting the majority class. Dynomap maintained strong sensitivity while improving specificity relative to most competing approaches, resulting in more balanced class discrimination. These differences are reflected in macro-F1 improvements of up to 7.6\% compared with existing methods.

Despite the absence of molecular relationships, the spatial layouts learned by Dynomap exhibited structured organization. Visualization of the feature embeddings showed that features contributing strongly to Parkinson’s disease and control predictions were not uniformly distributed across the learned image space (Fig.~\ref{fig:fig6_parkinson}a). Instead, high-attribution features formed localized regions. Quantitative spatial analysis confirmed this pattern. Observed Moran’s I values deviated from null expectations for both Parkinson- and control-associated features, indicating non-random clustering. Similarly, local neighborhood consistency measured by kNN purity exceeded null distributions (Fig.~\ref{fig:fig6_parkinson}a), demonstrating that features with similar attribution profiles occupied nearby spatial neighborhoods. These effects were reproducible across cross-validation folds, and fold-level layout similarity analysis showed stable spatial configurations across training splits (Supplementary Fig.~\ref{fig:figS33_parkinson_spatial_stats}).

Integrated Gradients analysis further revealed distinct class-specific attribution patterns (Fig.~\ref{fig:fig6_parkinson}b,c)~\cite{Sundararajan2017}. Parkinson’s disease predictions emphasized a subset of TQWT-derived spectral energy and entropy features, including multiple \textit{tqwt\_energy} and \textit{tqwt\_entropy\_shannon} coefficients, as well as perturbation-related metrics. In contrast, control predictions were associated with a different subset of features, including measures related to vocal stability and dispersion. These class-dependent features occupied partially segregated spatial regions in the learned layout, resulting in minimal overlap between Parkinson- and control-associated attribution hotspots. Many of the emphasized features correspond to acoustic descriptors previously used to characterize vocal impairment in Parkinson’s disease, including measures of frequency variation and nonlinear signal dynamics~\cite{Little2007,Tsanas2012,Sakar2019}. Dynomap was not provided with any grouping or domain-specific constraints on these features. Instead, the model learned a spatial arrangement that concentrated disease-associated signal into coherent neighborhoods aligned with prediction.

Importantly, the representational behavior observed here mirrors patterns seen in transcriptomic benchmarks, despite the fundamentally different feature semantics. In both molecular and phenomic settings, Dynomap learned layouts in which predictive features clustered spatially and reorganized according to the classification objective. This suggests that the spatial learning mechanism is not dependent on biological pathway structure but can operate on engineered acoustic measurements as well. While the Parkinson dataset lacks explicit biological networks, Dynomap nonetheless produced structured, class-dependent spatial organization and competitive predictive performance.

The analysis of the Parkinson dataset shows that even when applied to engineered acoustic descriptors with no explicit biological topology, the model organizes predictive features into coherent spatial regions and reallocates attribution in a class-dependent manner. The learned layouts remain stable across folds and reflect systematic clustering of disease-associated signal rather than random arrangement. These observations indicate that Dynomap’s representational behavior is driven by the predictive structure present in the data itself, rather than by assumptions about molecular organization. In this phenomic setting, Dynomap maintains balanced discrimination while preserving interpretable spatial structure, demonstrating that the framework extends naturally to high-dimensional engineered feature spaces.

\subsection{Dynomap achieves superior performance in analysis of a large TCGA-BRCA dataset and preserves PAM50 subtype structure in breast cancer}
Next, we evaluated Dynomap on TCGA-BRCA, a large-scale RNA-seq dataset with established PAM50 molecular subtype annotations~\cite{TCGA_BRCA_2012,Parker2009}. Expression profiles were obtained from the TCGA HiSeqV2 platform and aligned with curated clinical subtype labels. Five PAM50 classes were retained (LumA, LumB, Basal, HER2, and Normal-like), and a secondary binary task was defined by grouping Basal and HER2 tumors as aggressive relative to the remaining subtypes. LumA and LumB exhibit substantial transcriptional similarity, and class imbalance remains present despite the large cohort size~\cite{Pernas2018}. Unlike smaller benchmarks, TCGA-BRCA requires the model to maintain discriminative structure across thousands of genes and hundreds of samples without collapsing closely related subtypes.

\begin{figure}[pt]
\centering
\includegraphics[width=0.9\linewidth, page = 8]{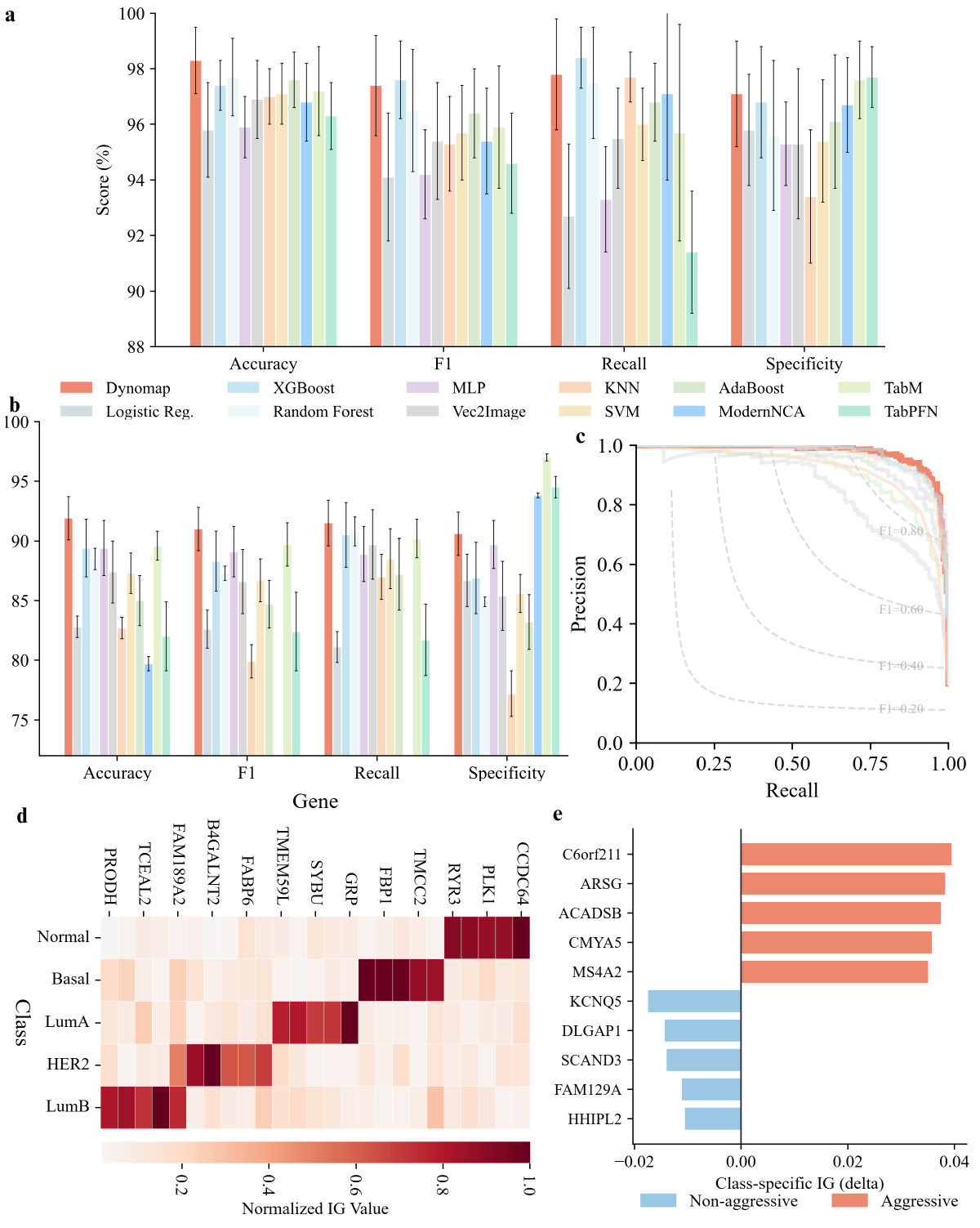}
\caption{\textbf{Binary and molecular subtype classification in breast cancer transcriptomes (TCGA-BRCA).} 
\textbf{a}, Comparative performance of Dynomap and eleven baseline methods for binary classification of non-aggressive versus aggressive tumors. Metrics include accuracy, macro-F1 score, recall, and specificity. Dynomap achieves the highest overall performance across class-balanced measures. Error bars indicate standard deviation across five cross-validation folds. 
\textbf{b}, Multiclass molecular subtype classification (Basal, HER2, LumA, LumB, Normal). Dynomap maintains superior accuracy and balanced performance across macro-F1, recall, and specificity relative to conventional and advanced tabular baselines. 
\textbf{c}, Precision-Recall curve for subtype classification. 
\textbf{d}, Normalized Integrated Gradients attribution for representative genes across molecular subtypes. Distinct gene groups are preferentially associated with individual subtypes. 
\textbf{e}, Differential attribution for binary stratification into non-aggressive and aggressive tumors. Bars represent class-specific attribution differences (delta |IG|), with positive values indicating greater contribution to aggressive prediction and negative values to non-aggressive prediction.}
\label{fig:fig7_tcga_brca}
\end{figure}

As seen in Fig.~\ref{fig:fig7_tcga_brca}a, for the binary task distinguishing aggressive from non-aggressive tumors, Dynomap achieved \textbf{98.3\%} accuracy with a macro-F1 of \textbf{97.4\%}, recall of \textbf{97.8\%}, and specificity of \textbf{97.1\%}. This represents the highest overall accuracy among all evaluated models and a balanced tradeoff between sensitivity and specificity. Tree-based ensembles such as XGBoost achieved slightly higher recall (98.4\%) and comparable F1 (97.6\%), while TabM exhibited marginally higher specificity (97.6\%). However, these methods did not simultaneously maximize accuracy and maintain comparable balance across both classes. Precision--recall and ROC curves further show that Dynomap maintains high precision across a broad recall range rather than achieving performance through a narrow high-confidence regime (Supplementary Fig.~\ref{fig:figS29_tcga_binary_performance}). These results indicate that Dynomap remains competitive with strong ensemble methods while preserving stable class discrimination at scale.

The multiclass PAM50 subtype task presents a more demanding benchmark in Fig.~\ref{fig:fig7_tcga_brca}b. Dynomap achieved \textbf{91.9\%} accuracy and a macro-F1 of \textbf{91.0\%}, exceeding the strongest baseline (TabM, 89.6\% accuracy) and outperforming XGBoost and MLP across macro-balanced metrics. Notably, improvements were most pronounced in macro-F1 and recall, indicating that performance gains were not driven by dominant classes but reflected improved discrimination across all subtypes. While TabM achieved higher macro-specificity, it did so with lower recall, suggesting a more conservative classification boundary. Precision--recall curves show that Dynomap maintains consistently higher precision across recall thresholds (Fig.~\ref{fig:fig7_tcga_brca}c, Supplementary Fig.~\ref{fig:figS30_tcga_subtype_performance}). These results demonstrate that Dynomap preserves subtype separability even when transcriptional profiles exhibit substantial inter-class similarity.

Integrated Gradients analysis further revealed subtype-specific redistribution of gene attribution (Fig.~\ref{fig:fig7_tcga_brca}d). Distinct PAM50 classes were characterized by different high-attribution gene sets with limited overlap among top-ranked features. For example, LumB predictions showed elevated attribution for \textit{PRODH}, \textit{TCEAL2}, and \textit{FAM189A2}, whereas HER2-associated predictions emphasized \textit{B4GALNT2} and \textit{FABP6}. LumA exhibited stronger attribution for \textit{TMEM59L}, \textit{SYBU}, and \textit{GRP}, while Basal predictions highlighted genes including \textit{FBP1} and \textit{TMCC2}. Normal-like samples emphasized a separate subset, including \textit{RYR3}, \textit{PLK1}, and \textit{CCDC64}. These patterns indicate that Dynomap does not rely on a single global gene ranking across classes. Instead, attribution magnitude and spatial localization adapt to the multiclass structure of the task.

A similar redistribution was observed in the binary grouping. Aggressive cancer predictions were associated with genes such as \textit{C6orf211} (\textit{ARMT1}), \textit{ARSG}, \textit{ACADSB}, \textit{CMYA5}, and \textit{MS4A2}, whereas non-aggressive predictions emphasized \textit{KCNQ5}, \textit{DLGAP1}, \textit{SCAND3}, \textit{FAM129A} (\textit{NIBAN1}), and \textit{HHIPL2}. Several of these genes are linked to distinct biological programs, including immune receptor signaling (\textit{MS4A2})~\cite{NCBI_MS4A2_2206}, mitochondrial metabolism (\textit{ACADSB})~\cite{NCBI_ACADSB_36}, and PCNA-associated regulation (\textit{ARMT1})~\cite{Armt1_PCNA_CellReports2015}, as well as neuronal or synaptic gene programs (\textit{KCNQ5}, \textit{DLGAP1})~\cite{NCBI_KCNQ5_56479,NCBI_DLGAP1_9229} and stress-response or proliferative regulation (\textit{FAM129A}, \textit{SCAND3}, \textit{HHIPL2})~\cite{NIBAN1_Review_Frontiers2022,NCBI_SCAND3_114821,HHIPL2_NSCLC_CellDeathDis2025}. These class-specific attribution profiles were spatially localized and showed minimal overlap in high-contribution regions. Importantly, genes emphasized in the binary task did not necessarily occupy identical spatial positions or maintain identical attribution magnitudes in the multiclass setting, indicating task-dependent reorganization of predictive weight.

Overall, the TCGA-BRCA analysis demonstrates that Dynomap scales to large, high-dimensional transcriptomic cohorts while preserving clinically defined subtype structure. The model achieves strong balanced performance in both binary and multiclass settings and learns stable, non-random spatial layouts in which subtype-associated genes form coherent modules. Rather than enforcing a static feature hierarchy, Dynomap reallocates attribution across genes according to task structure and class definition, maintaining interpretability and discriminative performance under substantial heterogeneity.

\subsection{Dynomap preserves lineage-specific transcriptional structure in large-scale single-cell classification}
Finally, we assessed Dynomap on the Tabula Muris single-cell transcriptomic atlas~\cite{TabulaMuris2018}, comprising 54{,}865 cells annotated into 55 cell types across 20 mouse organs. This benchmark differs from bulk or liquid biopsy settings because it involves fine-grained cellular identities embedded within overlapping lineage hierarchies, with substantial within-class variability and closely related immune and stromal populations. The task therefore probes whether Dynomap can maintain macro-balanced performance and learn structured, lineage-aligned organization when class structure is high-cardinality and partially hierarchical.

\begin{figure}[pt]
\centering
\includegraphics[width=0.95\linewidth, page = 9]{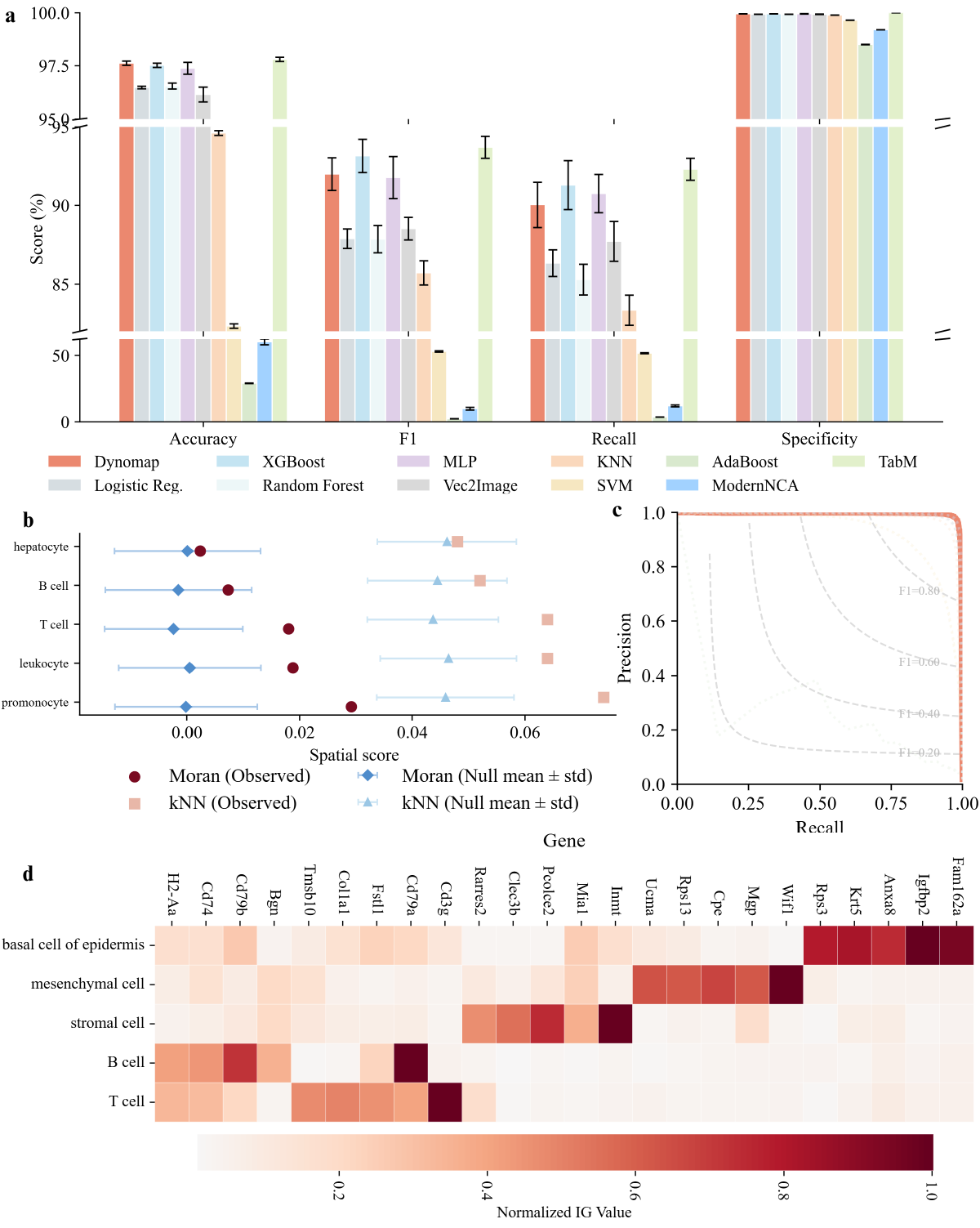}
\caption{\textbf{Cell-type classification and spatial organization in single-cell transcriptomes (Tabula Muris).} 
\textbf{a}, Comparative performance of Dynomap and eleven baseline methods for multiclass cell-type classification. Metrics include accuracy, macro-F1 score, recall, and macro-specificity. Dynomap achieves the highest overall performance across class-balanced measures. Error bars indicate standard deviation across cross-validation folds. 
\textbf{b}, Spatial organization analysis for representative cell types. Moran’s I (spatial autocorrelation) and kNN purity are shown alongside null expectations derived from shuffled layouts. Observed values exceed null distributions, indicating non-random clustering of cell-type–associated genes in the learned layout. 
\textbf{c}, Precision–recall curve demonstrating stable performance across recall thresholds. 
\textbf{d}, Normalized Integrated Gradients attribution for representative genes across selected cell types. Distinct gene groups are preferentially associated with individual cellular identities, indicating class-dependent attribution structure within the learned spatial representation.}
\label{fig:fig8_tabula_muris}
\end{figure}

Across this multiclass setting, Dynomap achieved 97.6\% accuracy with a macro-F1 score of 92.0\%, a macro-sensitivity of 90.0\%, and a macro-specificity of 100.0\% (Fig.~\ref{fig:fig8_tabula_muris}a). While overall accuracy approached the ceiling for several high-capacity models, macro-balanced metrics differentiated performance under extensive class heterogeneity. Dynomap maintained higher macro-F1 than logistic regression (87.9\%), random forest (87.9\%), and KNN (85.7\%), and substantially outperformed PCA+SVM and AdaBoost. Tree-based ensemble methods achieved comparable overall accuracy, with XGBoost reaching 97.5\% accuracy and a macro-F1 of 93.2\%, and TabM achieving 97.8\% accuracy with a macro-F1 of 93.7\%. Despite these narrow margins among the strongest ensemble-based approaches, Dynomap preserved competitive macro-balanced performance while maintaining near-perfect specificity, indicating stable discrimination across a large number of closely related cell identities. Precision--recall analysis further demonstrated that Dynomap sustained high precision across a broad recall range rather than concentrating performance on a limited subset of highly separable cell populations (Fig.~\ref{fig:fig8_tabula_muris}c, Supplementary Fig.~\ref{fig:figS22_tm_pr_roc}).

Given the high number of distinct classes, we next examined whether the learned representation exhibits structured spatial organization rather than diffuse attribution. Spatial autocorrelation analysis revealed that observed Moran’s I values consistently exceeded null expectations derived from shuffled layouts across representative cell types (Fig.~\ref{fig:fig8_tabula_muris}b; Supplementary Fig.~\ref{fig:figS23_tm_spatial_stats})~\cite{Moran1950,Anselin1995}. Similarly, kNN purity scores were high relative to null distributions, indicating that genes contributing strongly to a given cell type formed coherent local neighborhoods in the learned layout.

Integrated Gradients analysis provided a global view of class-specific attribution across cell types (Fig.~\ref{fig:fig8_tabula_muris}d; Supplementary Fig.~\ref{fig:figS27_tm_heatmap})~\cite{Sundararajan2017}. Instead of showing a uniform importance pattern, high-attribution genes appeared in lineage-restricted bands with minimal cross-class overlap. T-cell--associated populations had elevated attribution for \textit{Cd3g}, a core component of the T-cell receptor complex required for antigen recognition and signal transduction~\cite{Love2012TCR}. B-cell--related populations emphasized \textit{Cd79b} and \textit{Cd74}, central mediators of B-cell receptor signaling and MHC class II--dependent antigen presentation~\cite{Reth1992BCR,Neefjes2011MHCII}. Stromal and mesenchymal populations preferentially highlighted extracellular matrix genes such as \textit{Col1a1} and \textit{Col3a1}, which encode fibrillar collagens characteristic of connective tissue and fibroblast identity~\cite{RicardBlum2011Collagen}. Epithelial subtypes exhibited stronger attribution for keratin-associated genes including \textit{Krt5} and \textit{Krt15}, established markers of epithelial differentiation programs~\cite{Moll2008Keratin}. These lineage-aligned attribution modules are consistent with coordinated gene programs underlying cell identity in single-cell atlases~\cite{TabulaMuris2018,Zheng2017}.

Importantly, high-attribution genes were not globally dominant across multiple classes. Pairwise similarity analysis of attribution profiles showed strong diagonal dominance with limited off-diagonal overlap (Supplementary Fig.~\ref{fig:figS24_tm_jaccard}), indicating class-specific redistribution of predictive weight. Hierarchical clustering of cell-type attribution profiles further revealed biologically coherent groupings, with closely related immune populations forming neighboring clusters and stromal and epithelial lineages segregating into distinct branches (Supplementary Figs.~\ref{fig:figS25_tm_class_dendrogram}, \ref{fig:figS26_tm_gene_dendrogram}). Attribution profiles were lineage-specific, with limited overlap across cell types.

In summary, these results show that Dynomap scales to high-dimensional single-cell data with extensive class cardinality while preserving structured spatial organization and lineage-specific attribution patterns. In a setting characterized by subtle transcriptional boundaries and hierarchical relationships among 55 cell types, the model maintains strong macro-balanced performance and learns non-random spatial layouts in which predictive signal concentrates into biologically coherent modules. Rather than relying on a static ranking of genes, Dynomap dynamically redistributes attribution according to cell identity, indicating that its representational behavior remains structured and task-adaptive under large-scale single-cell heterogeneity.

\section{Discussion}
Tabular biomedical data are inherently unstructured, where features appear as independent dimensions, and any relationship among them must be inferred entirely by the learning algorithm for decision-making~\cite{Gorishniy2021,Grinsztajn2022}. In contrast, vision deep learning models exploit locality, weight sharing, and hierarchical aggregation, assumptions that are naturally aligned with image and signal domains~\cite{LeCun1998,LeCun2015,Goodfellow2016}. When applied directly to unordered tabular inputs, this row-column representation is not meaningfully grounded in feature relationships~\cite{Gorishniy2021}. Dynomap addresses this representational gap by learning spatial organization jointly with prediction (Fig.~\ref{fig:fig1_pipeline} and Fig.~\ref{fig:figS10_Dynomap_Pipeline}). Feature coordinates are optimized through a differentiable rendering process that is coupled directly to the prediction objective, rather than imposed through predefined similarity measures or external grouping rules. Across all datasets examined here, there are two consistent observations. First, Dynomap achieves strong predictive performance across regimes that vary in feature dimensionality, class imbalance, biological heterogeneity, and data modality~(Figs.~\ref{fig:fig2_rareseq_perf}--\ref{fig:fig8_tabula_muris}). Second, the learned spatial representations exhibit structured organization in which predictive features self-organize into coherent, task-dependent neighborhoods that are statistically distinguishable from null spatial arrangements (Figs.~\ref{fig:fig3_rareseq_interpret}, \ref{fig:fig4_gse_interpret}, \ref{fig:fig6_parkinson}, \ref{fig:fig7_tcga_brca}, \ref{fig:fig8_tabula_muris}; Supplementary Figs.~\ref{fig:figS10_rareseq_binary_spatial}--\ref{fig:figS12_rareseq_subtype_spatial}, \ref{fig:figS23_tm_spatial_stats}, \ref{fig:figS32_tcga_spatial_stats}). Taken together, these findings support the view that end-to-end spatial representation learning can extend convolutional inductive biases to traditionally non-spatial tabular domains while preserving interpretability.

A central observation across experiments is that Dynomap’s performance advantage is not dependent on a narrow dimensional regime. In RARE-Seq, strong performance is maintained under both full-gene and restricted signature-panel settings, and across binary detection, stage prediction, and subtype classification (Fig.~\ref{fig:fig2_rareseq_perf}). When the number of highly variable genes is systematically varied over a broad range, performance remains stable relative to competitive baselines (Supplementary Fig.~\ref{fig:figS18_rareseq_hvg_sweep}). In high-dimensional biomedical settings, predictive signal is often distributed and weak, and model performance can depend strongly on feature filtering and regularization strategies~\cite{Hastie2009,Grinsztajn2022}. The observed stability indicates that Dynomap does not rely on a single curated subset of features but adapts its spatial organization as the feature space expands or contracts. Similar robustness is observed in TCGA-BRCA, where subtype prediction under class imbalance remains competitive when evaluated using precision--recall curves, a metric better suited to imbalanced classification than accuracy or ROC analysis~\cite{Saito2015} (Fig.~\ref{fig:fig7_tcga_brca}; Supplementary Figs.~\ref{fig:figS29_tcga_binary_performance}, \ref{fig:figS30_tcga_subtype_performance}). In Tabula Muris, Dynomap maintains competitive performance despite a large number of closely related cell types with overlapping transcriptional programs~\cite{TabulaMuris2018} (Fig.~\ref{fig:fig8_tabula_muris}). Across these regimes, the representation reorganizes with the task rather than depending on a fixed feature ordering, supporting generalization across heterogeneous biomedical contexts.

Interpretability analyses further demonstrate that the learned spatial organization reflects structured predictive signal. Integrated Gradients attribution maps reveal localized regions that differ across objectives within the same dataset (Figs.~\ref{fig:fig3_rareseq_interpret} and \ref{fig:fig7_tcga_brca}), consistent with the principle that meaningful interpretation requires alignment between model structure and task-specific feature interactions~\cite{Rudin2019}. In RARE-Seq, shifting from binary detection to stage or subtype prediction reorganizes high-attribution genes into distinct spatial neighborhoods despite identical underlying features (Fig.~\ref{fig:fig3_rareseq_interpret}; Supplementary Figs.~\ref{fig:figS10_rareseq_binary_spatial}--\ref{fig:figS12_rareseq_subtype_spatial}). Spatial statistics provide quantitative support for this structure. Moran’s I~\cite{Moran1950} and kNN-based neighborhood coherence consistently exceed shuffled-layout baselines across bulk transcriptomic, single-cell, and phenomic datasets (Fig.~\ref{fig:fig4_gse_interpret}d; Fig.~\ref{fig:fig6_parkinson}c; Supplementary Figs.~\ref{fig:figS23_tm_spatial_stats}, \ref{fig:figS32_tcga_spatial_stats}). These results indicate that clustering of predictive features is unlikely to arise from arbitrary coordinate assignment. In Tabula Muris, the Jaccard similarity matrix shows strong within-class neighborhood stability and structured overlap among related lineages (Supplementary Fig.~\ref{fig:figS24_tm_jaccard}), consistent with shared developmental and transcriptional programs across immune and stromal populations~\cite{TabulaMuris2018}. High-attribution genes include canonical lineage markers that occupy coherent spatial regions (Fig.~\ref{fig:fig8_tabula_muris}d; Supplementary Fig.~\ref{fig:figS27_tm_heatmap}), supporting the biological plausibility of the learned layouts.

The emergence of coherent spatial neighborhoods follows directly from the optimization framework. Convolutional filters aggregate information locally and are most effective when spatial proximity corresponds to meaningful relationships~\cite{LeCun1998,LeCun2015,Bengio2013}. When tabular features are arranged without structure, local aggregation combines unrelated variables, limiting efficiency. By learning feature coordinates jointly with prediction, Dynomap aligns spatial proximity with predictive co-activation patterns. This alignment allows convolutional layers to capture distributed interactions through localized receptive fields while preserving feature identity. The repeated observation of task-dependent spatial reorganization across independent datasets suggests that the learned layouts reflect objective-driven structure rather than arbitrary embeddings.

In summary, the proposed approach, Dynomap, demonstrates that learning spatial organization directly from the prediction objective provides a general strategy for transforming heterogeneous tabular biomedical data into structured and interpretable representations. Across tasks that vary in dimensionality, imbalance, and biological complexity, the framework produces stable spatial layouts in which predictive features form coherent, task-adaptive neighborhoods associated with disease state, cellular identity, or phenotypic condition. In multiple benchmarks, Dynomap substantially improved predictive accuracy relative to the strongest baseline. Dynomap also maintained performance when the feature regime was restricted from \textbf{full feature set} to \textbf{signature panel}, indicating that the learned topography adapts to the available biological signal. By providing an explicit, task-specific feature organization through a differentiable mapping, Dynomap enables attribution-guided inspection and hypothesis generation in settings where feature interactions are otherwise implicit. These results indicate that spatial representation learning can extend vision-based modeling approaches to traditionally non-spatial tabular domains while preserving interpretability and enabling the discovery of structured biological signals across diverse biomedical settings.

\section{Methods}

\subsection{Overview of the Dynomap framework}

Dynomap converts high-dimensional tabular inputs into structured spatial representations that can be processed by convolutional models. The framework assigns each feature a learnable coordinate in a two-dimensional space and renders each sample into an image using a differentiable kernel-based mapping. Unlike fixed vector-to-image transformations, both spatial coordinates and rendering parameters are optimized jointly with the prediction objective. Let $\mathbf{x} \in \mathbb{R}^G$ denote an input feature vector with $G$ features. The model learns a mapping from $\mathbf{x}$ to a spatial image $I \in \mathbb{R}^{P \times P}$, while simultaneously optimizing a classifier defined on the rendered representation.

\subsection{Feature gating}

Before spatial organization, feature magnitudes are adaptively reweighted to allow task-dependent emphasis or suppression. Given an input vector $\mathbf{x}$, a dense layer with sigmoid activation produces gating weights $\mathbf{g} \in (0,1)^G$,
\[
\mathbf{g} = \sigma(\mathbf{W}_g \mathbf{x} + \mathbf{b}_g),
\]
where $\sigma$ denotes the sigmoid function. The gated feature vector is
\[
\tilde{\mathbf{x}} = \mathbf{x} \odot \mathbf{g}.
\]
This operation provides a differentiable mechanism for feature modulation prior to spatial rendering, allowing the representation to adapt to the prediction objective.

\subsection{Learnable spatial layout}

To introduce structure into unordered features, Dynomap assigns each feature a trainable coordinate in a two-dimensional space. For each feature $i$, a coordinate $\mathbf{c}_i \in \mathbb{R}^2$ is learned. Coordinates are initialized uniformly in $[-1,1]$ and optimized jointly with model parameters. To prevent degenerate configurations, spatial dispersion is enforced through three regularization terms.

A centering term ensures zero-mean layout,
\[
\mathcal{L}_{\text{center}} = \left\| \frac{1}{G} \sum_{i=1}^{G} \mathbf{c}_i \right\|_2^2.
\]

A spread term constrains global variance,
\[
\mathcal{L}_{\text{spread}} = \left( \mathrm{std}(\{\mathbf{c}_i\}) - 1 \right)^2.
\]

A repulsion term discourages coordinate collapse.
\[
\mathcal{L}_{\text{repel}} = \mathbb{E}_{i \neq j} \left[ \frac{1}{\|\mathbf{c}_i - \mathbf{c}_j\|_2^2 + \epsilon} \right],
\]
with $\epsilon = 10^{-4}$.

The total layout regularization is
\[
\mathcal{L}_{\text{layout}} = 0.01 \mathcal{L}_{\text{center}} + 0.01 \mathcal{L}_{\text{spread}} + 0.05 \mathcal{L}_{\text{repel}}.
\]
These terms maintain spatial dispersion while allowing coordinates to reorganize according to predictive structure.

\subsection{Differentiable rendering}
Once spatial coordinates are defined, each sample is rendered into a two-dimensional image of resolution $P \times P$, where $P$ denotes the number of pixels per spatial dimension and is treated as a model hyperparameter. This step converts feature values into spatial intensity patterns while preserving differentiability. For each grid location $(u,v)$ in the $P \times P$ image,

\[
I(u,v) = \sum_{i=1}^{G} \tilde{x}_i 
\exp\left( - \frac{\| (u,v) - \mathbf{c}_i \|_2^2}{2 \sigma_i^2} \right).
\]

Kernel width is feature-dependent,
\[
\sigma_i = 0.5 + 4.5 \tanh(|\tilde{x}_i|).
\]

The rendering grid is dynamically defined based on the coordinate bounding box with fixed padding. Each rendered image is standardized per sample. Because the rendering operation is differentiable with respect to both $\tilde{\mathbf{x}}$ and $\mathbf{c}_i$, gradients propagate through the spatial mapping during optimization.

\subsection{End-to-end joint optimization}

All components of Dynomap, including feature gating, spatial coordinates, rendering, and classification layers, are optimized jointly under a unified objective. Let $f(\mathbf{x}; \theta)$ denote the full model with parameters $\theta = \{\theta_{\text{gating}}, \theta_{\text{layout}}, \theta_{\text{CNN}}, \theta_{\text{head}}\}$. The total loss is defined as
\[
\mathcal{L}_{\text{total}} = \mathcal{L}_{\text{CE}}(f(\mathbf{x}; \theta), y) + \mathcal{L}_{\text{layout}},
\]
where $\mathcal{L}_{\text{CE}}$ denotes categorical cross-entropy and $\mathcal{L}_{\text{layout}}$ represents the spatial regularization terms described above. Model parameters are updated via gradient-based optimization,
\[
\theta_{t+1} = \theta_t - \eta \nabla_{\theta} \mathcal{L}_{\text{total}},
\]
where $\eta$ denotes the learning rate.

Because the rendering operation is differentiable with respect to both feature values and spatial coordinates, gradients propagate from the classification loss through the convolutional layers and rendering module to the coordinate parameters. As a result, spatial organization evolves directly in response to predictive error, allowing feature neighborhoods to adapt to the task objective.

\subsection{Layout stabilization}

During early training, spatial coordinates may undergo substantial movement as the model searches for predictive organization. To prevent continuous drift and to stabilize topology, coordinate velocity is monitored across epochs. Mean displacement is defined as
\[
v^{(t)} = \frac{1}{G} \sum_{i=1}^{G} \left\| \mathbf{c}_i^{(t)} - \mathbf{c}_i^{(t-1)} \right\|_2.
\]
If $v^{(t)}$ falls below a predefined threshold for a fixed number of consecutive epochs, coordinate updates are halted and the layout is frozen. Model selection is performed only after stabilization to ensure evaluation under a fixed spatial configuration.

\subsection{Convolutional modeling and training}
Rendered images of resolution $P \times P$ with $P=64$ are processed using a CNN designed to capture hierarchical spatial structure through shared local filters~\cite{LeCun1998,LeCun2015}. The backbone consists of sequential two-dimensional convolutional layers with ReLU activation~\cite{Nair2010}, applied with same padding to preserve spatial dimensions. Max pooling layers reduce spatial resolution, and global pooling aggregates spatial responses into a compact feature vector.

To preserve direct access to original tabular information, the convolutional feature vector is concatenated with the gated feature vector $\tilde{\mathbf{x}}$. The combined representation is passed through a fully connected classifier head consisting of a dropout layer with rate 0.5~\cite{Srivastava2014}, a hidden dense layer with 64 units and ReLU activation, and a final linear layer producing class logits.

The model is trained by minimizing categorical cross-entropy loss with label smoothing~\cite{Szegedy2016}. Optimization is performed using the Adam optimizer~\cite{Kingma2014} with learning rate $1\times10^{-3}$. Mini-batch size is set to 16 and models are trained for up to 1000 epochs. During training, spatial coordinates are monitored for stabilization as described above (patience = 15 epochs, velocity threshold = 0.002).

All models were implemented in TensorFlow with reproducible random seed initialization across Python, NumPy, and TensorFlow operations. GPU acceleration was used during training. Identical hyperparameters were applied across all datasets to ensure comparability.

\subsection{Integrated Gradients attribution}
To quantify feature contributions to model predictions, we employed Integrated Gradients (IG)~\cite{Sundararajan2017}. Let $f(\mathbf{x})$ denote the logit corresponding to a target class and $\mathbf{x} \in \mathbb{R}^G$ the input feature vector. Given a baseline input $\mathbf{x}'$ (set to the zero vector), the integrated gradient for feature $i$ is defined as

\[
\mathrm{IG}_i(\mathbf{x}) = (x_i - x'_i) \int_{0}^{1}
\frac{\partial f\left(\mathbf{x}' + \alpha(\mathbf{x} - \mathbf{x}')\right)}
{\partial x_i} \, d\alpha.
\]

This formulation accumulates gradients along the straight-line path from the baseline to the input. In practice, the integral is approximated using a Riemann sum with $m$ discrete steps. For multiclass settings, IG was computed with respect to the logit of the predicted class. Attribution magnitudes were aggregated across samples within each class to obtain class-specific importance profiles. These attribution vectors were subsequently used for spatial and clustering analyses.

\subsection{Spatial autocorrelation using Moran’s I}
To quantify whether high-attribution features cluster spatially beyond random expectation, we computed Moran’s I statistic~\cite{Moran1950}. Let $a_i$ denote the attribution score for feature $i$ and let $\mathbf{c}_i$ denote its learned spatial coordinate. A spatial weight matrix $W = \{w_{ij}\}$ is defined based on neighborhood relationships in the learned layout, where $w_{ij}=1$ if feature $j$ is among the $k$ nearest spatial neighbors of feature $i$, and $w_{ij}=0$ otherwise. Let $\bar{a}$ denote the mean attribution value and $N$ the number of features.

Moran’s I is defined as

\[
I = 
\frac{N}{\sum_{i}\sum_{j} w_{ij}}
\cdot
\frac{\sum_{i}\sum_{j} w_{ij} (a_i - \bar{a})(a_j - \bar{a})}
{\sum_{i} (a_i - \bar{a})^2}.
\]

Positive values indicate spatial clustering of similar attribution magnitudes, whereas values near zero indicate spatial randomness. To assess statistical significance, we generated null distributions by randomly permuting attribution values across fixed coordinates and recomputing Moran’s I across multiple iterations.

\subsection{k-nearest neighbor spatial coherence}
In addition to global spatial autocorrelation, we evaluated local neighborhood consistency using a k-nearest neighbor (kNN) coherence measure. For each feature $i$, let $\mathcal{N}_k(i)$ denote the set of its $k$ nearest neighbors in the learned coordinate space. Define a binary indicator $z_i$ reflecting whether feature $i$ belongs to the top-$q$\% highest attribution values for a given class.

The local coherence score for feature $i$ is defined as

\[
C_i = \frac{1}{k} \sum_{j \in \mathcal{N}_k(i)} \mathbf{1}(z_j = z_i),
\]

where $\mathbf{1}(\cdot)$ denotes the indicator function. The overall spatial coherence is computed as the mean of $C_i$ across features classified as high attribution. This statistic quantifies whether high-importance features preferentially cluster within their local spatial neighborhoods.

Significance was assessed by comparing observed coherence scores to null distributions obtained by randomly permuting attribution labels across fixed coordinates. Together with Moran’s I, this provides complementary global and local measures of spatial organization in the learned layout.

\section{Datasets}

Unless otherwise specified, highly variable genes (HVGs) were selected based on variance across samples or cells within each dataset. All tabular inputs were standardized prior to modeling.

\subsection{Circulating cell-free RNA profiling (RARE-Seq).}
We evaluated Dynomap on a circulating cell-free RNA (cfRNA) cohort derived from a multi-cancer liquid biopsy study~\cite{Nesselbush2025}. The dataset contains plasma-derived cfRNA profiles spanning healthy individuals and multiple cancer types with annotated clinical stage and tissue of origin. Gene-level expression matrices were constructed from TPM values and restricted to Ensembl gene identifiers. Features with zero variance across samples were removed. Three classification tasks were defined. For binary detection, samples were grouped into Healthy (Control, LDCT Control, Meta-Reference Control) and Cancer (LUAD, LUAD (TKI), LIHC, PAAD, PRAD). For subtype prediction, cancer samples were mapped to LUAD, LIHC, PAAD, and PRAD categories, with LUAD (TKI) grouped under LUAD. For stage classification, only cancer samples with available stage annotations were retained, and samples with missing stage information were excluded. For the full-gene setting, we selected the top 4,000 highly variable genes based on variance across samples. In the signature setting, we used all genes within the curated 622-gene cfRNA panel without additional HVG filtering. Expression matrices were standardized using z-score normalization prior to modeling. All downstream analyses, including attribution and spatial statistics, were performed separately for the HVG and signature feature regimes.

\subsection{Platelet RNA sequencing in solid tumors (GSE68086).}
To assess performance on bulk transcriptomic data derived from tumor-educated platelets, we used the GSE68086 dataset~\cite{best2015rna}. This cohort comprises platelet RNA-seq profiles from patients with multiple cancer types and healthy controls. Samples include breast cancer (BRCA), colorectal cancer (CRC), glioblastoma (GBM), lung cancer, pancreatic cancer, and healthy individuals. Raw gene expression matrices were obtained from the Gene Expression Omnibus (GEO). After alignment with series matrix metadata and removal of zero-variance genes, the top 5,000 highly variable genes based on variance across samples were retained for modeling. Expression values were standardized prior to modeling. Two tasks were defined: (i) binary cancer versus healthy classification and (ii) five-class tumor type prediction (Breast, CRC, GBM, Lung, Pancreas).

\subsection{TCGA breast cancer (TCGA-BRCA).}
We analyzed RNA-seq expression data from The Cancer Genome Atlas (TCGA) breast cancer cohort~\cite{TCGA_BRCA_2012}. Gene expression and clinical annotations were obtained from the UCSC Xena repository~\cite{goldman2020visualizing}. Expression values correspond to HiSeqV2 RNA-seq profiles and were aligned to clinical PAM50 subtype annotations~\cite{Parker2009}. Two classification tasks were defined: (i) binary aggressive versus non-aggressive tumor classification, where Basal and HER2-enriched tumors were labeled as aggressive, and (ii) multiclass PAM50 subtype prediction (LumA, LumB, Basal, HER2, Normal-like). From the aligned expression matrix, we selected the top 4,000 highly variable genes based on variance across samples. Expression values were standardized prior to modeling. No additional feature engineering was performed.

\subsection{Parkinson’s disease voice measurements.}
To evaluate generalization beyond transcriptomics, we used the Parkinson’s disease voice dataset originally described by Sakar \textit{et al.}~\cite{Sakar2019}. This dataset contains engineered acoustic features extracted from sustained phonations of individuals with Parkinson’s disease and healthy controls. Features include jitter, shimmer, harmonicity, nonlinear dynamical metrics, and wavelet-derived descriptors. We used the dataset in its published tabular form without feature selection or dimensionality reduction. All features were standardized prior to modeling. The task was binary classification of Parkinson’s disease versus control.

\subsection{Single-cell transcriptomics (Tabula Muris).}
To evaluate Dynomap in a large multicellular context, we used the Tabula Muris single-cell RNA-seq atlas~\cite{schaum2018single}. The dataset comprises 54,865 cells across 55 annotated cell types collected from 20 murine organs. We utilized a preprocessed expression matrix and selected the top 1,089 highly variable genes based on variance across cells to represent core transcriptional programs. Expression values were standardized prior to modeling. The task was multiclass cell-type classification across all annotated populations.

\section*{Data availability}
All datasets analyzed in this study are publicly available from established repositories. The circulating cell-free RNA (RARE-Seq) cohort was obtained from the multi-cancer liquid biopsy study reported by Nesselbush \textit{et al.}~\cite{Nesselbush2025}. Tumor-educated platelet RNA-sequencing data (GSE68086) were obtained from the Gene Expression Omnibus (GEO) under accession number GSE68086~\cite{best2015rna}. Breast cancer RNA-sequencing data (TCGA-BRCA) and associated clinical annotations were obtained from The Cancer Genome Atlas (TCGA)~\cite{TCGA_BRCA_2012} via the UCSC Xena repository~\cite{goldman2020visualizing}. The Parkinson’s disease voice dataset was obtained from the publicly released dataset described by Sakar \textit{et al.}~\cite{Sakar2019}. The Tabula Muris single-cell RNA-sequencing atlas was obtained from the Tabula Muris consortium~\cite{schaum2018single}.

All processed expression matrices, feature-selection outputs, and intermediate data files required to reproduce the analyses reported in this study are provided within the associated Code Ocean capsule prepared for peer review. Detailed preprocessing procedures and model training protocols are described in the Methods. No new human or animal data were generated for this study.

\section*{Code availability}
All code used to perform the analyses in this study has been deposited in a Code Ocean capsule and shared with the editors and reviewers as part of the peer-review process. The capsule contains the complete implementation of the Graph2Image framework, including data preprocessing, image construction, model training, and evaluation pipelines.

Upon acceptance of the manuscript, the Code Ocean capsule will be made publicly available with a persistent digital object identifier (DOI) to enable full reproducibility of the results.

\bibliography{main}

\clearpage
\appendix
\setcounter{figure}{0}
\renewcommand{\thefigure}{S\arabic{figure}}
\setcounter{table}{0}
\renewcommand{\thetable}{S\arabic{table}}
\section*{Supporting Information}

\begin{figure}[pt]
\centering
\includegraphics[width=\linewidth, page = 10]{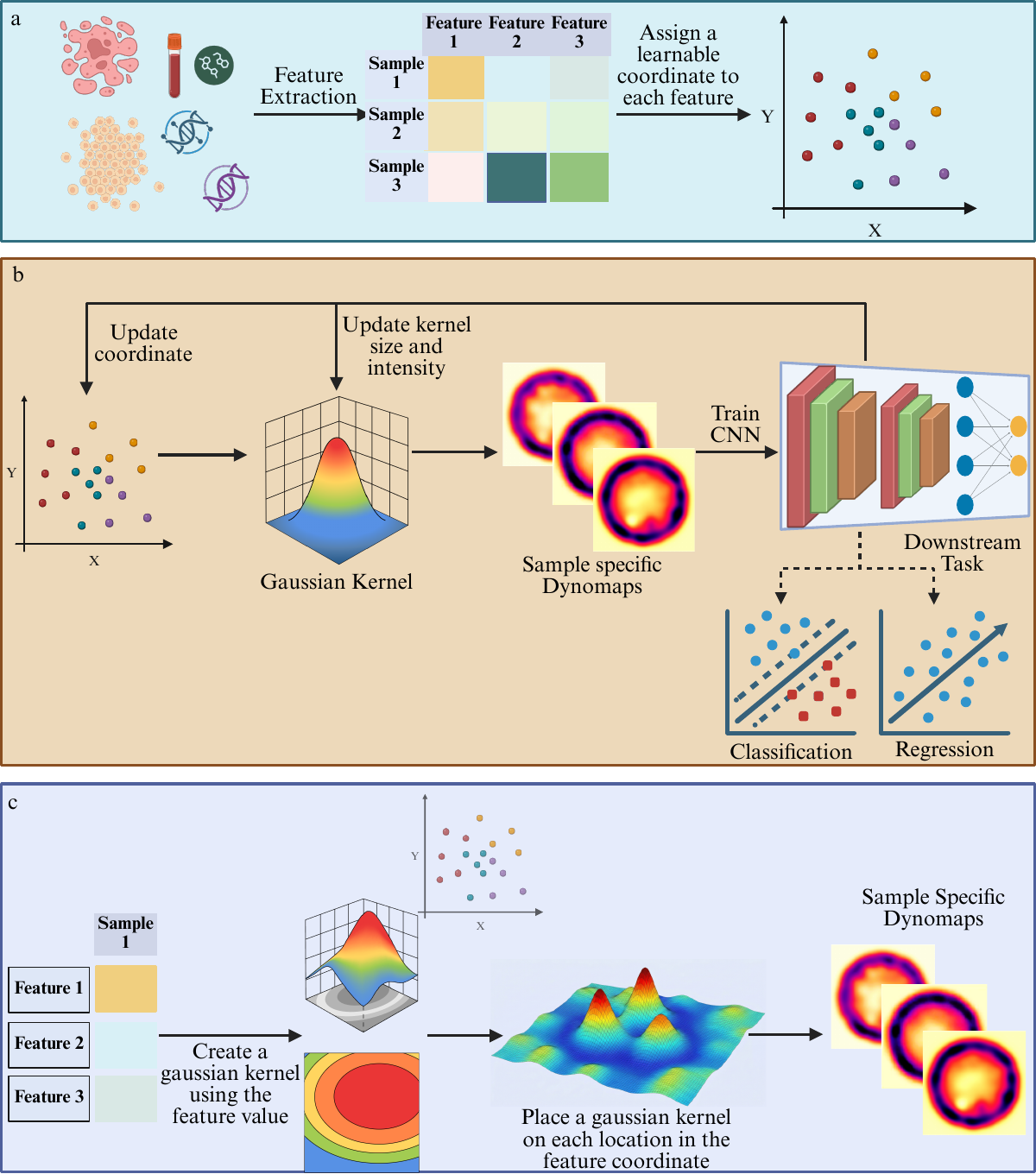}
\caption{\textbf{Detailed illustration of Dynomap construction from learned feature coordinates.}
\textbf{a}, Tabular biomedical measurements are first organized into a feature matrix, where each column corresponds to a biological variable and each row represents a sample. Dynomap assigns every feature a learnable coordinate in a continuous two-dimensional space, establishing a spatial layout that reflects task-dependent relationships among variables. 
\textbf{b}, During model training, both the feature coordinates and rendering parameters are updated jointly with the predictive model. Feature values are projected onto the spatial grid through Gaussian kernels whose location corresponds to the learned feature coordinate and whose intensity reflects the feature magnitude. The resulting spatial maps are used as inputs to a convolutional neural network to perform downstream tasks such as classification or regression. 
\textbf{c}, Detailed rendering process for a single sample. Each feature value generates a Gaussian kernel centered at its learned coordinate. The superposition of all kernels produces a continuous two-dimensional surface, which is discretized into an image representation referred to as a Dynomap. This process preserves feature identity while translating tabular measurements into a spatial representation that can be analyzed using vision-based models.}
\label{fig:figS10_Dynomap_Pipeline}
\end{figure}

\begin{figure}[pt]
\centering
\includegraphics[width=0.9\linewidth, page = 11]{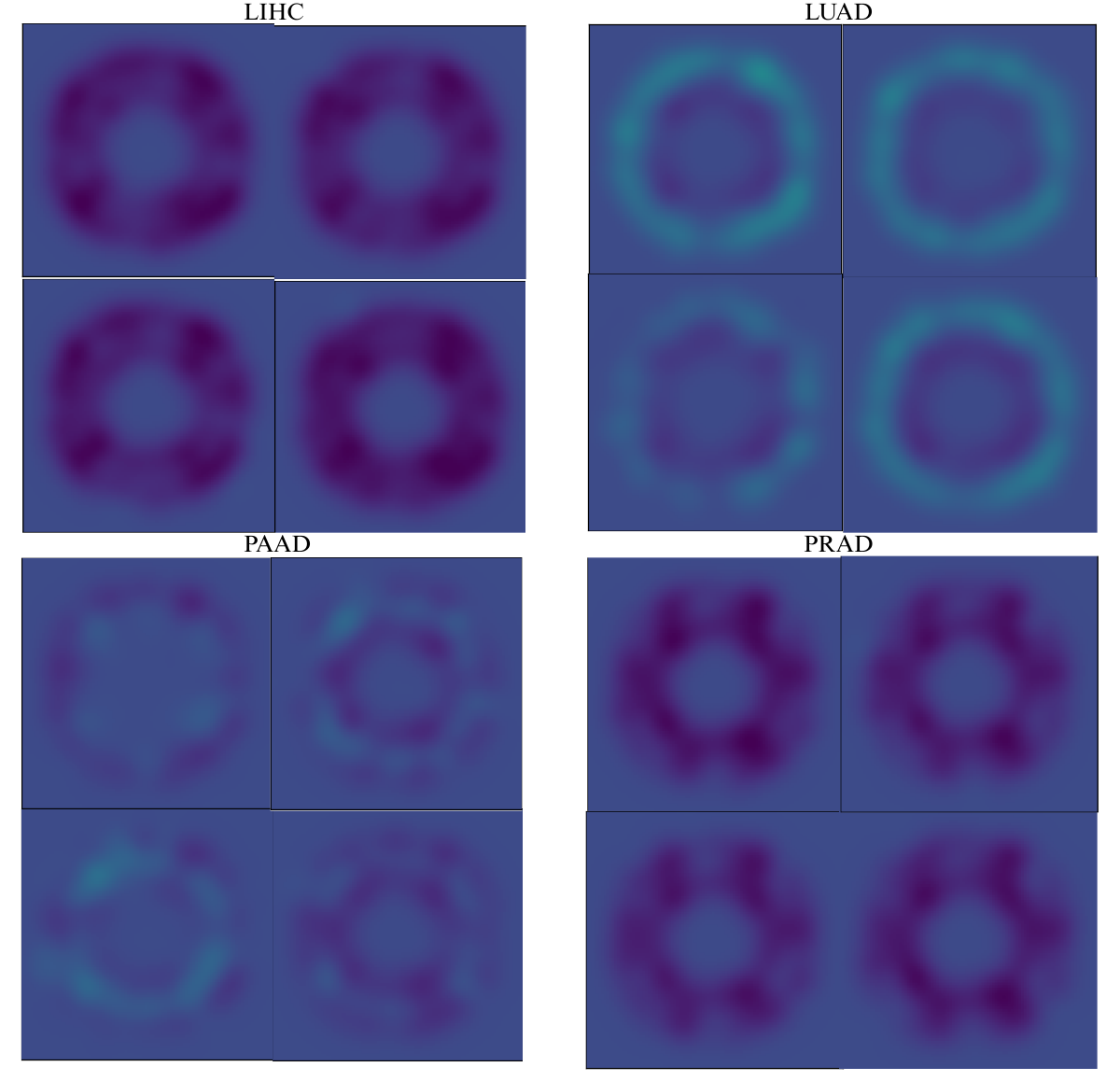}
\caption{\textbf{Representative Dynomap image embeddings across RARE-Seq classes.} Dynomap-generated image representations for individual RARE-Seq cfRNA samples, shown as class-wise examples across tasks. Within each class, samples exhibit visually consistent spatial organization, whereas different classes display distinct spatial configurations. These qualitative patterns illustrate that Dynomap maps tabular expression profiles into structured image-like representations without relying on a predefined gene layout. Color intensity denotes the normalized painted signal on the learned spatial canvas, with higher intensity corresponding to stronger localized contribution.}
\label{fig:figS9_rareseq_class_examples}
\end{figure}

\begin{figure}[pt]
\centering
\includegraphics[width=0.95\linewidth, page = 12]{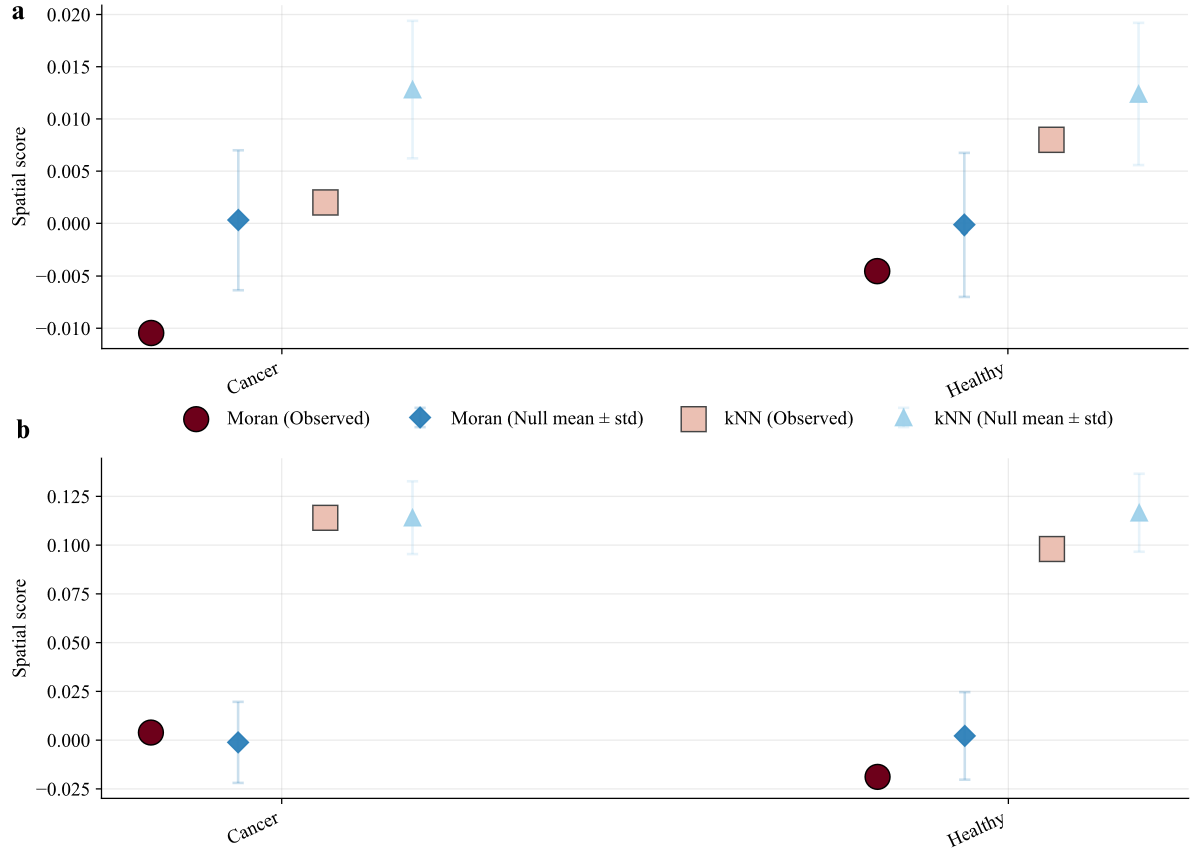}
\caption{\textbf{Spatial organization of class-specific gene attributions in RARE-Seq binary classification.}
Spatial autocorrelation and neighborhood consistency of Integrated Gradients (IG) attribution maps for healthy versus cancer prediction, quantified using Moran’s~$I$ and kNN purity.
\textbf{a}, Observed Moran’s~$I$ values for cancer- and healthy-associated genes compared with null distributions generated by spatial permutation. Cancer-associated attributions exhibit stronger deviation from null expectations, indicating non-random spatial organization.
\textbf{b}, kNN-based neighborhood purity for the same attribution maps. Cancer-associated genes show elevated local neighborhood consistency relative to null expectations, whereas healthy-associated attributions display weaker spatial structure.
Together, these results demonstrate that Dynomap organizes cancer-relevant genes into spatially coherent regions even in a binary classification setting.}
\label{fig:figS10_rareseq_binary_spatial}
\end{figure}

\begin{figure}[pt]
\centering
\includegraphics[width=0.95\linewidth, page = 13]{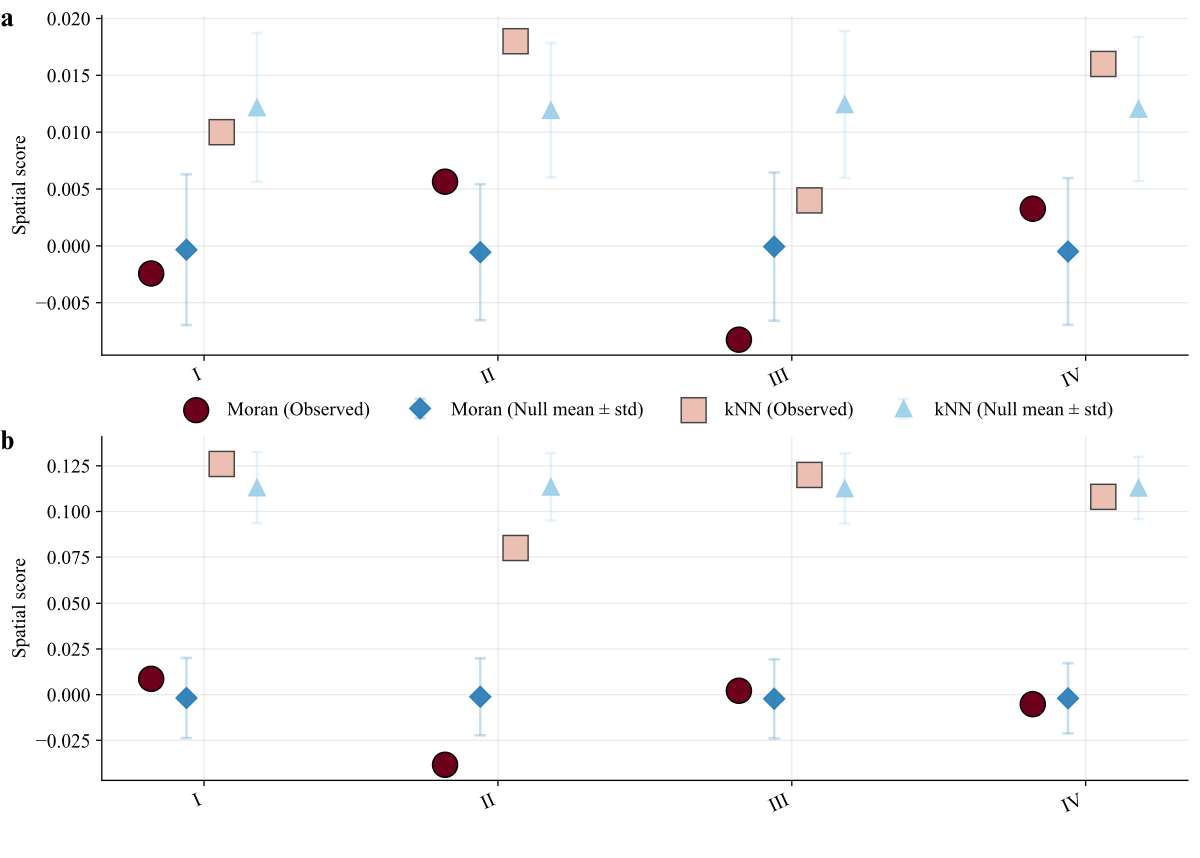}
\caption{\textbf{Stage-specific spatial structure of gene attributions in RareSeq.}
Spatial statistics of Integrated Gradients (IG) attribution maps for cancer stage prediction (Stages~I–IV), assessed using Moran’s~$I$ and kNN purity.
\textbf{a}, Observed Moran’s~$I$ values across stages compared with null distributions, revealing heterogeneous spatial organization of stage-specific gene importance. Stages II and IV show stronger spatial autocorrelation than expected under random layouts.
\textbf{b}, kNN purity analysis indicates elevated local neighborhood consistency for multiple stages, confirming that genes emphasized for the same stage tend to cluster spatially.
These results indicate that Dynomap adapts its spatial organization to disease progression, redistributing predictive gene signal in a stage-dependent manner.}
\label{fig:figS11_rareseq_stage_spatial}
\end{figure}

\begin{figure}[pt]
\centering
\includegraphics[width=0.95\linewidth, page = 14]{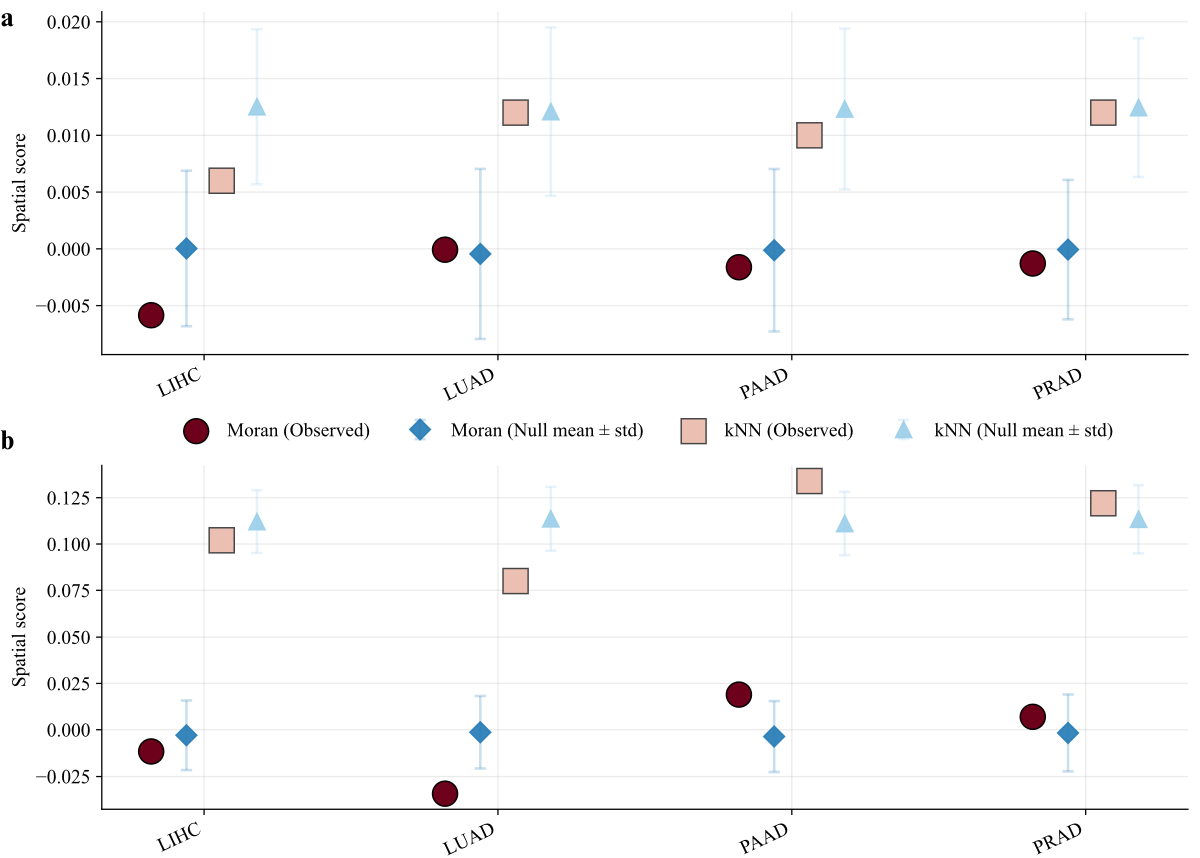}
\caption{\textbf{Subtype-dependent spatial clustering of gene attributions in RareSeq.}
Quantification of spatial autocorrelation and neighborhood consistency for subtype-specific Integrated Gradients (IG) attribution maps across multiple cancer types.
\textbf{a}, Moran’s~$I$ values for liver hepatocellular carcinoma (LIHC), lung adenocarcinoma (LUAD), pancreatic adenocarcinoma (PAAD), and prostate adenocarcinoma (PRAD), shown alongside null expectations. Subtypes exhibit distinct autocorrelation profiles, reflecting task-specific spatial organization.
\textbf{b}, kNN purity scores demonstrate consistently elevated neighborhood consistency across subtypes relative to null distributions, indicating robust spatial clustering of subtype-relevant genes.
These patterns show that Dynomap induces subtype-specific spatial structure rather than relying on a shared set of globally dominant genes.}
\label{fig:figS12_rareseq_subtype_spatial}
\end{figure}

\begin{figure}[pt]
\centering
\includegraphics[width=0.9\linewidth, page = 15]{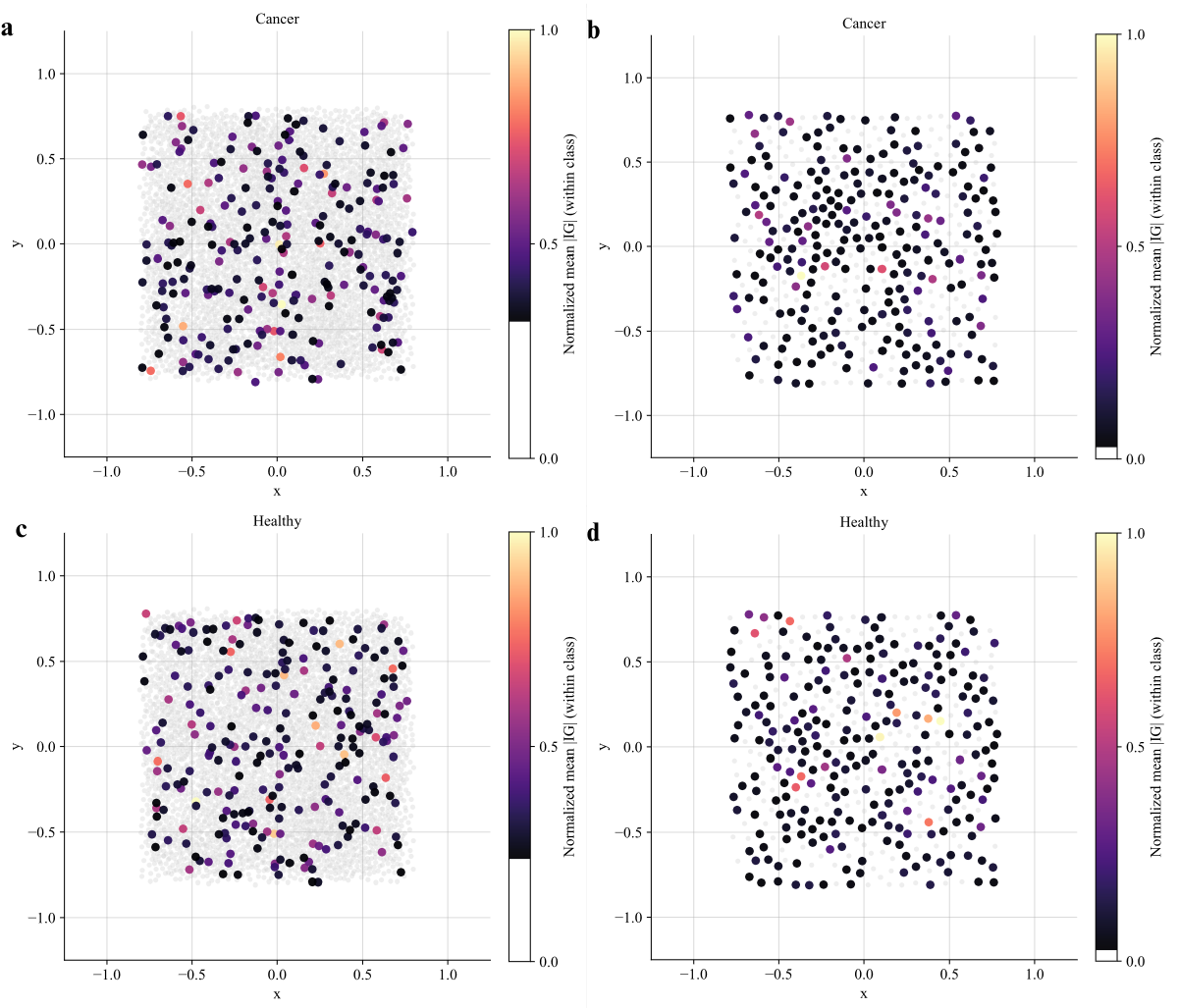}
\caption{\textbf{Class-specific feature attribution projected onto the Dynomap layout for RARE-Seq binary classification.}
\textbf{a,b}, Cancer-class layouts showing gene positions colored by normalized mean Integrated Gradients (IG) values within class.
\textbf{c,d}, Corresponding healthy-class layouts.
Gray points indicate all genes in the layout, while colored points highlight genes with higher class-specific attribution.}
\label{fig:figS13_rareseq_layout_binary}
\end{figure}

\begin{figure}[pt]
\centering
\includegraphics[width=0.9\linewidth, page = 16]{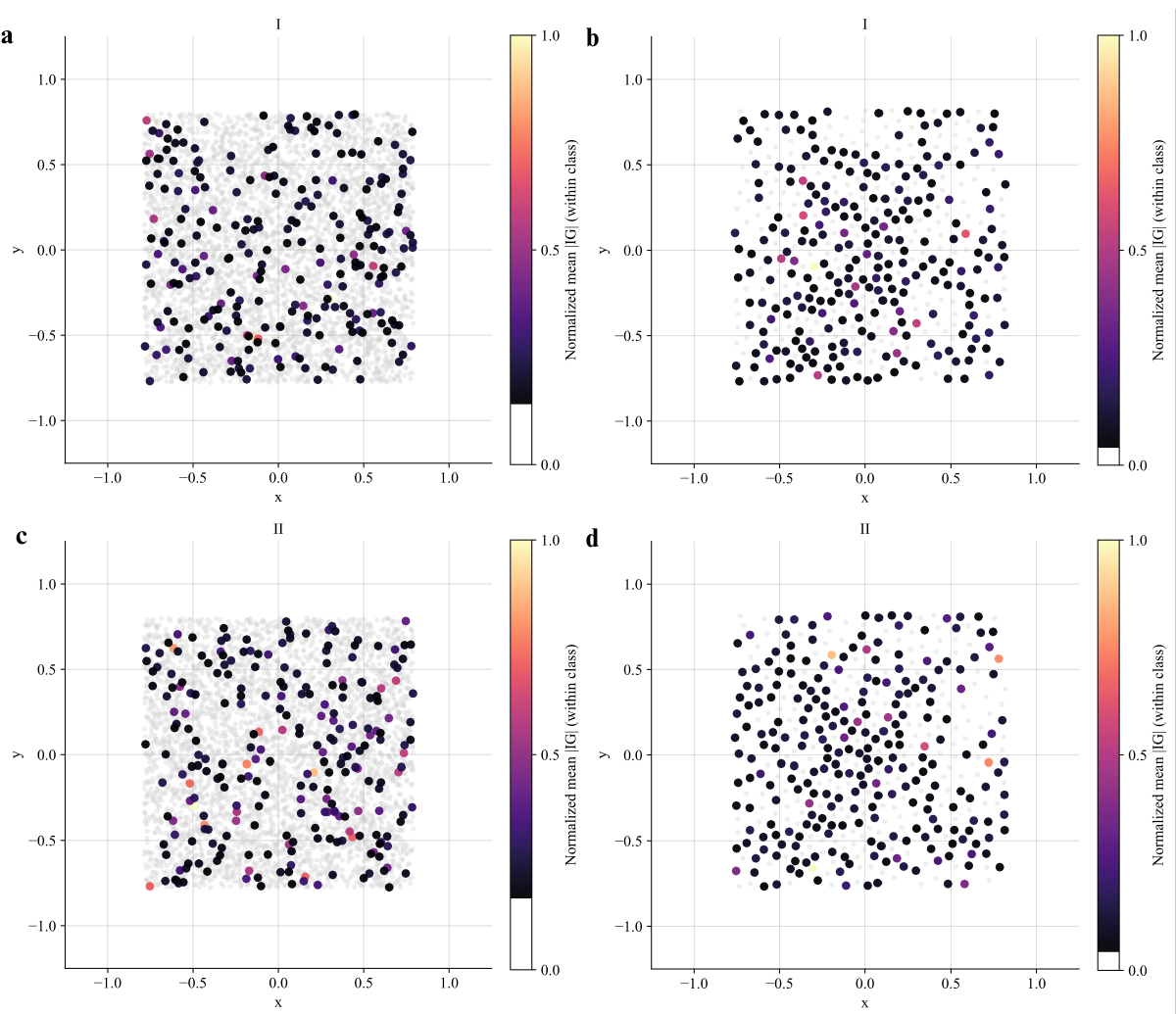}
\caption{\textbf{Stage-specific Dynomap layouts for RARE-Seq stages I and II.}
\textbf{a,b}, Stage I layouts showing IG-weighted gene importance.
\textbf{c,d}, Stage II layouts.
Gene positions are fixed across stages, enabling direct comparison of how discriminative regions shift with disease progression.}
\label{fig:figS14_rareseq_layout_stage_12}
\end{figure}

\begin{figure}[pt]
\centering
\includegraphics[width=0.9\linewidth, page = 17]{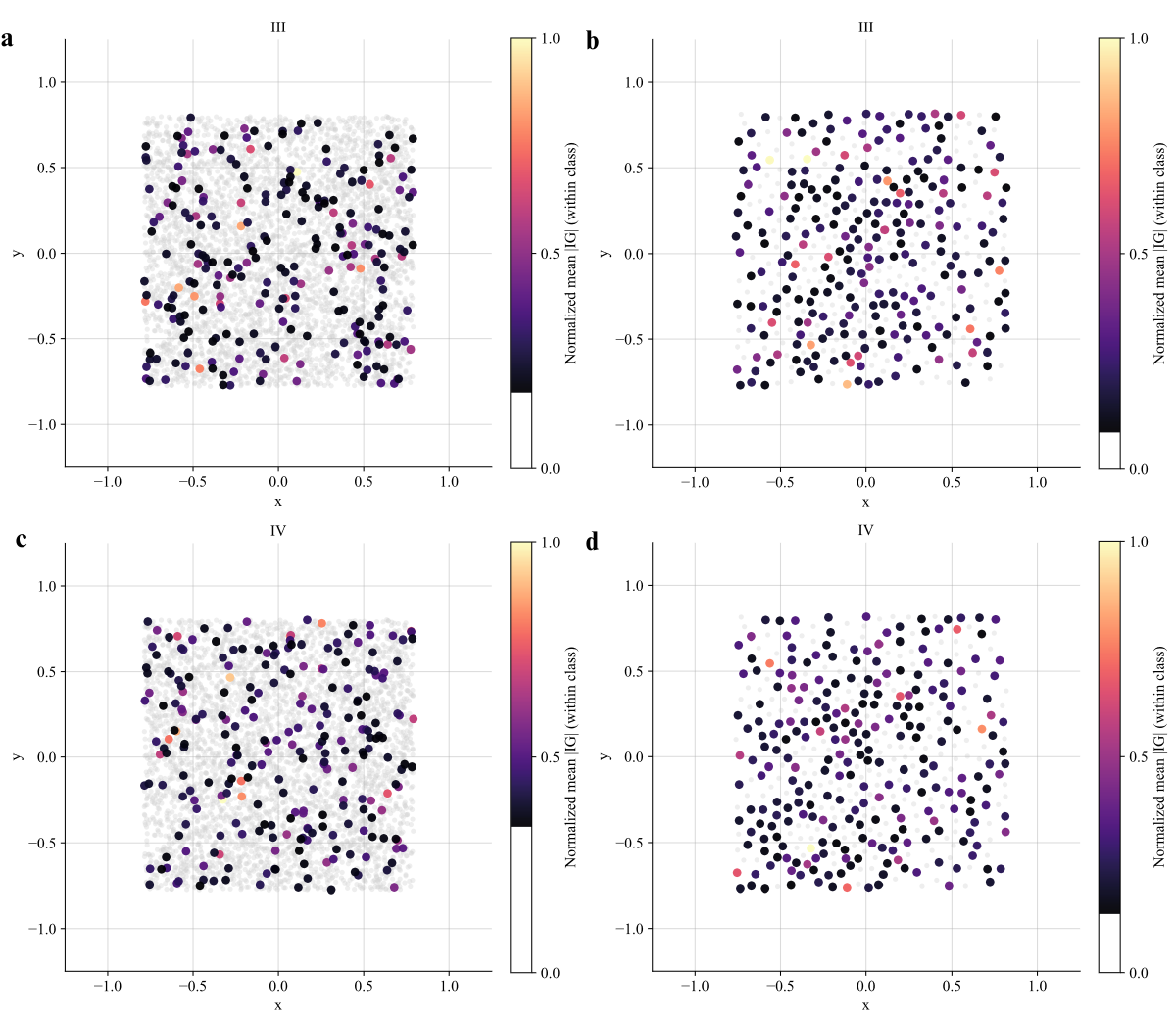}
\caption{\textbf{Stage-specific Dynomap layouts for RARE-Seq stages III and IV.}
\textbf{a,b}, Stage III layouts colored by normalized IG values.
\textbf{c,d}, Stage IV layouts.
Later-stage disease is associated with altered spatial emphasis patterns while preserving the underlying gene layout geometry.}
\label{fig:figS15_rareseq_layout_stage_34}
\end{figure}

\begin{figure}[pt]
\centering
\includegraphics[width=0.9\linewidth, page = 18]{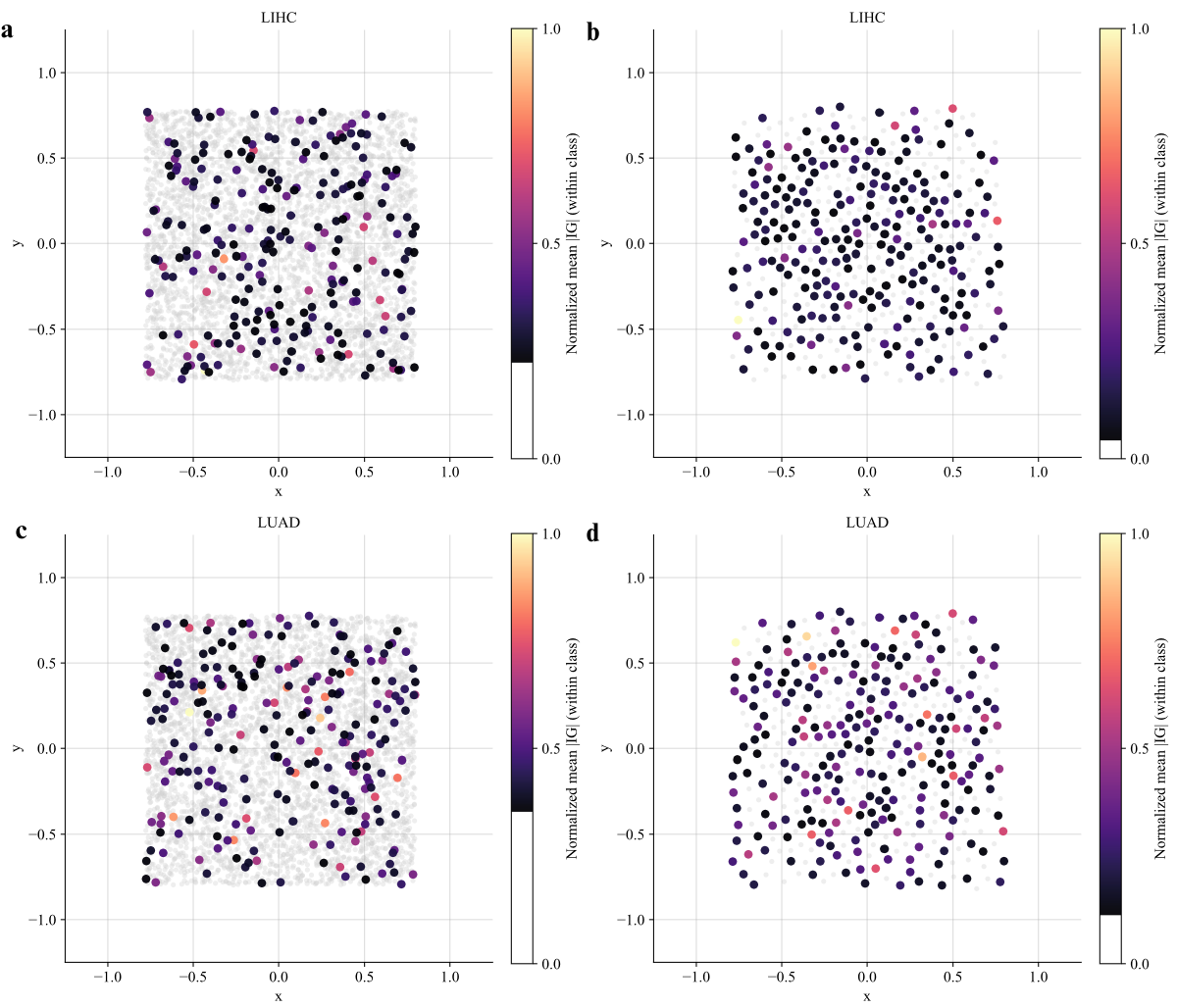}
\caption{\textbf{Subtype-specific Dynomap layouts for RARE-Seq (subset 1).}
\textbf{a,b}, IG-weighted gene layouts for two molecular subtypes.
Colored points represent genes with higher subtype-specific attribution, while gray points denote the full gene set.
Distinct spatial emphasis patterns are observed across subtypes within a shared layout.}
\label{fig:figS16_rareseq_layout_subtype_1}
\end{figure}

\begin{figure}[pt]
\centering
\includegraphics[width=0.9\linewidth, page = 19]{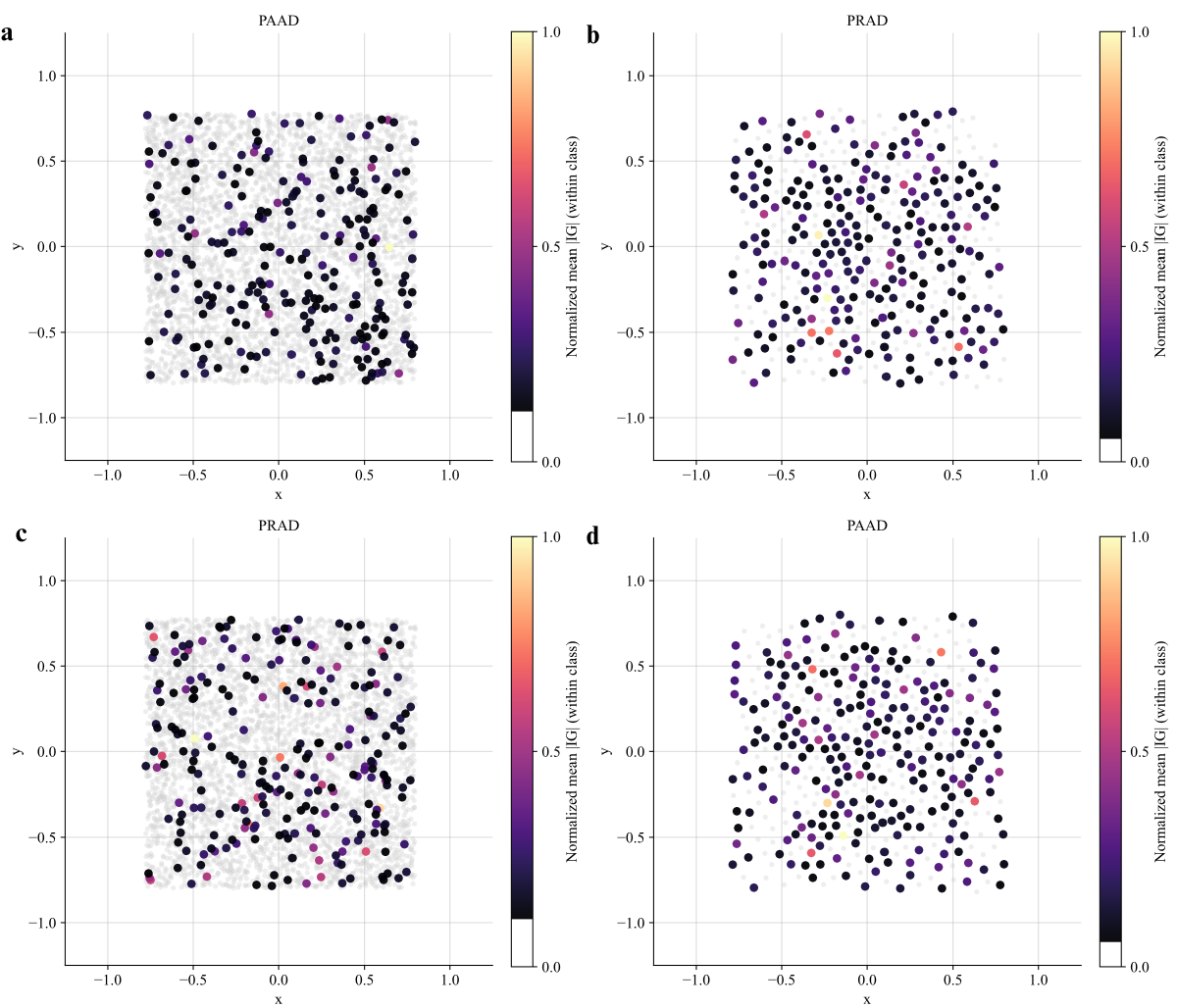}
\caption{\textbf{Subtype-specific Dynomap layouts for RARE-Seq (subset 2).}
\textbf{a,b}, Additional subtype layouts highlighting class-specific attribution patterns.
\textbf{c,d}, Corresponding complementary subtypes.
The reuse of a common spatial scaffold allows subtype distinctions to emerge through localized attribution rather than global rearrangement.}
\label{fig:figS17_rareseq_layout_subtype_2}
\end{figure}

\begin{figure}[pt]
\centering
\includegraphics[width=0.9\linewidth, page = 20]{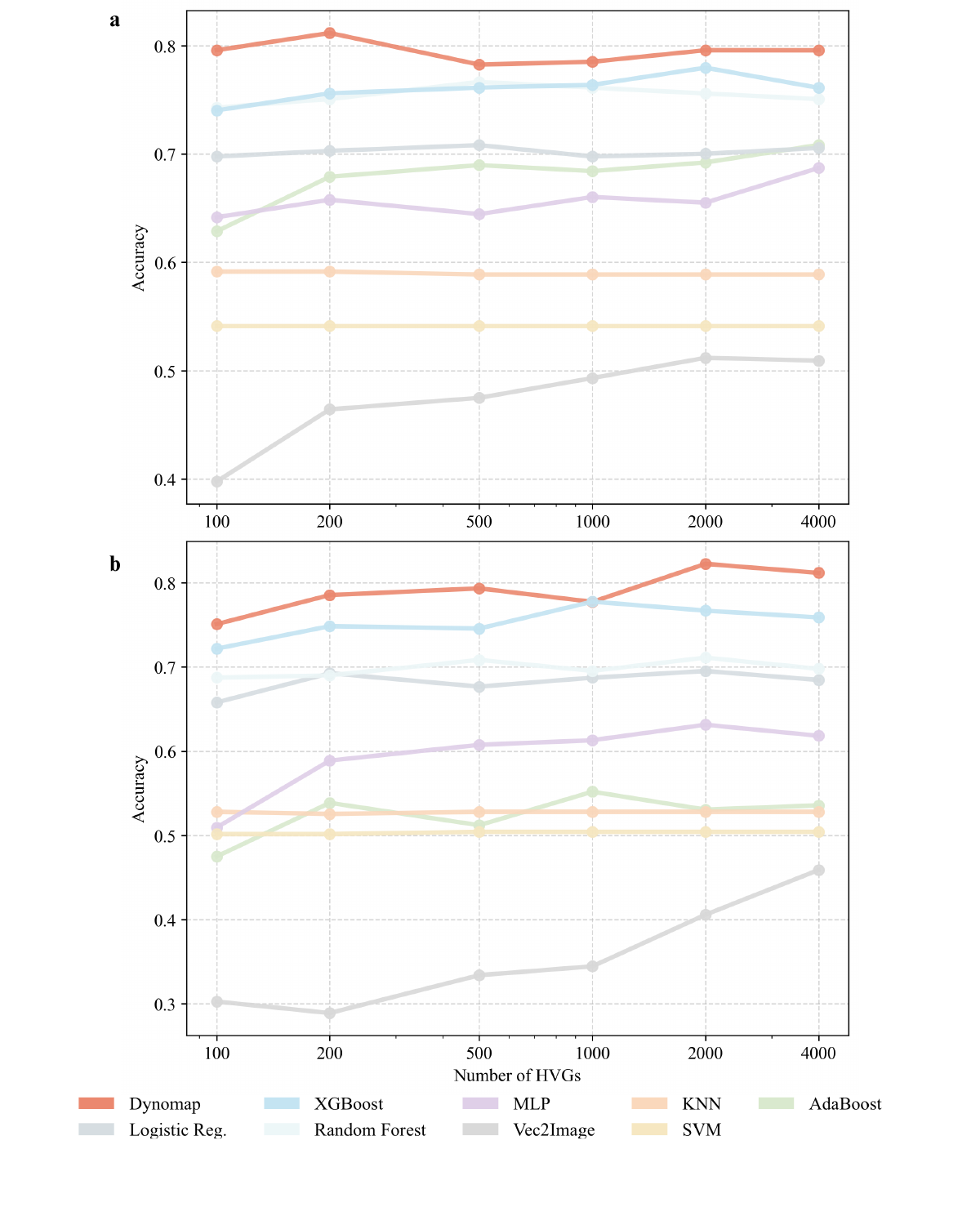}
\caption{\textbf{Effect of highly variable gene (HVG) selection on classification performance in RareSeq.}
Classification accuracy is shown as a function of the number of HVGs used to construct the input representation.
\textbf{Top}, Binary cancer versus healthy classification.
\textbf{Bottom}, Multiclass subtype classification.
Dynomap maintains consistently higher accuracy across a wide range of HVG counts compared with baseline methods, demonstrating robustness to feature dimensionality and stability of the learned representation under varying gene selection regimes.}
\label{fig:figS18_rareseq_hvg_sweep}
\end{figure}

\begin{figure}[pt]
\centering
\includegraphics[width=0.9\linewidth, page = 21]{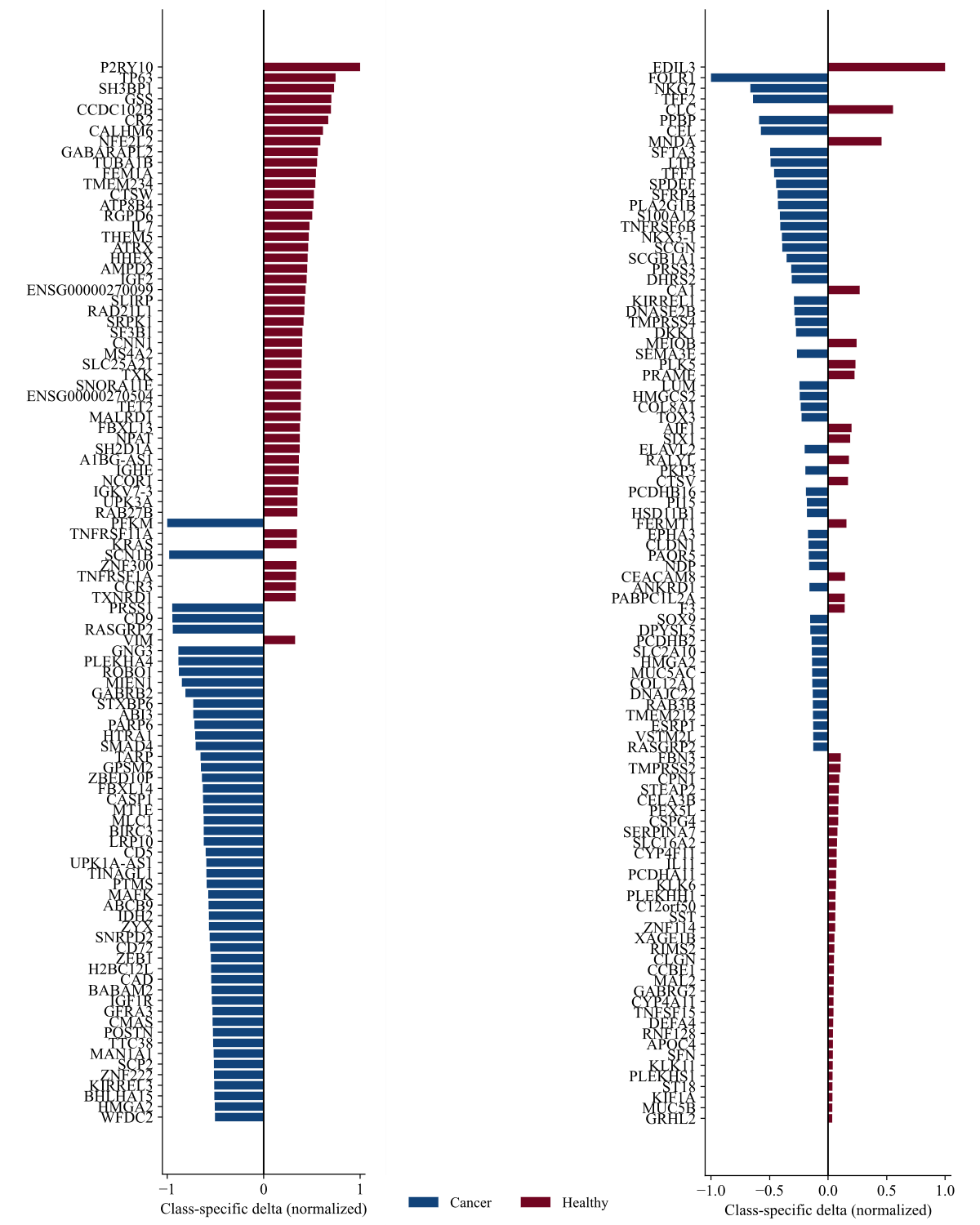}
\caption{\textbf{Differential gene importance for binary healthy vs. cancer detection.} Horizontal bars indicate the normalized difference in Integrated Gradients (IG) attribution strength between classes. \textbf{a}, Differential attribution using the full gene set shows distinct molecular signatures for healthy (red) and cancer (blue) predictions. \textbf{b}, Differential attribution using the 622-gene signature panel reveals a more concentrated set of discriminative features. The shift in specific top-ranked genes across settings demonstrates that Dynomap adapts its feature selection strategy based on the available genomic space.}
\label{fig:figS19_rareseq_binary_ig}
\end{figure}

\begin{figure}[pt]
\centering
\includegraphics[width=0.7\linewidth, page = 22]{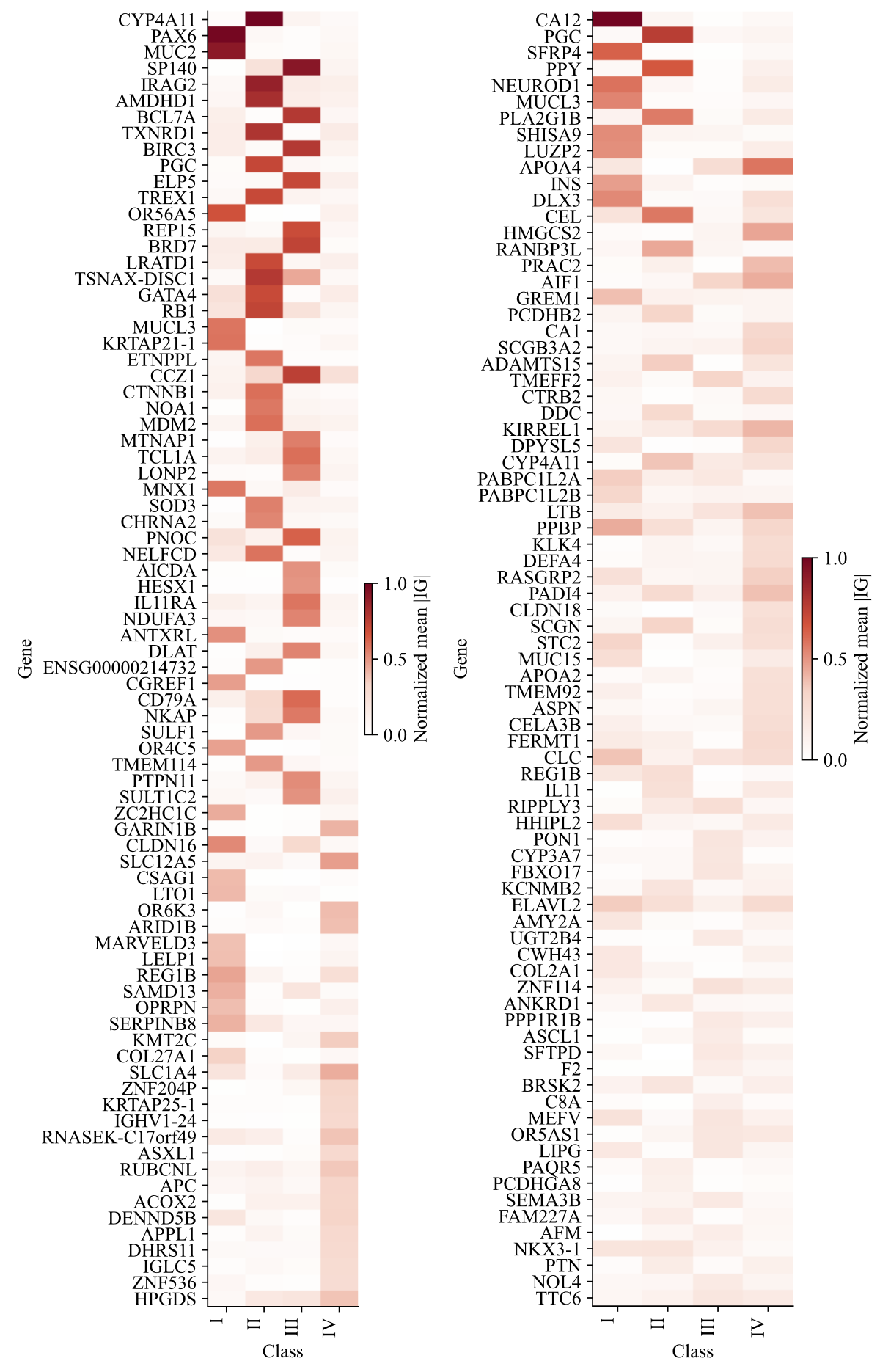}
\caption{\textbf{Subtype-specific gene attribution across cfRNA profiles.} Heatmaps of normalized Integrated Gradients (IG) show distinct transcriptional programs utilized for multiclass cancer subtype classification. \textbf{a}, Gene utilization for subtype prediction (LIHC, LUAD, PAAD, PRAD) using the full gene set. \textbf{b}, Subtype-specific attribution patterns using the 622-gene signature panel. While top predictive genes differ between feature regimes, the model consistently identifies non-overlapping gene sets for each tumor class, supporting the emergence of class-separable spatial representations.}
\label{fig:figS20_rareseq_subtype_ig}
\end{figure}

\begin{figure}[pt]
\centering
\includegraphics[width=0.7\linewidth, page = 23]{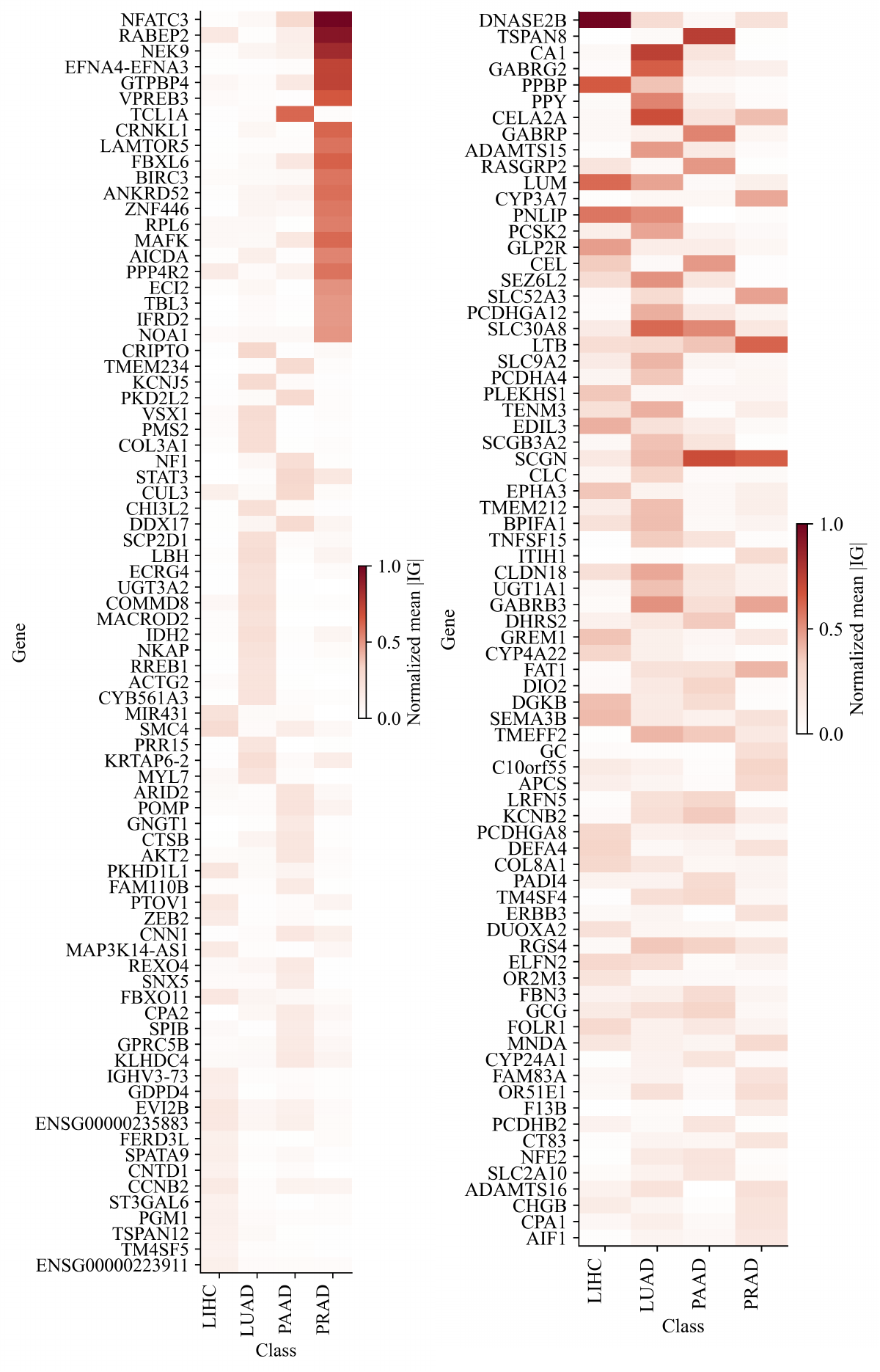}
\caption{\textbf{Gene-level attribution for liquid biopsy cancer stage prediction.} Task-adaptive feature importance profiles across clinical stages (I, II, III, and IV). \textbf{a}, Normalized mean Integrated Gradients (|IG|) for top predictive genes using the full gene set. \textbf{b}, Stage-specific attribution using the 622-gene signature panel. The model identifies a more compact set of stage-informative genes, reflecting a redistribution of predictive signal when the feature space is restricted. The distinct patterns across columns confirm that Dynomap captures subtle signals necessary for fine-grained clinical stratification.}
\label{fig:figS21_rareseq_stage_ig}
\end{figure}


\begin{table}[p]
  \centering
  \caption{Tabula Muris SHAP gene list (part 1 of 3; first 400 genes).}
  \label{supp_table:tm_shap_genes_part_1}
  \begin{tabular}{llllllll}
    \hline
    Fcer1g & Ubb & Cd79a & Cd79b & Cd3g & Bst2 & Il1b & Vpreb3 \\
    Cfh & Cst3 & Fxyd5 & Tmsb10 & Laptm5 & Lsp1 & Gnas & Gng11 \\
    S100a4 & Alox5ap & Ly6c1 & Plvap & Ctsl & Irf8 & H2\_K1 & Tpm2 \\
    Klf2 & Lyz2 & Sepp1 & Ccl3 & Vim & Socs3 & Lyz1 & Igfbp2 \\
    Krt14 & Ramp2 & Ckb & Oaz1 & Gm1821 & Marcksl1 & Arpc2 & Arpc3 \\
    Hbb\_b2 & Car2 & Scgb1a1 & Prss34 & Ccl9 & Defb1 & Tyrobp & Gzma \\
    AW112010 & Nkg7 & Ifitm1 & Tg & H2\_D1 & Emp3 & Hmgb2 & Lgals1 \\
    Dek & Ly6c2 & Coro1a & Slc12a1 & Fbln1 & Ccl6 & Ctsd & Hmgn2 \\
    S100a8 & Krt19 & Shisa5 & Rps15a\_ps6 & Clic1 & Rps20 & B2m & Cmtm7 \\
    Igj & Crip1 & Igfbp7 & Ifitm3 & Fabp4 & Sparcl1 & Selk & Cd81 \\
    Jund & Capzb & Krt7 & Serp1 & Cd52 & A130040M12Rik & Ctsb & Sln \\
    Myl4 & Nppa & Ifitm6 & Retnlg & S100a9 & x1100001G20Rik & Laptm4a & Rhob \\
    Tmsb4x & Cxcl15 & Sftpb & Sftpc & Ly6a & Atp1b1 & Ly6e & Ppap2b \\
    Tuba1a & Clec3b & Sod3 & Pcolce2 & Tnfaip6 & Dntt & Ptprcap & Grcc10 \\
    Sh3bgrl3 & Pck1 & Slc34a1 & Errfi1 & Mt2 & Btg1 & Hspa1a & Ccnl1 \\
    Meg3 & Vcam1 & Id3 & Ndufa2 & x1110003E01Rik & Upk1a & Upk1b & Ifitm2 \\
    Cfl1 & Rpl37 & Gadd45b & Zfp36 & Jun & Rps6 & Postn & Rpl13 \\
    Rps16 & Dpt & Ppp1r15a & Rpl7 & Srsf5 & Npc2 & Cryab & Gatm \\
    Pabpc1 & Rps26 & Atp5k & Cyc1 & Tceb2 & Sec61g & x2410015M20Rik & Actr3 \\
    Ubl5 & Mdh2 & Vdac1 & Atox1 & Aplp2 & Cox6c & Lamp1 & Cdc42 \\
    x1190002H23Rik & Cks2 & Ccl7 & Cyr61 & Aqp3 & Perp & Cd9 & Atp5d \\
    Tbca & Mif & Pim3 & Eif6 & Trim29 & Jup & Krt6a & Vsig8 \\
    Rab25 & Fabp5 & Ly6d & Rtn4 & Nfkbia & D14Ertd449e & Rpl9 & Rpl27a \\
    Rps10 & Eif4a1 & Lgals3 & x2010001M09Rik & Chchd10 & Ndufc2 & Psap & Ier2 \\
    Sepw1 & Psmb8 & Rhoa & Gabarap & Arpc5 & Tmed10 & Mal & Tpt1 \\
    Lum & Csrp1 & H3f3b & Cd24a & S100a11 & Cotl1 & Taldo1 & Ndufa7 \\
    Actb & Rpl37a & Rplp2 & Serinc3 & Tmem176a & Gdi2 & H2\_Aa & Rps15a\_ps4 \\
    Btg2 & Ccl17 & Tmem176b & Atf4 & Fosb & Gadd45g & Rnase4 & Cbr2 \\
    Ptpn18 & Rps2 & Rps11 & Rps17 & Atp5b & Rpl8 & Rplp1 & Txn1 \\
    Rpl18a & Rps8 & Sat1 & Ndufb10 & Chchd2 & Ccdc72 & Cox8a & Eif4g2 \\
    Atp5a1 & Btf3 & Ndufv3 & Myl12a & Rnasek & Foxq1 & Ivl & S100a6 \\
    Wfdc2 & Rplp0 & Pdlim1 & Rpl19 & Sdc1 & Igfbp3 & Gltscr2 & Krt5 \\
    Hspa8 & Rpl15 & S100a10 & Mrfap1 & Pkm2 & Psmb3 & Sfn & Csnk1a1 \\
    Rbm3 & Rps5 & Txndc17 & Cldn5 & H2\_Ab1 & Cd74 & H2\_Eb1 & Lmna \\
    Cd164 & Krt8 & H3f3a & Romo1 & Hmgn1 & Rps28 & H2afv & Mdh1 \\
    Eif5a & Rpl23 & Rps24 & Aldh2 & Rpl22l1 & Tomm6 & Fos & Gpx1 \\
    Arpc1b & Shfm1 & Klf9 & Cox7a2l & Rpl27 & Rpl18 & x1500012F01Rik & Eef1a1 \\
    Atf3 & Cct4 & Adamts1 & Pa2g4 & Atp6v0e & Timp3 & Rpl32 & Rpl36al \\
    Rps15 & Rpl41 & Rps27 & Hnrnpk & Ran & Pcbp1 & Usmg5 & Cycs \\
    Cnbp & Eif2s2 & Map1lc3a & Myl6 & Cyba & Sec11c & Hint1 & Birc5 \\
    Anp32e & Cenpa & Eif1 & Zfp706 & Atp1a1 & Bag1 & Rpl5 & Rpl35 \\
    Ifi27l2a & Atp6v0b & Slc38a2 & Ier5 & Pnrc1 & Tpm4 & Vtn & Apoa1 \\
    Akr1c6 & Gnmt & Srp9 & Nfe2l2 & x2310003F16Rik & Gsr & Hp & Cyp2f2 \\
    S100a1 & x1810037I17Rik & Rac2 & Brp44 & Nr4a1 & Egr1 & Arf1 & Ndufb4 \\
    Hist1h2bc & Hras1 & Sepx1 & Apoe & Ifrd1 & Prdx5 & Umod & Clta \\
    Hnrnpf & Rpl4 & Rps27a & Use1 & Tmed2 & Bsg & Tomm7 & Tpm3 \\
    Uqcrc1 & Ctla2a & Gmfg & Gas6 & Rpl35a & Pcbp2 & Sgk1 & Rps9 \\
    Psca & x2700060E02Rik & Rps14 & Gnb2l1 & Rpl29 & Eef2 & Lsm4 & Dusp1 \\
    Junb & Nme2 & Rpl11 & C3 & Igfbp4 & Tmem123 & Bgn & Wif1 \\
    Myl9 & Elane & Ms4a3 & Anxa1 & Glrx & H2afx & Mki67 & Pfn1 \\
    \hline
  \end{tabular}
\end{table}

\begin{table}[p]
  \centering
  \caption{Tabula Muris SHAP gene list (part 2 of 3; next 400 genes).}
  \label{supp_table:tm_shap_genes_part_2}
  \begin{tabular}{llllllll}
    \hline
    Park7 & Rpl17 & Sub1 & Ube2i & Erp29 & Mcpt8 & Icam1 & Tsc22d1 \\
    Atp5e & Fxyd3 & Rps18 & Ndufb9 & x3110003A17Rik & Dmkn & Uqcrfs1 & Rps19 \\
    Rps25 & Ltb & Hnrnpab & Plac9 & Dnaja1 & Itm2b & Id1 & Ier3 \\
    Rarres2 & Csrp2 & Mmp3 & Calm1 & Gnai2 & Pfdn5 & Sumo2 & Ctnnb1 \\
    Atp5j & Cox4i1 & Rpsa & Slc25a4 & Cct7 & Ddx5 & Atp5h & Rpl23a \\
    Snrpd2 & Tagln & Chmp2a & Ndufa4 & Ccl5 & Nbl1 & Rps21 & Uqcrb \\
    Rpl10 & Arpc1a & Rpl10a & Akr1a1 & Etfb & Pomp & Srgn & Ubc \\
    Ftl1 & Rpl28 & Rps27l & Stmn1 & Tubb5 & H2afy & Hist1h2ae & Ssr4 \\
    Ctsg & Top2a & Aes & Rpl21 & Rpl3 & Reep5 & Cct5 & Tmbim6 \\
    Rps4x & Ywhaq & Atp6v1g1 & Arl6ip1 & Aqp1 & Ldhb & Cd36 & Cxcl12 \\
    Cd14 & C1qa & C1qb & Ndufa8 & Edf1 & Hprt & Fkbp1a & Top1 \\
    Mfap5 & Mmp2 & Ccl11 & Pi16 & Col6a1 & Ugdh & Dcn & Serpinf1 \\
    Mt1 & Myoc & Smoc2 & Pcolce & Cebpd & Serping1 & Cxcl1 & Fgl2 \\
    Zfp36l1 & Rps29 & Tmem134 & D8Ertd738e & Slc25a5 & Smc4 & x2410006H16Rik & Chmp4b \\
    Hk2 & Ahnak & Sdc4 & Atp5o & Dynlrb1 & Dad1 & Aldoa & Map1lc3b \\
    Hdgf & Uqcrh & Cox6b1 & Nme1 & Rpl7a & Rps12 & Rps7 & Nhp2l1 \\
    Polr1d & Cdkn1a & Ywhaz & Rps3 & Maff & Ndufb8 & Klf6 & Mgst3 \\
    Rps23 & Gnb2 & Hspe1 & Rpl12 & Pgk1 & Crlf1 & Hist1h1c & Klf5 \\
    Phb2 & Arf5 & Avpi1 & Pebp1 & Psmb5 & Crip2 & Ptma & Prdx1 \\
    Psmb6 & Snrpb & Rpl6 & Uba52 & Cox6a1 & Nap1l1 & Hnrnpa2b1 & Ndufa12 \\
    Rps3a & Zfand5 & Eef1b2 & Eif3h & x1110008F13Rik & Vamp8 & Xist & Rpl31 \\
    H2afj & Tst & Dstn & Dbi & Slc25a3 & Pycard & Rnh1 & Pir \\
    Ppp1r14b & Serbp1 & x2610528A11Rik & Rpl14 & Arf6 & Dnajb1 & Cstb & Emp2 \\
    Rpl22 & Fosl1 & Pkp1 & Ovol1 & Hbegf & Nhp2 & Ndufa4l2 & Capg \\
    Calml3 & Them5 & Capns2 & Gpx2 & S100a14 & Ptgr1 & Lypd3 & Krt16 \\
    x2310033E01Rik & Slpi & Spink5 & Actg1 & Eef1d & Nedd8 & Ndufa3 & Plac8 \\
    Fau & Ndufb11 & Sqstm1 & Acta2 & Atp5l & Pdia3 & Spint2 & Gsto1 \\
    Ost4 & Rbx1 & Ppp1ca & Rps13 & Polr2l & Actn4 & Mgp & Swi5 \\
    Hsp90aa1 & Snrpg & Limd2 & Pkp3 & Srsf3 & Slc3a2 & Arpp19 & Tuba1b \\
    Hsp90b1 & Sec61b & Glrx3 & Rac1 & Ccnd3 & Ltf & Ngp & Ap3s1 \\
    Tkt & Gpi1 & Chi3l3 & Camp & Fcnb & Atpif1 & Minos1 & Ndufc1 \\
    Prdx2 & Gm11428 & Hmox1 & Atp5f1 & Ccl4 & Napsa & Ctss & x1500032L24Rik \\
    Alas2 & Pglyrp1 & Eif3i & Nop10 & Clic4 & Lgals7 & Metap2 & Ppia \\
    Eif3m & Tubb4b & Atp5g3 & Gja1 & Ezr & Ldha & Hspd1 & Anxa8 \\
    Gm94 & Ube2d3 & Uqcrq & Thy1 & Tm4sf1 & Glul & F3 & Krt18 \\
    Snrpe & Anp32b & H2afz & Cltb & Gnb1 & Fth1 & Lpl & S100a16 \\
    Cyb5 & Odc1 & Tacstd2 & Psma7 & Rpl36a & Timp2 & Mrpl52 & Serf2 \\
    Srsf2 & Higd1a & Ndufb5 & Tuba4a & Myeov2 & Cd8a & Phlda1 & Cd8b1 \\
    Rpl31\_ps12 & Skp1a & Tmem59 & Blvrb & Uqcr10 & Psma2 & Rps15a & Hspa5 \\
    Pcna & Oat & Ncl & Ybx1 & Gpx3 & x2200002D01Rik & Gsta4 & Hes1 \\
    Sparc & Cpe & Fmod & Eif3k & Chad & Prg4 & Srsf7 & Pdcd4 \\
    Serpine2 & Gcat & Ppp1r2 & Rpl39 & Cox5a & Tpi1 & Rab10 & Chit1 \\
    Ndufs6 & Uqcr11 & Ndufa6 & Ucp2 & Ndufa13 & Pgam1 & Id2 & Atp5g2 \\
    Rcan1 & Arhgdib & Hnrnpu & x2810417H13Rik & Krtcap2 & Hist1h1e & Rrm2 & Manf \\
    Timm13 & Tspo & Naca & Fam162a & Sfr1 & Nrp1 & Acsm2 & Apob \\
    Hpx & Rabac1 & Cdo1 & Atp6v1f & Acaa1b & Fxyd2 & Akr1c21 & Slc27a2 \\
    Hpd & Apoh & Rgn & Serpina1b & Fabp1 & Mup2 & Ndrg1 & Upk3a \\
    Gclc & Col3a1 & Dynll1 & Hmgb1 & Gng5 & Ralbp1 & Ucma & Col11a1 \\
    Comp & Pdzk1ip1 & Ndufb7 & Calm4 & Anxa5 & Apoc4 & Gm13889 & Krt15 \\
    Eif5 & Glycam1 & Ube2c & Psma3 & Rpl24 & Rpl38 & Eef1g & Gapdh \\
    \hline
  \end{tabular}
\end{table}

\begin{table}[p]
  \centering
  \caption{Tabula Muris SHAP gene list (part 3 of 3; remaining 288 genes).}
  \label{supp_table:tm_shap_genes_part_3}
  \begin{tabular}{llllllll}
    \hline
    Acp5 & Rap1b & Hspb8 & Col2a1 & Il6 & Tomm20 & Dsp & Klf4 \\
    Malat1 & Epsti1 & Ccl2 & Has1 & Neat1 & Prelid1 & Calr & Hspb1 \\
    P4hb & Ppib & Fdps & Mup3 & Cxcl14 & Cxcl2 & x2010107E04Rik & Hrsp12 \\
    Miox & Cat & Col8a1 & Cxcl10 & Mgst1 & C1qc & Rpl13a & Mpo \\
    Lcn2 & Prtn3 & Eif3e & Gas5 & Impdh2 & Npm1 & Tagln2 & Cox7b \\
    Hsp90ab1 & Tcp1 & Procr & Psmb1 & Trf & Emp1 & x1500015O10Rik & Hist1h2ao \\
    Myl7 & Tpm1 & Rab11a & Akr1b8 & Rpl36 & Scgb3a2 & Sbpl & Scgb3a1 \\
    Tff2 & Ndufa1 & Serpinb5 & Eif3f & Prdx6 & Areg & Cldn4 & Hnrnpa3 \\
    Eln & Tnfrsf12a & Tmem27 & Acy3 & Aldob & Cyp4b1 & Guca2b & Ranbp1 \\
    Anxa2 & Klk1 & Tppp3 & Psmb2 & Calm2 & Ttc36 & Tcn2 & Tecr \\
    Hilpda & Gstm1 & Ttr & Cyp2e1 & Scd1 & Esd & Sep15 & Egf \\
    Wfdc15b & Sprr1a & x1110032A04Rik & Bpifa1 & Mia1 & Sftpa1 & Fgg & Gltp \\
    Aldh3a1 & Dsg1a & Timp1 & Ccl19 & Fgb & Sult2b1 & Cd63 & Homer2 \\
    Rbp4 & Fmo5 & Sbsn & Snrpf & Cox5b & Scp2 & Atp5j2 & Cox7c \\
    Cma1 & Retnla & Ctnnbip1 & Erh & Serpinh1 & Fn1 & Ywhae & Set \\
    Apoc3 & Fga & Cyp3a11 & Mup20 & Spink3 & Mrpl33 & Apoa2 & Alb \\
    Serpina1d & Hn1 & Thbs1 & Fam25c & Gjb2 & Barx2 & Krt24 & Gem \\
    Ube2s & Spp1 & Gc & Krtdap & Car3 & Gsta3 & Ahsg & Krt10 \\
    Serpina1c & Serpina3k & Apoc1 & Kdm6b & Dut & Gpx4 & Tubb2a & Csf3 \\
    Epcam & Expi & Clu & S100g & Tmem14c & Atp5c1 & Col1a2 & Fgfbp1 \\
    Sprr1b & Cnfn & Selenbp1 & Dapl1 & Cst6 & Krt13 & Krt4 & Ccl20 \\
    Krt6b & x2310002L13Rik & Ly6g6c & Krt23 & Tgm3 & Atp5g1 & Col1a1 & Nupr1 \\
    Lmo1 & Tesc & Inmt & Psapl1 & Efhd2 & Apod & Calb1 & Col10a1 \\
    Krt17 & Chi3l1 & Serpinb2 & Mfge8 & Calca & Reg3g & Selp & Krt35 \\
    Igfbp5 & Serpinb3a & Elovl4 & x2310002J15Rik & Cldn7 & Ctsk & Prr9 & Msmp \\
    x2310079G19Rik & Lce3c & Serpina1a & Rn45s & Serpinb12 & Defb4 & Rbp2 & Klk14 \\
    Ptx3 & Adh7 & Gstp1 & Sox9 & Csn3 & Lrrc15 & Kap & x8430408G22Rik \\
    Atp2a2 & Myh6 & Ttn & Ada & Lor & Krt76 & Slurp1 & Crct1 \\
    Darc & Krt36 & Acan & Cytl1 & Basp1 & Krt84 & Tchh & Chi3l4 \\
    Lipf & x3110079O15Rik & G0s2 & Bpifb1 & Slc14a2 & Krt85 & Sprr2i & Otor \\
    Dcpp2 & Krt33b & Dcpp3 & Krtap3\_3 & Mfap4 & x1600029D21Rik & Car1 & Aqp2 \\
    Lyg1 & Serpina1e & Krtap13\_1 & Serpinb3c & Pf4 & Ppbp & Sprr3 & Ppa1 \\
    Krt34 & Krt81 & Bhmt & Krt31 & Dcpp1 & Krtap3\_1 & Krt42 & x5430421N21Rik \\
    Krtap13 & Krtap9\_3 & Psors1c2 & Krt33a & Krt86 & Krtap3\_2 & x2310034C09Rik & x2310061N02Rik \\
    Cyp2c69 & Hamp & Klk10 & Ptms & Cox7a2 & Mt4 & Pam & Pgls \\
    Aldh1a1 & Capns1 & Sod1 & Cebpb & Fstl1 & Gsn & Ifi205 & Igfbp6 \\
    \hline
  \end{tabular}
\end{table}

\begin{figure}[pt]
\centering
\includegraphics[width=0.8\linewidth, page = 24]{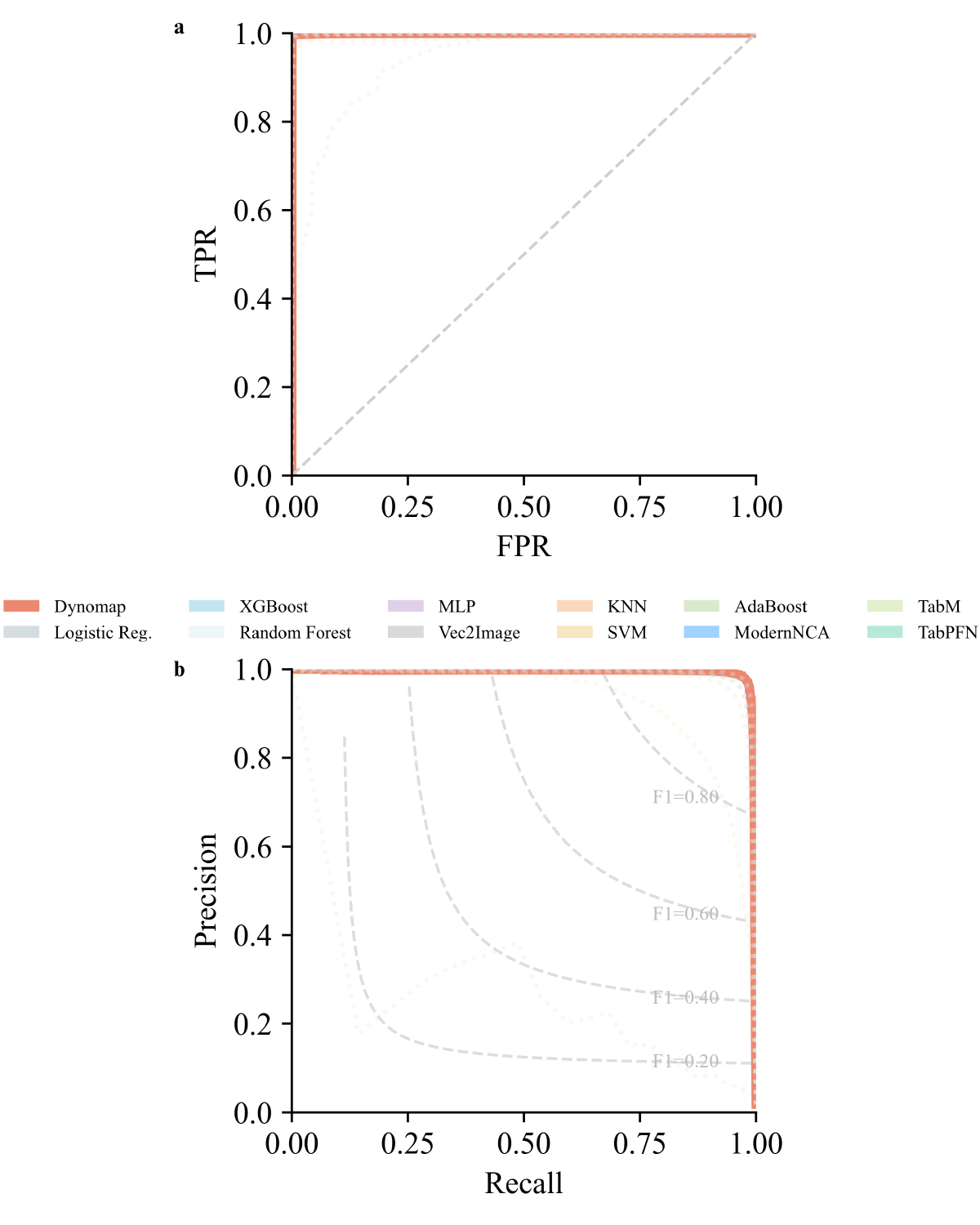}
\caption{\textbf{Precision-Recall and ROC analysis for Tabula Muris cell type classification.} \textbf{a}, Receiver Operating Characteristic (ROC) curve showing the True Positive Rate (TPR) against the False Positive Rate (FPR). Dynomap (solid orange line) achieves near-perfect area under the curve (AUC), significantly outperforming baseline methods. \textbf{b}, Precision-Recall curves across various cell types. Dynomap maintains high precision across the entire recall range, demonstrating robust classification even for rare or heterogeneous cell populations.}
\label{fig:figS22_tm_pr_roc}
\end{figure}

\begin{figure}[pt]
\centering
\includegraphics[width=0.8\linewidth, page = 25]{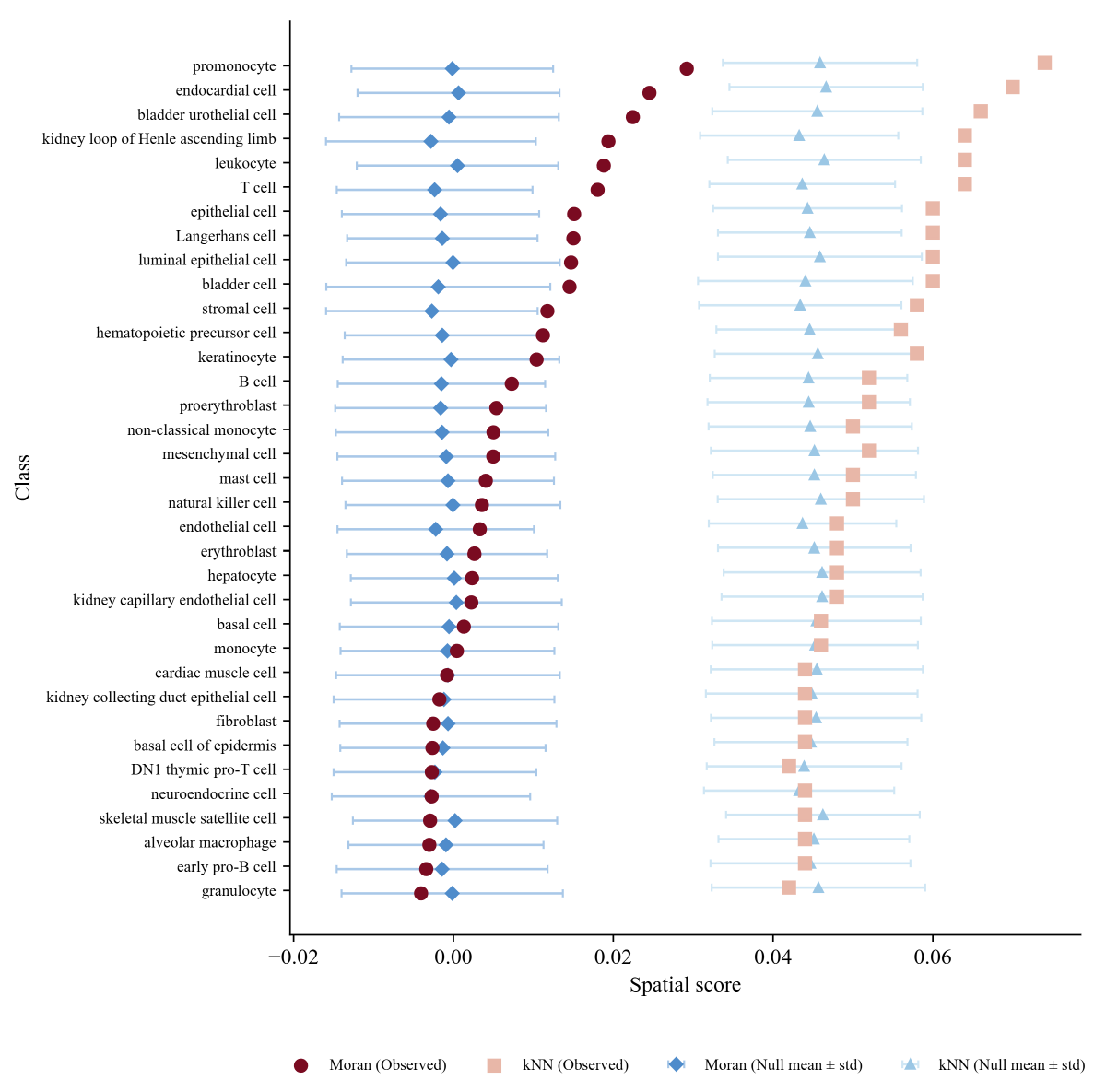}
\caption{\textbf{Quantification of spatial organization across cell lineages in the Tabula Muris atlas.} Observed Moran’s I (spatial autocorrelation, dark red circles) and kNN purity (neighborhood consistency, light orange squares) are compared against null distributions (blue diamonds and triangles with error bars). For nearly all cell types, including immune and stromal populations, the observed scores significantly exceed null expectations. This confirms that cell-type-specific transcriptional programs are organized into coherent spatial neighborhoods within the learned Dynomap layout.}
\label{fig:figS23_tm_spatial_stats}
\end{figure}

\begin{figure}[pt]
\centering
\includegraphics[width=\linewidth, page = 26]{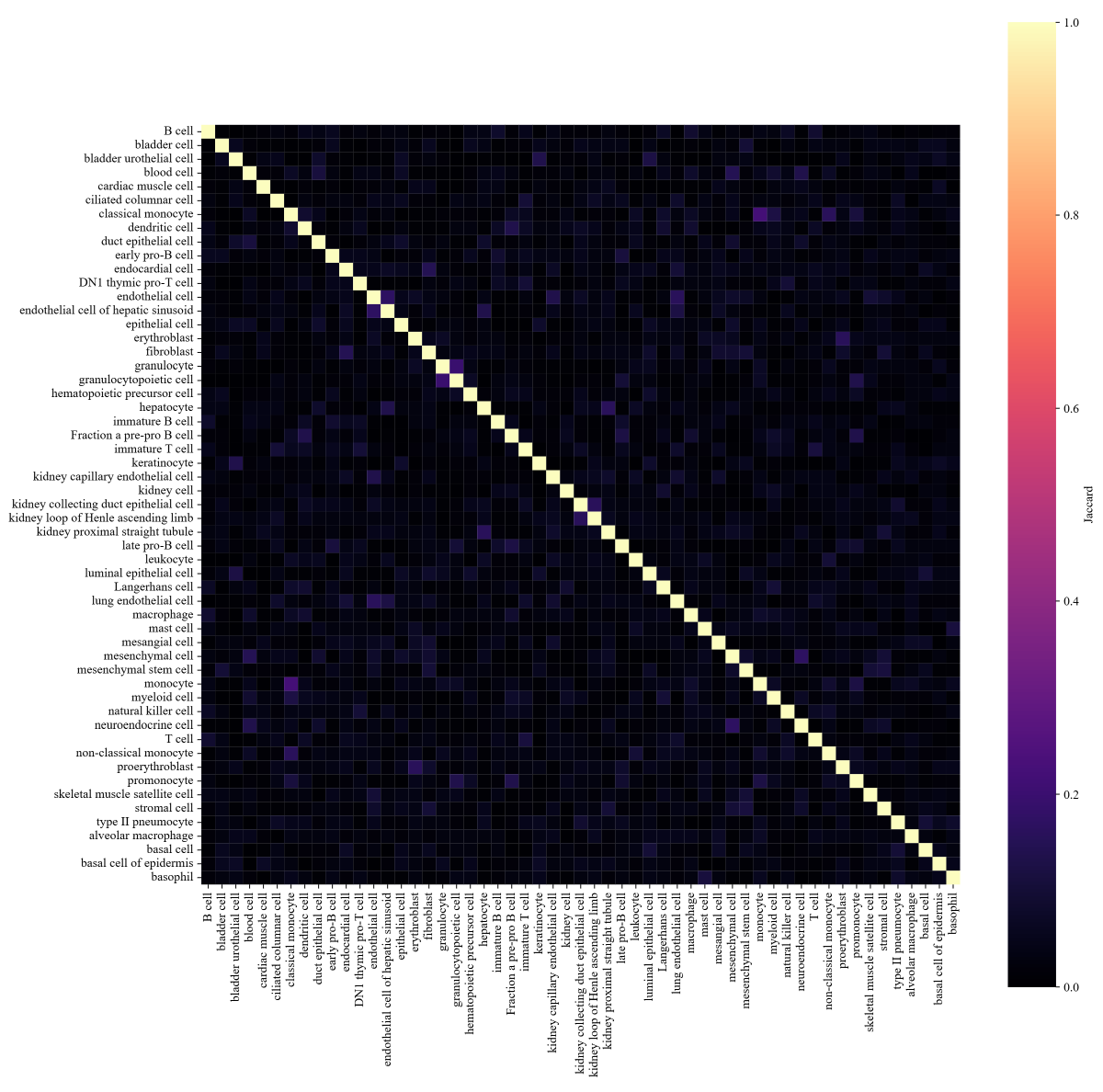}
\caption{\textbf{Jaccard similarity matrix of learned spatial neighborhoods for cell types.} Heatmap values represent the overlap between the top predictive feature sets of different cell types. The strong diagonal indicates highly stable and distinct spatial neighborhoods for individual classes. Sparse off-diagonal signals reveal biologically expected similarities between related cell lineages, such as different subtypes of immune or endothelial cells.}
\label{fig:figS24_tm_jaccard}
\end{figure}

\begin{figure}[pt]
\centering
\includegraphics[width=0.32\linewidth, page = 27]{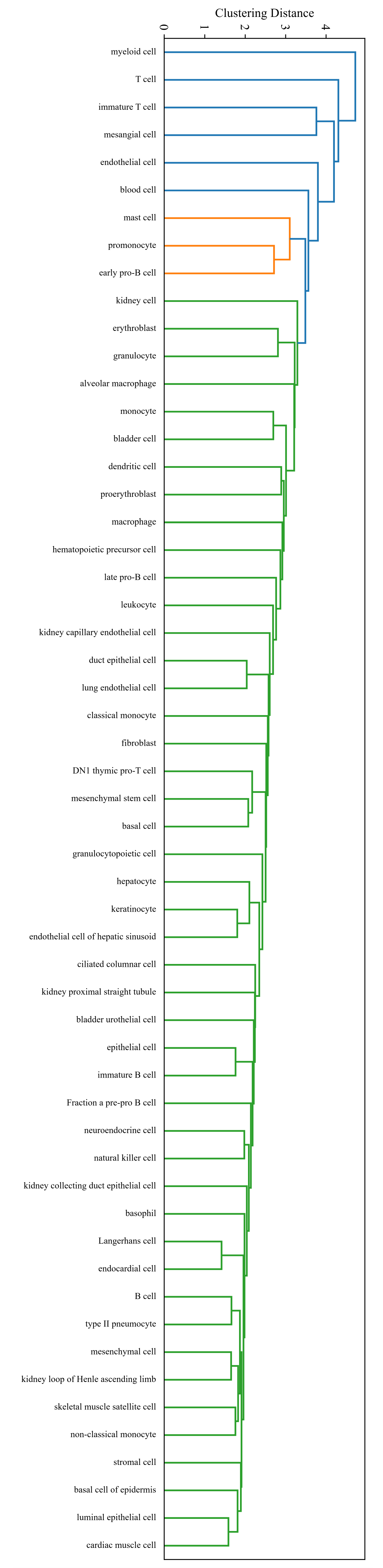}
\caption{\textbf{Hierarchical clustering of cell type attribution profiles.} This dendrogram illustrates the relationships between cell types based on their Integrated Gradients (IG) feature importance. Closely related cell types, such as various B cell stages or lung-specific populations, cluster together. This demonstrates that the Dynomap framework captures a hierarchical representation of biological identity directly from single-cell transcriptomic data.}
\label{fig:figS25_tm_class_dendrogram}
\end{figure}

\begin{figure}[pt]
\centering
\includegraphics[width=0.35\linewidth, page = 28]{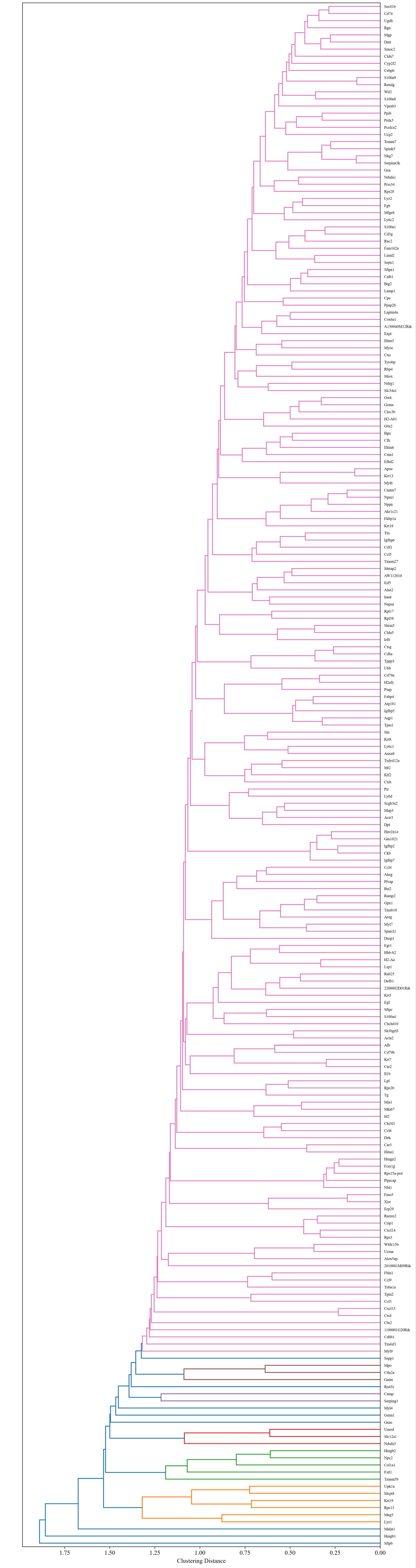}
\caption{\textbf{Hierarchical organization of predictive genes in the Tabula Muris dataset.} The dendrogram shows clustering of genes based on their spatial proximity and attribution patterns across cell types. Functional gene modules associated with specific cellular processes form distinct branches. This indicates that Dynomap learns to group genes with similar biological roles into organized spatial structures.}
\label{fig:figS26_tm_gene_dendrogram}
\end{figure}

\begin{figure}[pt]
\centering
\includegraphics[width=0.95\linewidth, page = 29]{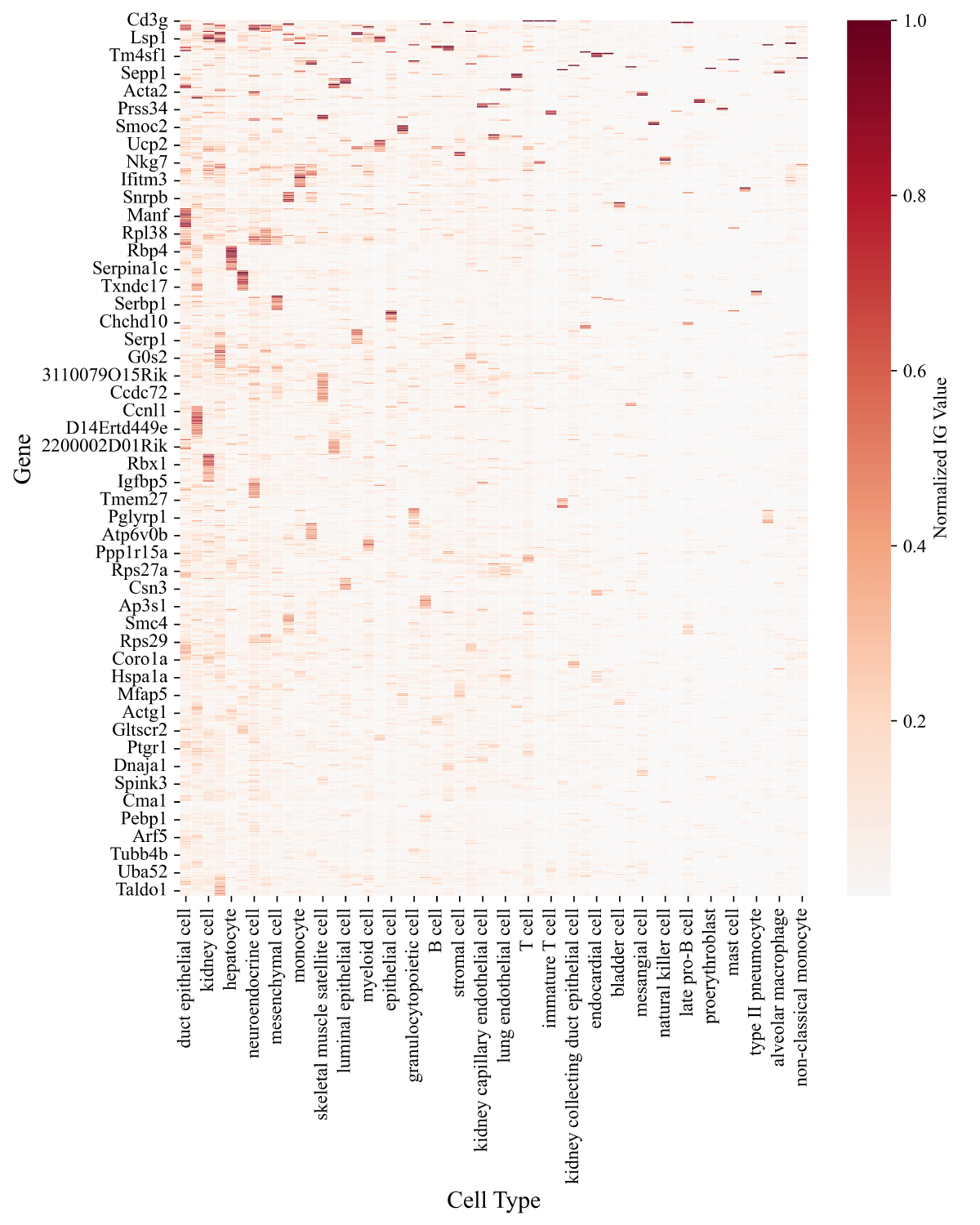}
\caption{\textbf{Comprehensive feature attribution heatmap for single-cell lineages.} The heatmap displays normalized Integrated Gradients (IG) values for top predictive genes (rows) across dozens of cell types (columns). Distinct vertical bands highlight marker genes uniquely utilized by the model for specific lineages, such as \textit{Cd3g} for T cells and \textit{Hepatocyte}-specific markers. The sparse structure confirms that the model relies on specialized transcriptional signatures to distinguish between complex cellular identities.}
\label{fig:figS27_tm_heatmap}
\end{figure}

\begin{figure}[pt]
\centering
\includegraphics[width=0.95\linewidth, page = 30]{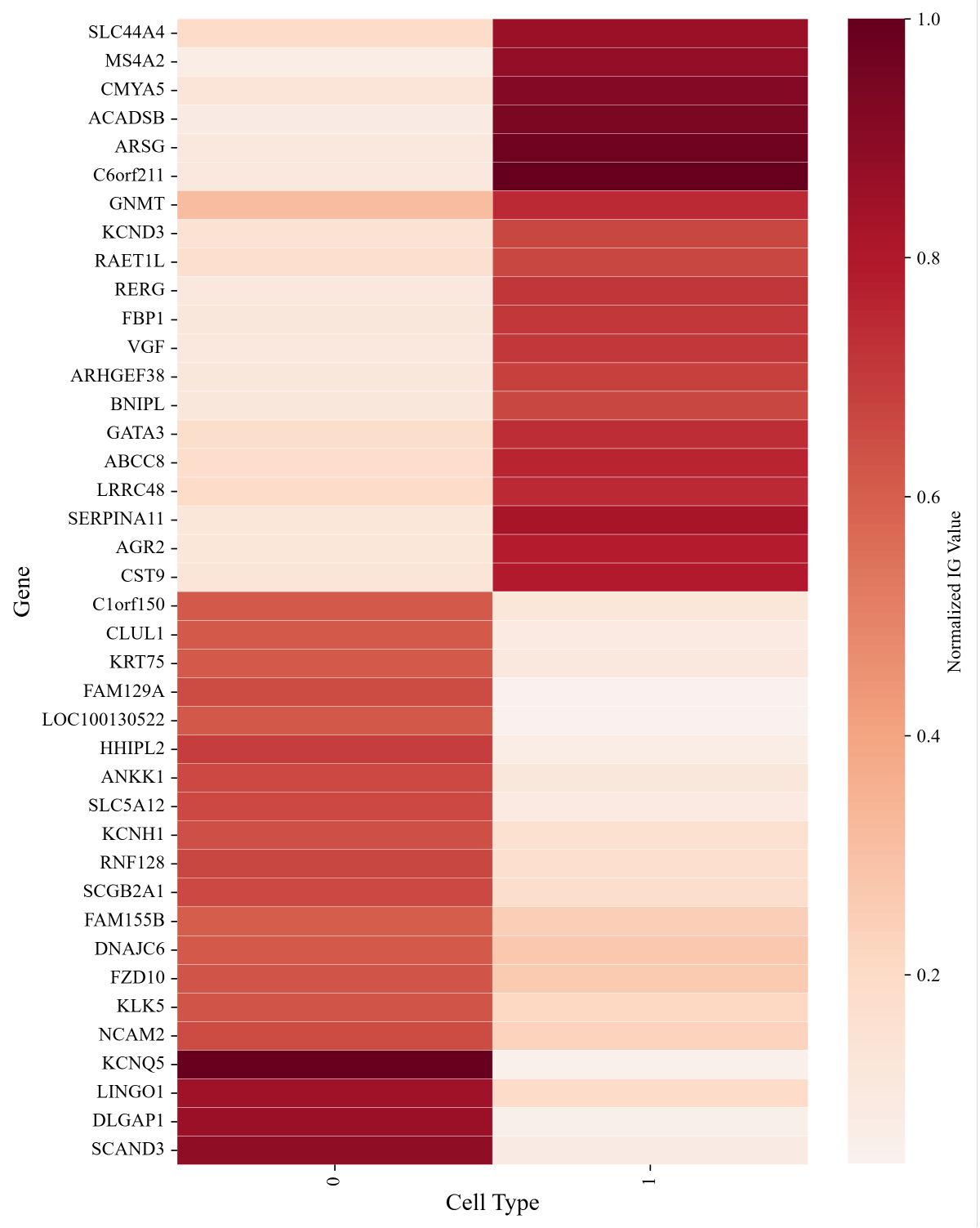}
\caption{\textbf{Class-specific gene attribution for breast cancer risk stratification.} Heatmap of normalized Integrated Gradients (IG) for the top 50 predictive genes in the binary task (Aggressive vs. Non-aggressive). The model utilizes distinct sets of genes to identify aggressive phenotypes, including \textit{C6orf211} and \textit{ARSG}, while non-aggressive tumors are characterized by a different molecular signature featuring \textit{KCNQ5} and \textit{DLGAP1}. The clear vertical separation of high-attribution scores confirms that Dynomap learns non-overlapping transcriptional programs for clinical risk assessment.}
\label{fig:figS28_tcga_binary_heatmap}
\end{figure}

\begin{figure}[pt]
\centering
\includegraphics[width=0.95\linewidth, page = 31]{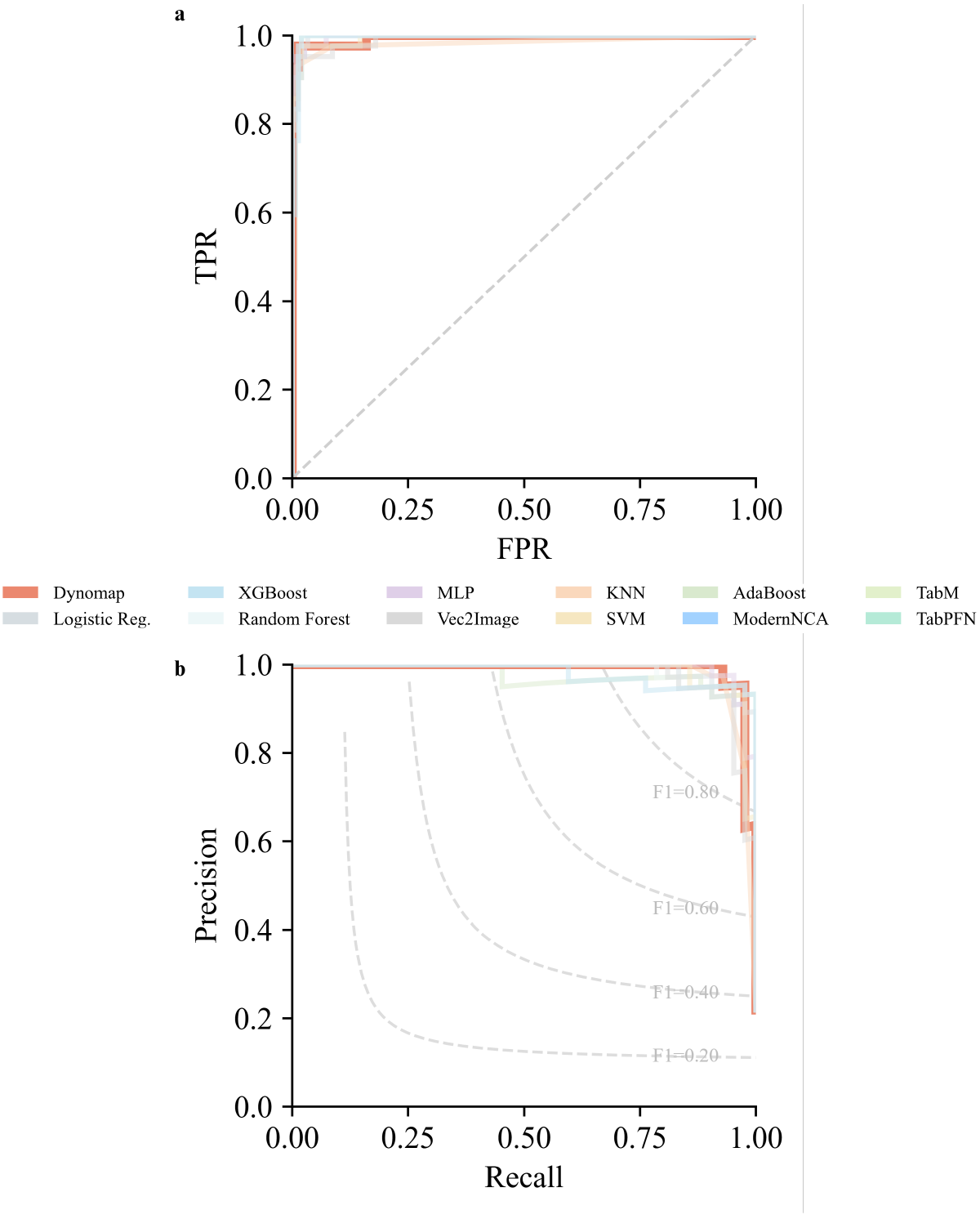}
\caption{\textbf{Predictive performance for binary risk stratification in TCGA-BRCA.} \textbf{a}, Receiver Operating Characteristic (ROC) curve and \textbf{b}, Precision-Recall (PR) curve for the Aggressive vs. Non-aggressive classification task. Dynomap (solid line) achieves high area under the curve (AUC) values, demonstrating robust discrimination even in the presence of tumor heterogeneity. The high precision maintained across recall levels indicates that the model's risk assessments are highly reliable for aggressive cases.}
\label{fig:figS29_tcga_binary_performance}
\end{figure}

\begin{figure}[pt]
\centering
\includegraphics[width=0.95\linewidth, page = 32]{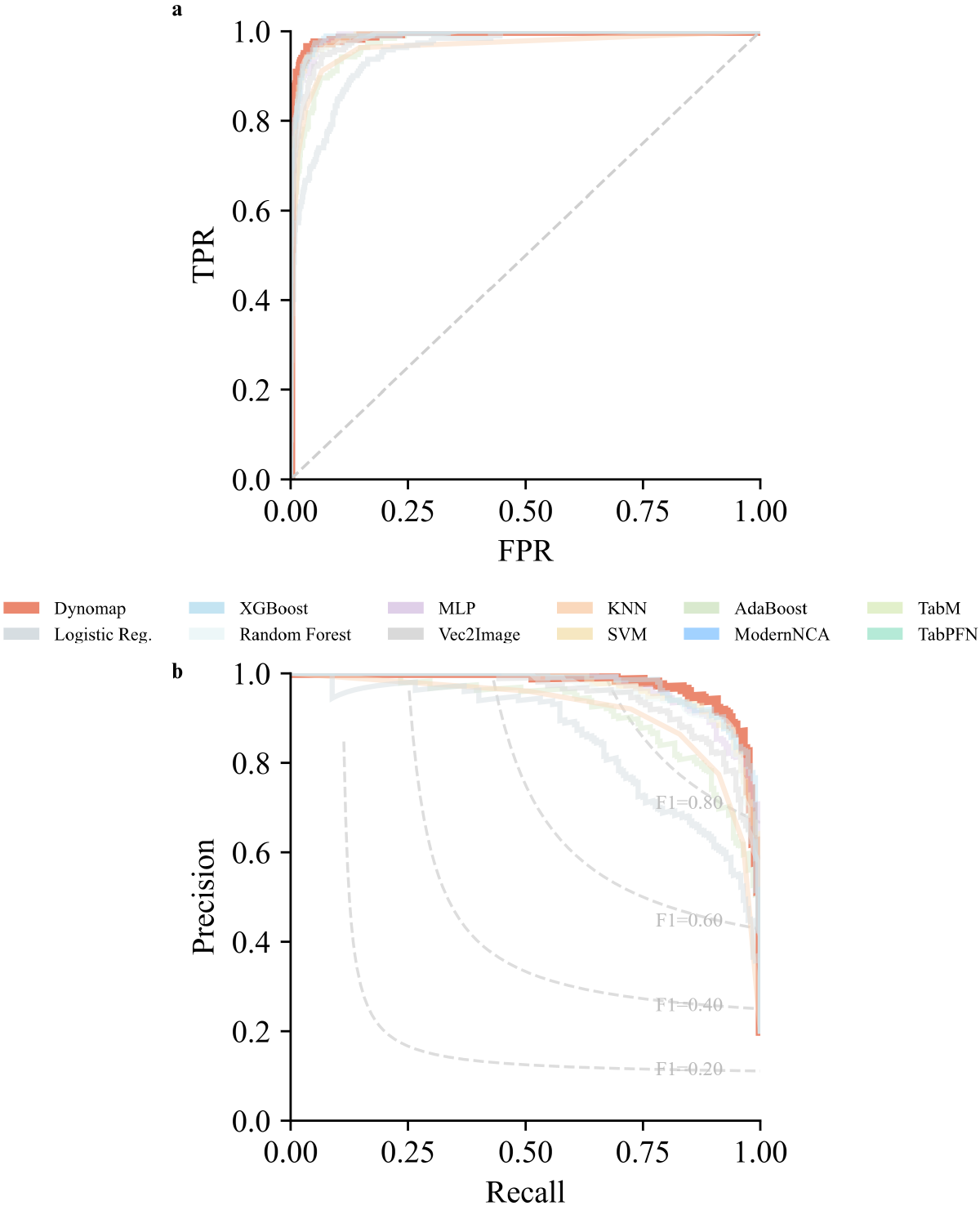}
\caption{\textbf{Multiclass performance for breast cancer molecular subtype prediction.} \textbf{a}, ROC and \textbf{b}, Precision-Recall curves for five intrinsic subtypes (LumA, LumB, Basal, HER2, and Normal-like). Dynomap maintains superior precision-recall trade-offs across all subtypes compared to baseline averages. Performance is particularly strong for the Basal and HER2-enriched subtypes, which exhibit the most distinct molecular signatures in the learned spatial representation.}
\label{fig:figS30_tcga_subtype_performance}
\end{figure}

\begin{figure}[pt]
\centering
\includegraphics[width=0.5\linewidth, page = 33]{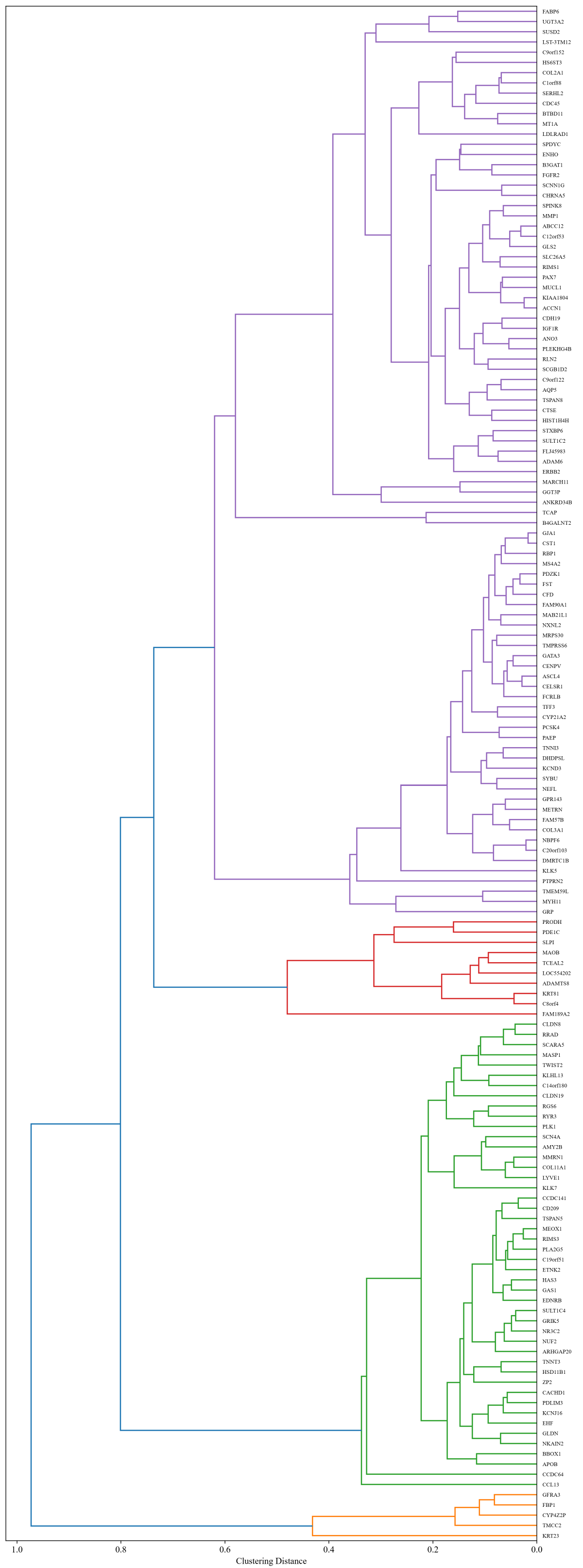}
\caption{\textbf{Hierarchical relationship of breast cancer subtype attribution profiles.} The dendrogram illustrates the clustering of molecular subtypes based on similarities in their feature attribution (Integrated Gradients) profiles. Luminal A and Luminal B subtypes cluster together, reflecting their shared hormonal signaling pathways, while the Basal-like subtype forms a distinct branch. This hierarchical structure confirms that the model’s internal logic aligns with established clinical and biological relationships between tumor subtypes.}
\label{fig:figS31_tcga_dendrogram}
\end{figure}

\begin{figure}[pt]
\centering
\includegraphics[width=0.95\linewidth, page = 34]{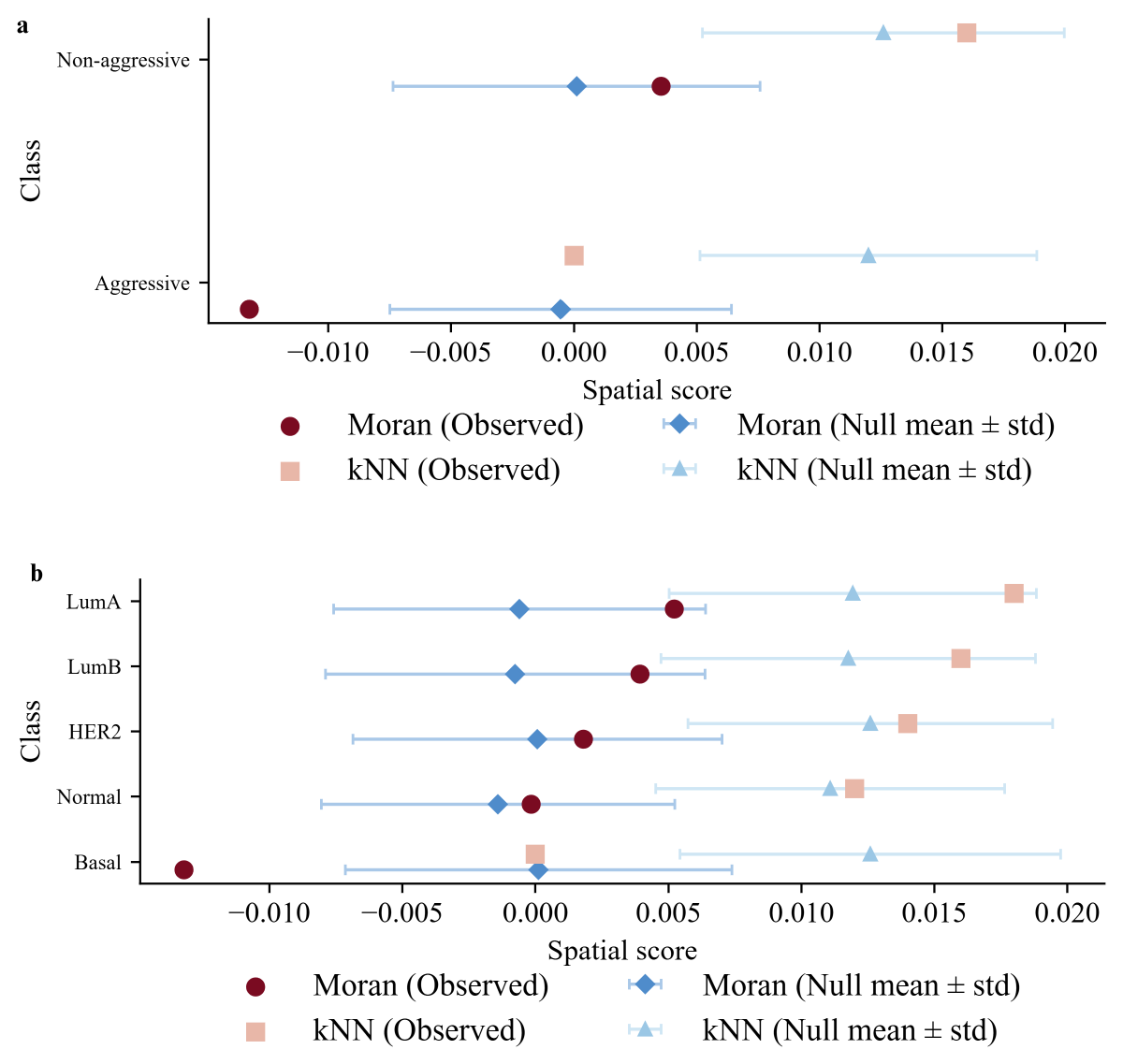}
\caption{\textbf{Spatial quantification of learned representations in TCGA-BRCA.} Observed Moran’s I (spatial autocorrelation) and kNN purity (neighborhood consistency) for the binary (red) and multiclass (blue) tasks are compared against null distributions. In both experimental settings, the observed metrics significantly exceed random expectations. This demonstrates that Dynomap consistently organizes subtype-specific and risk-associated genes into coherent spatial neighborhoods regardless of the classification objective.}
\label{fig:figS32_tcga_spatial_stats}
\end{figure}

\begin{figure}[pt]
\centering
\includegraphics[width=0.95\linewidth, page = 35]{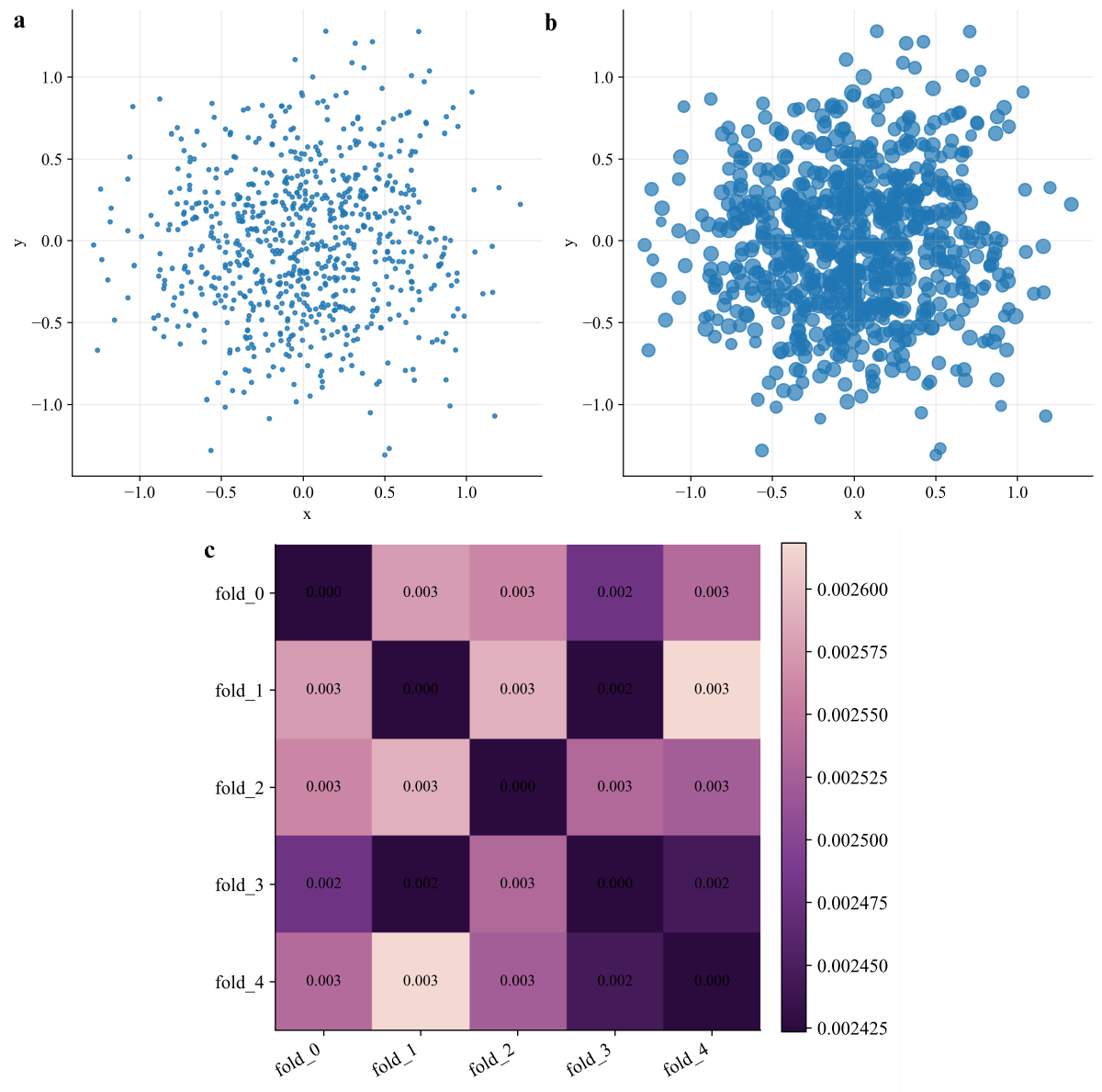}
\caption{\textbf{Reproducibility and stability of learned spatial layouts in Parkinson’s disease classification.}
a, Mean consensus layout across repeated training runs. Feature coordinates are averaged following Procrustes alignment to account for rotational and translational invariance. The resulting configuration demonstrates stable spatial organization of phenomic features across independent model initializations.
b, Layout consensus uncertainty. The spatial dispersion of each feature across repeated runs is quantified as coordinate variance after alignment. Most features exhibit low positional variance, indicating convergence toward a consistent spatial configuration.
c, Pairwise Procrustes distance matrix between independently trained layouts. Distances are computed after optimal alignment of feature coordinates. Low pairwise distances confirm high structural similarity between learned layouts across runs, demonstrating that the spatial representation is reproducible rather than driven by random initialization.}
\label{fig:figS33_parkinson_spatial_stats}
\end{figure}

\end{document}